\documentclass[acmsmall]{acmart}
\usepackage{natbib}
\usepackage{hyperref}

\AtBeginDocument{%
  }

\setcopyright{none}
\renewcommand\footnotetextcopyrightpermission{}
\begin{document}

\title{Multimodal Large Language Models for Image, Text, and Speech Data Augmentation: A Survey}

\author{Ranjan Sapkota*}
\affiliation{%
  \institution{Cornell University}
  \city{Ithaca}
  \state{New York}
  \country{USA}}
\email{rs2672@cornell.edu}

\author{Shaina Raza}
\affiliation{%
  \institution{Vector Institute}
  \city{Toronto}
  \state{Ontario}
  \country{Canada}}

\author{Maged Shoman}
\affiliation{%
  \institution{University of Tennessee}
  \city{Knoxville}
  \state{Tennessee}
  \country{USA}}

\author{Achyut Paudel}
\affiliation{%
  \institution{Orchard Robotics}
  \city{Bellevue}
  \state{Washington}
  \country{USA}}

\author{Manoj Karkee}
\affiliation{%
  \institution{Cornell University}
  \city{Ithaca}
  \state{New York}
  \country{USA}}
\email{}

\renewcommand{\shortauthors}{Sapkota et al.}

\begin{abstract}
In the past five years, research has shifted from traditional Machine Learning (ML) and Deep Learning (DL) approaches to leveraging Large Language Models (LLMs) , including multimodality, for data augmentation to enhance generalization, and combat overfitting in training deep convolutional neural networks. However, while existing surveys predominantly focus on ML and DL techniques or limited modalities (text or images), a gap remains in addressing the latest advancements and multi-modal applications of LLM-based methods. This survey fills that gap by exploring recent literature utilizing multimodal LLMs to augment image, text, and audio data, offering a comprehensive understanding of these processes. We outlined various methods employed in the LLM-based image, text and speech augmentation, and discussed the limitations identified in current approaches. Additionally, we identified potential solutions to these limitations from the literature to enhance the efficacy of data augmentation practices using multimodal LLMs. This survey serves as a foundation for future research, aiming to refine and expand the use of multimodal LLMs in enhancing dataset quality and diversity for deep learning applications. 
Paper GitHub: https://github.com/WSUAgRobotics/data-aug-multi-modal-llm.
\end{abstract}

\maketitle

\section{Introduction}

\begin{figure}[htbp]
    \centering
    \includegraphics[width=0.75\linewidth]{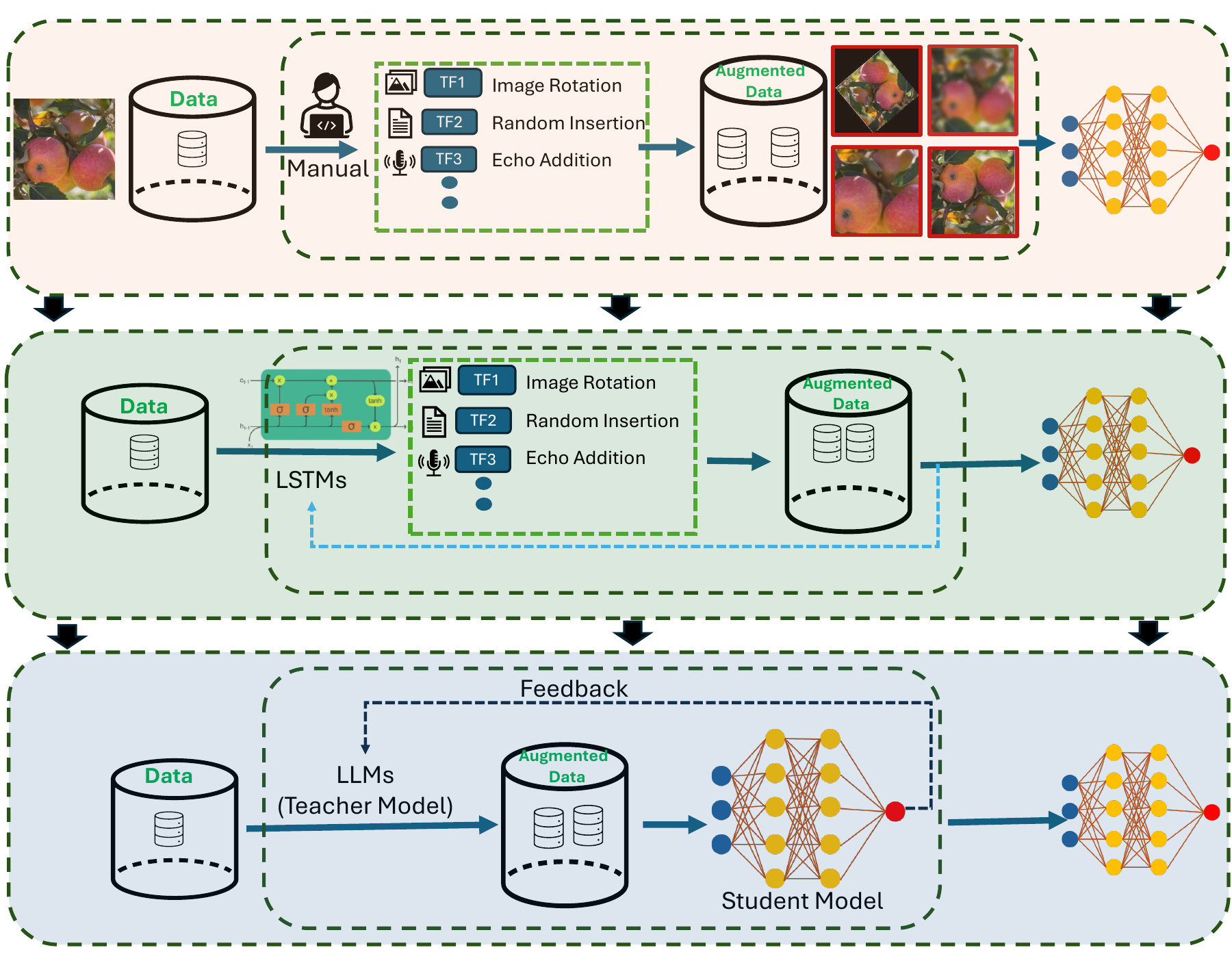}
    \caption{Evolution of data augmentation techniques (from top to bottom) a) Manual transformation functions like image rotation for training dataset expansion. b) LSTM-based automation generating synthetic data. c) Use of Generative LLMs for advanced, context-aware synthetic data creation, marking a shift to AI-driven data augmentation methods.}
    \label{fig:importanceofDataAugmentation}
\end{figure}

Data augmentation is a fundamental technique in machine learning (ML) that enhances the size and diversity of training datasets through the generation of modified versions of existing data samples \cite{iglesias2023data, shorten2019survey}. This practice uses various transformation functions (TFs), methods like rotating images or changing words, that adjust the original data to produce new variations, as illustrated in Figure \ref{fig:importanceofDataAugmentation}. 
Data manipulation experts applied these TFs manually to generate new examples that help train deep learning models more effectively. Specifically, TFs such as image rotation, gaussian blur, zooming in/out  (Figure \ref{fig:importanceofDataAugmentation}) could transform a single image of apples from orchards into multiple orientations, effectively increasing the dataset size for models focused on image processing. Similar TFs for text and audio data, like random insertion and specific audio modifications, broaden datasets for natural language processing (NLP) and audio analysis applications, respectively \cite{ratner2017learning, mumuni2022data}. These augmentation strategies not only increase the volume of data available for training but also introduce a spectrum of variations that models might encounter in real-world scenarios, thereby enhancing their robustness and generalization capabilities. This foundational practice is depicted through a visual representation in Figure \ref{fig:importanceofDataAugmentation} a, showcasing the augmentation of image data with rotated apple images as an example.

Traditionally, data augmentation was performed manually, but with the advent of Long Short-Term Memory (LSTM) networks,it has become a more automated and widespread practice \cite{iwana2021empirical, maharana2022review}. LSTMs facilitate the automatic generation of synthetic data across various applications, including time series forecasting, natural language processing (NLP), and human activity recognition\cite{bayer2023data}. This shift reduced the reliance on manual data creation, as illustrated in Figure \ref{fig:importanceofDataAugmentation} (b), which depicts the transition to LSTM-based data augmentation.

The LSTM-based augmentation became a cornerstone in data-driven fields until the emergence of Large Language Models (LLMs) and generative AI. With the surge in popularity following innovations like ChatGPT, LLMs have begun to redefine data augmentation, particularly by integrating and automating cross-modal synthesis. As shown in Figure \ref{fig:importanceofDataAugmentation} (c), this new era leverages the contextual intelligence of multi-modal LLMs to perform data augmentation. These methods also  move beyond traditional and LSTM-based methods by providing more sophisticated, contextually aware synthetic data generation across multiple data types. 

Data augmentation is crucial for enhancing the robustness and performance of DL models across various domains such as computer vision, NLP, and speech recognition. In computer vision, techniques like random cropping and flipping are normally used to prevent overfitting by promoting generalization across different orientations and expressions \cite{kaur2021data}. Similarly, in NLP, synonym replacement and paraphrasing help models to generalize across diverse lexical and linguistic formulations, crucial for applications like sentiment analysis and chatbot interactions \cite{abonizio2021toward, shorten2019survey}. In the realm of speech recognition, strategies such as noise injection enable models to perform reliably in noisy environments by simulating various acoustic scenarios \cite{pervaiz2020incorporating, veluri2024look}. 

Data augmentation enables model training under varied conditions, like lighting in autonomous driving or medical scenarios in imaging, reducing reliance on costly data collection.\cite{zhang2023perception, jockel2019safe, shin2018medical, chlap2021review}. It also addresses class imbalance and enhances dataset diversity, which is important for tasks requiring high levels of accuracy in real-world setting, such as machine translation and sound identification \cite{ csahin2022augment, chen2023empirical}. Data augmentation synthetically increases training data, reducing costs, speeding development, and maximizing data resource ROI \cite{shorten2019survey, wang2020survey}.

Building on the existing foundation of data augmentation methods, the advent of multi-modal LLMs has brought a lot of changes in the field. These models go beyond traditional applications such as machine translation and sentiment analysis, introducingpseudo data generation for classification and dataset enhancement for regression analysis \cite{ding2024data, sufi2024generative}. This shift introduces a transition to more dynamic and functional data augmentation techniques that not only diversify available methods but also deepen our understanding model training and performance \cite{hu2023survey}. 

\textbf{Necessity of this survey}
While numerous review articles on data augmentation in AI research have explored various techniques, most focus on traditional ML and DL approaches \cite{mumuni2022data, maharana2022review, iwana2021empirical, kaur2021data, chlap2021review, gracia2023data, al2024speech, abayomi2022data, liu2020survey, wang2020survey, cauli2022survey, bayer2022survey, kumar2024image, fayaz2024advancements, mumuni2024survey}, including GAN-based methods \cite{shin2018medical, sheng2018data, qian2019data, wali2022generative}. However, these studies often focus on a single modality, such as NLP or image processing. This survey addresses the gap by exploring ML and DL techniques across three modalities, Image, Text, and Speech, while also covering the latest advancements in LLMs and generative AI methods.
This survey critically evaluates the diverse methodologies for data augmentation that have emerged over the past  five years. In particular, with the rapid advancements and capabilities of LLMs from 2020 to the present, significant transitions in these methods have taken place. We, particularly, focus on the application of multimodal LLMs in data augmentation in generating coherent and contextually relevant synthetic data.

\textbf{Key Contributions}
The main contributions of this survey are summarized as follows:

\begin{enumerate}
    \item \textbf{Coverage of Multi-modality and LLM-Based Data Augmentation Methods:} To the best of our knowledge, this is the first survey to comprehensively cover three key modalities in ML research: Image, Text, and Speech. Additionally, it offers an in-depth exploration of the technical methods employed for data augmentation using LLMs across these modalities.  We identify and discuss the limitations and challenges inherent in current LLM-based data augmentation techniques for all three data modalities.
    
    \item \textbf{Ethical AI Research Conduct}: This survey adheres to the principles of ethical AI research, ensuring transparency, fairness, accountability, and integrity throughout the research process. We maintain these standards by carefully collecting literature, respecting copyright laws, and designing the study to be fully reproducible.

    \item \textbf{Analysis, Challenges, and Solutions:} We categorically present the results of our literature research for each modality, highlight the limitations and challenges, and propose potential solutions with the aim of advancing the field.
\end{enumerate}
The survey is structured as follows: Section \ref{method} outlines the \textit{Methodology}. Section \ref{background} provides a brief \textit{Background}, categorizing data augmentation techniques into traditional methods (1990–2010) and ML/DL methods (2010–2020). Section \ref{results}, \textit{Results and Discussion}, presents findings for Image, Text, and Speech modalities, analyzing LLM-based augmentation techniques, their limitations, and potential solutions. Finally, Section \ref{conclusion}, \textit{Conclusion}, synthesizes insights, discusses future research directions, and highlights the evolution of LLM applications in data augmentation.

\section{ Literature Review Methodology}
\label{method}
To compile a comprehensive and relevant list of papers for our review, we conducted a comprehensive literature review,  adhering to established methodology principles \cite{carrera2022conduct}. Our search query and extraction methods are detailed below:
\subsection{Databases Searched}
\begin{figure}[htbp]
    \centering
    \includegraphics[width=0.98\linewidth]{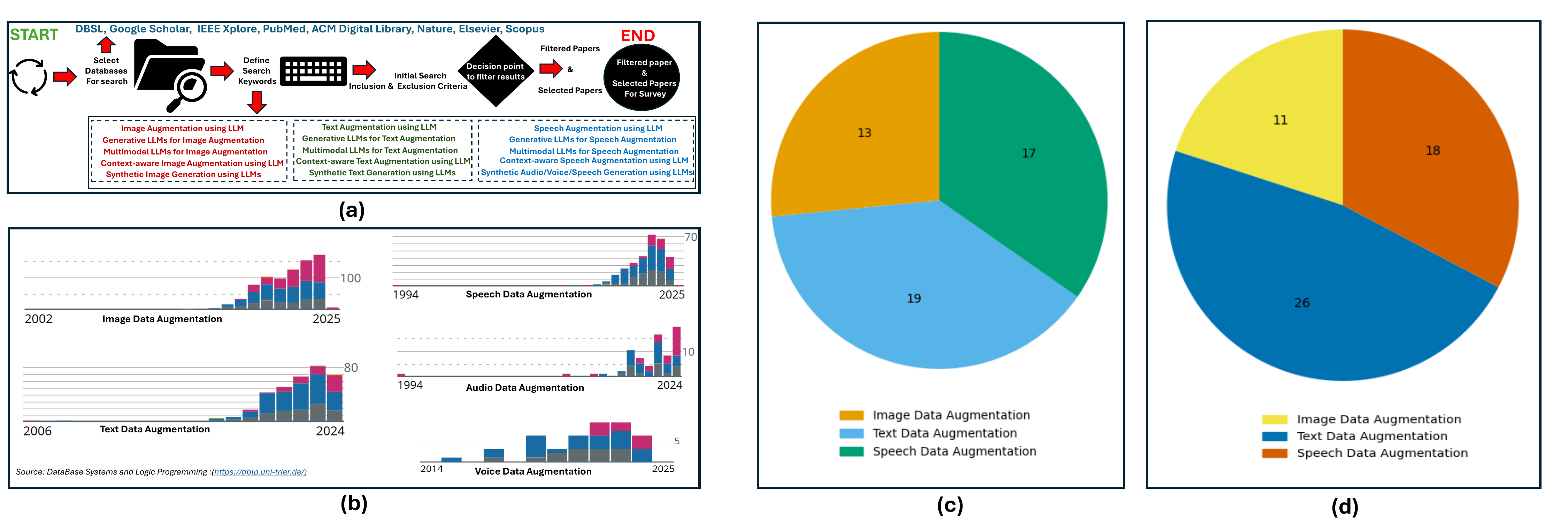}
    \caption{Survey methodology and results overview:    \textbf{(a)} Flowchart of the structured survey method, showing specific search keywords and steps from initial search to study selection; \textbf{(b)}  Graph displaying initial search outcomes across the DBSL database, illustrating paper distribution;  \textbf{(c)} Pie chart of selected peer-reviewed papers, showing thorough screening; \textbf{(d)} Pie chart of preprints distribution, indicating extensive preliminary research review. Keywords for image, text, and speech augmentation are marked in red, green, and blue, respectively.}
    \label{fig:searchresults}
\end{figure}

We conducted an extensive literature review across several renowned academic databases, starting with the DataBase systems and Logic Programming (DBSL) platform\footnote{\url{https://dblp.uni-trier.de/}},which  a comprehensive computer science bibliography. Subsequent searches were carried out across Google Scholar, IEEE Xplore, PubMed, ACM Digital Library, Nature, Elsevier, and Scopus, utilizing a color-coded keyword strategy as illustrated in Figure \ref{fig:searchresults} (a).  

\subsection{Search Keywords}
This search yielded 766 papers on "image data augmentation", 336 on "text data augmentation", and 414 on speech-related augmentations using a variety of keywords, including "speech", "audio", "voice", "synthetic data generation", "augmentation techniques", "data preprocessing", \textit{"noise injection"}, \textit{"feature transformation"}, and \textit{"data synthesis"}.

\subsection{Inclusion Exclusion Criteria}
We adhere to ethical guidelines for data search and collection \cite{nihguidingprinciples}, categorizing the historical development of data augmentation into distinct eras. 
The period from 1990 to 2010 represents the era of traditional methods, characterized by basic techniques. From 2010 to 2020, the focus shifted to advanced machine learning and deep learning (ML/DL) methods. Post-2020 marks the era dominated by LLMs, which have significantly enhanced the scope and effectiveness of data augmentation across various modalities. This historical framework, segmented by technological advancements in image, text, and speech data augmentation, guided our review. The distinctive methods and their evolution in each era were essential to understanding the current landscape shaped by LLMs, as demonstrated by the substantial literature retrieved during the initial search phase, shown in Figure \ref{fig:searchresults} (a).

The screening and selection process identified the most relevant studies for in-depth analysis, as shown in Figure \ref{fig:searchresults} (b), which summarizes the total papers retrieved for image, text, and speech (including audio and voice) data augmentation. These studies formed the basis for understanding the evolution of data augmentation methods across traditional, ML/DL, and LLM-based eras. Further refinement targeted studies utilizing LLMs for data augmentation, detailed in Figures \ref{fig:searchresults} (c) for peer-reviewed journals and \ref{fig:searchresults} (d) for preprints. These figures highlight the rigorous selection process that distilled the most pertinent contributions in the field.

\subsection{Quality Assessment}  
We evaluated the articles using the inclusion and exclusion criteria described earlier. In cases of uncertainty, the paper was briefly reviewed and subsequently included or excluded based on a consensus among the co-authors. The selected papers were then thoroughly reviewed and subjected to a quality assessment using a set of questions.  
An article was considered eligible for inclusion in our review if it received a “yes” or “partially” response to any of the following questions:  
\begin{enumerate}
    \item Does the article present a method for data augmentation in ML/DL/LLM?  
    \item Does the article present a method for data augmentation in image, text, or speech?  
    \item Does the article propose a framework, tool, or methodology?  
\end{enumerate}

\subsection{Results}
In total, our survey evaluated 24 studies focused on image data augmentation from 2020 onwards, which included 13 peer-reviewed papers and 11 pre-prints extensively discussing LLM-based augmentation techniques. For text data augmentation, 45 studies were reviewed, consisting of 19 peer-reviewed articles and 26 preprints. The speech data augmentation modality was represented by 35 studies, encompassing 17 peer-reviewed papers and 18 preprints. Besides, the comprehensive evaluation of the current research landscape, we also highlighted emerging trends and advancements in the application of LLMs.

\section{Background}
\label{background}
\subsection{Historical Context}
Data augmentation has undergone significant evolution since the 1990s, as depicted in the upper part of Figure \ref{fig:BackgroundAugmentation}, where traditional methods are highlighted in blue, red, and green boxes representing image, text, and speech augmentation respectively. Initially, these traditional techniques, such as image flipping, rotating, and scaling, were employed to enhance dataset variability and robustness. Entering the 2010s, a shift occurred towards more sophisticated ML and DLapproaches, illustrated in the lower part of Figure \ref{fig:BackgroundAugmentation}. These ML and DL methods introduced advanced, automated augmentation processes tailored to specific modalities, significantly improving the creation of diverse, realistic datasets that closely mimic real-world complexity. This section provides an overview of the progression from traditional to ML/DL-based data augmentation, setting the stage for our exploration and analysis of multimodal LLM-based data augmentation methods from 2020 onwards.

\subsection{Traditional Data Augmentation Methods (1990-2010)} 

\begin{figure*}[htbp]
    \centering
    \includegraphics[width=0.98\linewidth]{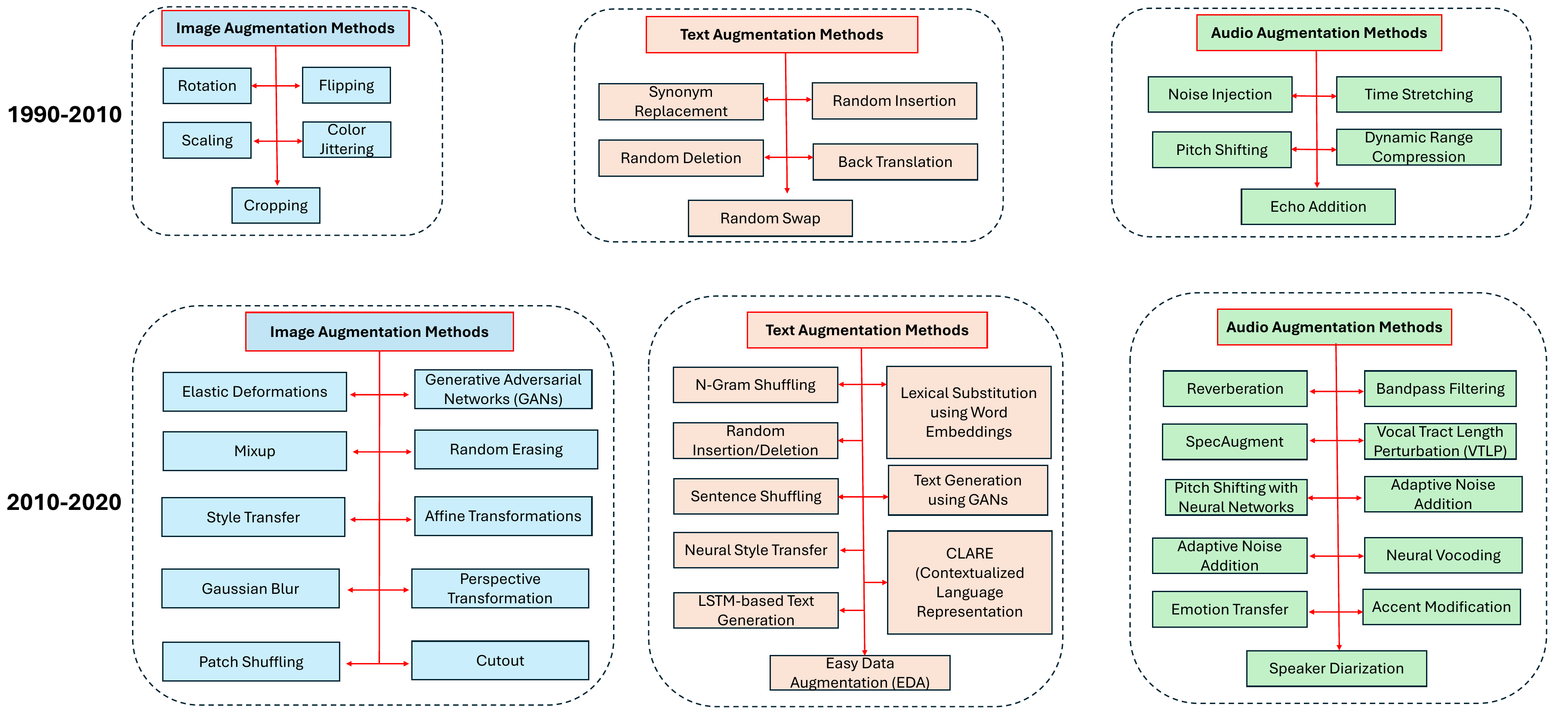}
    \caption{A comprehensive overview of data augmentation techniques, divided into two main eras: 1990 to 2010, focusing on traditional methods for image, text, and audio augmentation, and 2010 to 2020, highlighting m achine learning and deep learning-based advancements.}
    \label{fig:BackgroundAugmentation}
\end{figure*}

\begin{itemize}
    \item \textbf{Image Data Augmentation:} Traditional image augmentation methods from the 1990s to early 2000s primarily used geometric and color transformations to enhance ML training datasets, as shown in Figure \ref{fig:BackgroundAugmentation} \cite{joshi2008colour, starck2003gray}. These methods improved image quality and interpretability by overcoming the limitations of then-current imaging technology with several innovative techniques. Histogram hyperbolization was introduced to enhance contrast more effectively than standard methods by applying a memory-less nonlinear transformation tailored to human brightness perception \cite{frei1977image}. Additionally, non-recursive techniques that utilizes local statistics were developed for noise filtering and contrast enhancement, ideal for real-time applications \cite{lee1980digital}. Other techniques, such as included local histogram equalization and dynamic video gain adjustment for enhancing TV-type imagery in varying lighting \cite{ketcham1976real}, shock filters based on nonlinear differential equations for reducing oscillations and enhancing feature recognition \cite{osher1990feature}, and fuzzy set theory to improve image clarity and contrast by manipulating pixel properties \cite{pal1980image}.

    Traditional image data augmentation methods, while pioneering for their time, had limitations that could compromise the effectiveness of augmented datasets. Techniques such as image rotation and flipping were straightforward and computationally light, helping to enhance dataset size and variability, but often at the cost of losing important data or introducing artifacts like empty corners or misleading features for asymmetrical objects \cite{park2003super}. Scaling images to simulate depth perception proved useful for creating scale-invariant models; however, this often led to pixelation and resolution loss when images were enlarged \cite{gibson2002accurate, varma2008statistical}. Color jittering adjusted images to simulate different lighting conditions, but excessive alteration could render images unnatural \cite{rubio2006jittering}. Similarly, cropping and Gaussian blurring were employed to focus model attention and simulate blur effects respectively, yet these could remove vital image details or overly simplify textures, potentially undermining model training \cite{aghagolzadeh1992transform, goljan2008camera, greenspan2000image}.

    \item \textbf{Text Data Augmentation:} Traditional text data augmentation methods, uch as synonym replacement, random deletion, random insertion, back translation and random swapm have been typically used in the training of language models by introducing lexical and structural diversity.  Synonym replacement allows for the substitution of words with their synonyms, which helps models in grasping semantic similarities \cite{zhu2007text, edmonds1999semantic}. However, these replacements are sometimes contextually inappropriate, as synonym databases may not capture nuanced meanings or colloquialisms, which occasionally alter intended sentiments or textual nuances \cite{zhu2007text, edmonds1999semantic, barnard2005word}.  Random Deletion methods that removes words to mimic incomplete information, compelling models to deduce missing context, though excessive use risked omitting essential information and skewing data representation \cite{boulis2005text, ben2007addition, mcnamara1996good, nakagawa2008missing}.
Random Insertion is employed to enhance model robustness by inserting random words into unpredictable positions within the text. This method aims to prepare models for handling unexpected or out-of-context inputs. However, it compromises grammatical integrity and adding irrelevant content, which could muddy the training process \cite{kukich1992techniques, mcroy2003augmented}.  Back Translation is another traditional method to translate sentences into a foreign language and then back to the original language, resulting in subtly different paraphrases. This process was dependent on the quality of the translation systems and was noted for being both time-intensive and computationally demanding, especially for longer texts or rarer languages \cite{haritaoglu2001scene, yan2008augmenting, kay1997proper, kraaij2003embedding, fung1997technical}. Random Swap reorganized words within sentences to reduce the model's sensitivity to specific word order, benefiting languages with flexible syntax. However, this could also disrupt the natural flow of text and complicate the learning of language rules \cite{dhillon2002iterative, mcroy2003augmented}. Some of the popular text data augmentation methods during the 1990 to early 2000 are illustrated in upper mid part of Figure \ref{fig:BackgroundAugmentation}.
    
    \item \textbf{Speech Data Augmentation:} Traditional methods of speech data augmentation have been typicall used for auditory experiences across various applications since the early 1990s. Early studies like Cohen (1993) introduced augmented audio reality by merging computer-generated sounds with natural environments, enhancing telepresence and virtual reality interactions \cite{cohen1993augmented}. Other  contributions include Brandenburg et al. (1992), who used autogenous fat for vocal cord augmentation, showing improvements in voice quality \cite{brandenburg1992vocal}, and Watanabe (1985), who enhanced teleconferencing systems by integrating audio and visual information \cite{watanabe1985audio}. Schmandt (1990) also incorporated speech input into window systems, facilitating voice-controlled navigation \cite{schmandt1990augmenting}, and Adams (1992) utilized the Lombard effect to increase voice intensity in Parkinson’s patients \cite{adams1992can}.
    
  \textit{Challanges:}  These pioneering methods, while useful, face challanges such as alignment accuracy in synthetic audio with real-world environments, vulnerability to background noise, and substantial computational demands for real-time processing. Throughout the late 1990s and early 2000s, techniques like noise injection, pitch shifting, and dynamic range compression were developed to train robust audio recognition systems capable of operating under varied real-world conditions \cite{van2003speech, won2008improving, nishimura2006speech}. However, these methods often introduced artifacts, distorted audio signals, or overfitting to specific noise types, highlighting a persistent need for improvement in simulating authentic auditory environments for deep learning applications \cite{harma2004augmented, shu2008power, nishimura2006speech}. 

\end{itemize}

\subsection{Machine Learning and Deep Learning-based Data Augmentation (2010-2020)}

\begin{itemize}
    \item \textbf{Image Data Augmentation:} Over the past decade, DL/ML based image data augmentation techniques including the methods like rotations, flips, and crops (Figure \ref{fig:BackgroundAugmentation}) have been extensively used to train neural networks to be invariant to changes in object orientation, scale, and position, essential in sectors such as medical imaging \cite{yadav2019deep, altaf2019going, chlap2021review} and autonomous driving \cite{shu2021adversarial, jockel2019safe}. Advanced techniques have also emerged to handle increased task complexity. For instance, elastic deformations and affine transformations including scaling, shearing, and translating are employed to mimic real-world variations and perspective changes. Techniques like Gaussian blur and perspective transformations enhance model tolerance to sensor imperfections and variable viewpoints, while patch shuffling and methods such as cutout or random erasing challenge models by altering image contexts, compelling reliance on less obvious features for predictions \cite{su2021data, gracia2023data}.

    Additionally, innovative augmentation methods have driven substantial progress in fields requiring complex visual processing. MixUp technique blends images and labels to prevent overfitting by diluting pattern recognition, whereas style transfer and GANs generate stylistically varied or completely synthetic images to enrich training datasets, addressing issues like data scarcity and class imbalance \cite{jo2022dagam, sapkota2023creating}. These methods have found applications beyond traditional realms, enhancing video data robustness in surveillance, adjusting images in agriculture to predict crop health under varied conditions \cite{su2021data, gracia2023data}, and creating realistic patient data in healthcare for better disease prediction models \cite{hernandez2022synthetic, garcea2023data, sapkota2024multi}. 

\textit{Challenges:}However, despite these advances, challenges persist such as potential image quality degradation from techniques like rotation or scaling \cite{shijie2017research, mumuni2022data}, which can detrimentally affect training outcomes. Additionally, the complexity of generating sufficiently diverse images that accurately reflect real-world variability necessitates the use of resource-intensive generative models, highlighting the need for continued innovation in augmentation technology \cite{vyas2024data}. 

    \item \textbf{Text Data Augmentation:} From 2010 to 2020, the integration of ML/DL techniques advanced text data augmentation to a great extent. These methods are used to refine the processes that enhance text-based applications like machine translation, information retrieval \cite{bayer2022survey, bayer2023data}. The most famous DL/ML based text augmentation methods are illustrated in Figure \ref{fig:BackgroundAugmentation}. The period was marked by the application of statistical methods to linguistic databases, greatly improving the handling of textual data \cite{wang2020data}. For instance, early work by Fung in 1994, which augmented Chinese dictionaries with statistically collected character groups from corpora, significantly expanded domain-specific and regional word coverage, thereby enhancing natural language applications by enriching the lexicon with words, idioms, and technical compounds \cite{fung1994statistical}. Similarly, Dietterich's 1995 study demonstrated how learning algorithms such as ID3 and backpropagation could improve English text-to-speech systems by enhancing text mapping to phonemes and stresses \cite{dietterich1995comparison}. Additionally, Rubin's 1987 introduction of the Sampling/Importance Resampling (SIR) algorithm offered a non-iterative alternative for handling missing textual data, advantageous in settings with modest amounts of missing information \cite{rubin1987comment}.

\textit{Challanges:}    Despite these advancements, traditional text augmentation methods often struggled to generate contextually relevant data, as synthesized text frequently lacked the accuracy and fluency of human-generated language \cite{bayer2022survey, bayer2023data}. Simple techniques like synonym replacement, while initially effective, risked disrupting semantic coherence by introducing contextually inappropriate words, leading to model misunderstandings and reduced performance on downstream tasks \cite{liu2020survey, onan2023srl}. However, the evolution of text augmentation techniques, as illustrated in Figure \ref{fig:BackgroundAugmentation}, included diverse methods like n-gram shuffling, which added syntactic diversity, and lexical substitution using word embeddings, which enhanced lexical variety while maintaining semantic content \cite{troiano2023theories, liu2023summary}. Advanced approaches like neural style transfer and GANs were developed to produce high-quality, contextually accurate synthetic text, addressing issues of data scarcity and class imbalance in specialized tasks such as medical report generation \cite{haritaoglu2001scene, yan2008augmenting}.

    \item \textbf{Audio Data Augmentation:} The rapid evolution of DL technologies has greatly influenced fields such as NLP, computer vision, and especially speech recognition \cite{alam2020survey, khan2023exploring, al2024speech}. These models' effectiveness depends significantly on the availability of large, high-quality datasets \cite{braik2024automated, zuo2024machine}. In recent years, the advent of multi-modal LLMs has marked a significant shift in data augmentation strategies, introducing innovative techniques aimed at mitigate data scarcity that enhances the performance of models across various modalities \cite{raiaan2024review}. This advancement is vital for increasing data volume and improving model robustness and adaptability, thereby enabling models to perform reliably under real-world conditions and preventing overfitting \cite{khan2023exploring}. Furthermore, the role of data augmentation extends to supporting ethical AI development by fostering diversity in training datasets, which is crucial for reducing biases and promoting fairness, particularly in sensitive applications \cite{braik2024automated}. 

\textit{Challanges:}   Audio data augmentation poses unique challenges, particularly in replicating authentic sound environments, which are critical for developing effective speech recognition systems \cite{wei2020comparison, cohen2022study}. Traditional simple methods like noise addition often fall short of capturing the complexity of real acoustic environments \cite{zhou2022analysis}. Moreover, the lack of diverse and high-quality training samples across various languages and dialects significantly hampers the capability of these systems \cite{abayomi2022data}. Addressing these challenges necessitates the development of more sophisticated and rigourous data augmentation methods that can generate realistic and contextually appropriate synthetic audio data. These advanced methods are essential not only for enhancing the performance of speech recognition systems but also for ensuring they are adaptable and effective across different acoustic and linguistic contexts.

\end{itemize}

\section{Literature Review and Comparative Analysis}
\label{results}
In this section, we present the results of our literature review.
\subsection{LLM-based Image Data Augmentation}
\subsubsection{Process overview}
Image data augmentation represents a complex, multi-step process designed to significantly enhance machine learning models by diversifying the training datasets \cite{rayavarapu2024comprehensive, kaur2021data} . Initially, the process starts with Image Encoding, where raw images are converted into a computable format, typically using a vision encoder to distill essential visual information into feature vectors. This step is crucial for preparing images for further processing, as seen in studies like those by \cite{li2024itimca}, where image-text contrastive learning is utilized. 

Following the encoding, Prompt Generation involves the LLM generating a textual description of the encoded image. This textual prompt accurately reflects the image’s content and serves as a bridge to the next augmentation steps \cite{sapkota2024synthetic}. Augmentation Instruction Generation then takes these descriptions to create detailed transformation instructions. For instance, DALL-E \cite{sapkota2024synthetic} uses these prompts to generate synthetic images that closely mimic the textual description, enhancing datasets with varied visual representations. 

The generated instructions are translated into executable code during the Natural Language to Code Translation step. This code is applied to the original images in the Code Execution phase, effectively implementing the desired augmentations. This step is exemplified in the work by \cite{li2024itimca}, where augmented images are created to enhance disease identification accuracy in cassava leaves. Subsequent to the transformation, the Quality Assessment phase ensures the augmented images meet high-quality standards, crucial for maintaining the natural appearance and utility of the images in training scenarios, as noted by \cite{liu2024enhanced} in their work on breast cancer diagnosis. Metadata Generation follows, documenting the augmentation details, which is essential for reproducibility and further analytical work \cite{kirilenko2024generative}. 

\begin{figure*}[htbp]
    \centering
    \includegraphics[width=0.98\linewidth]{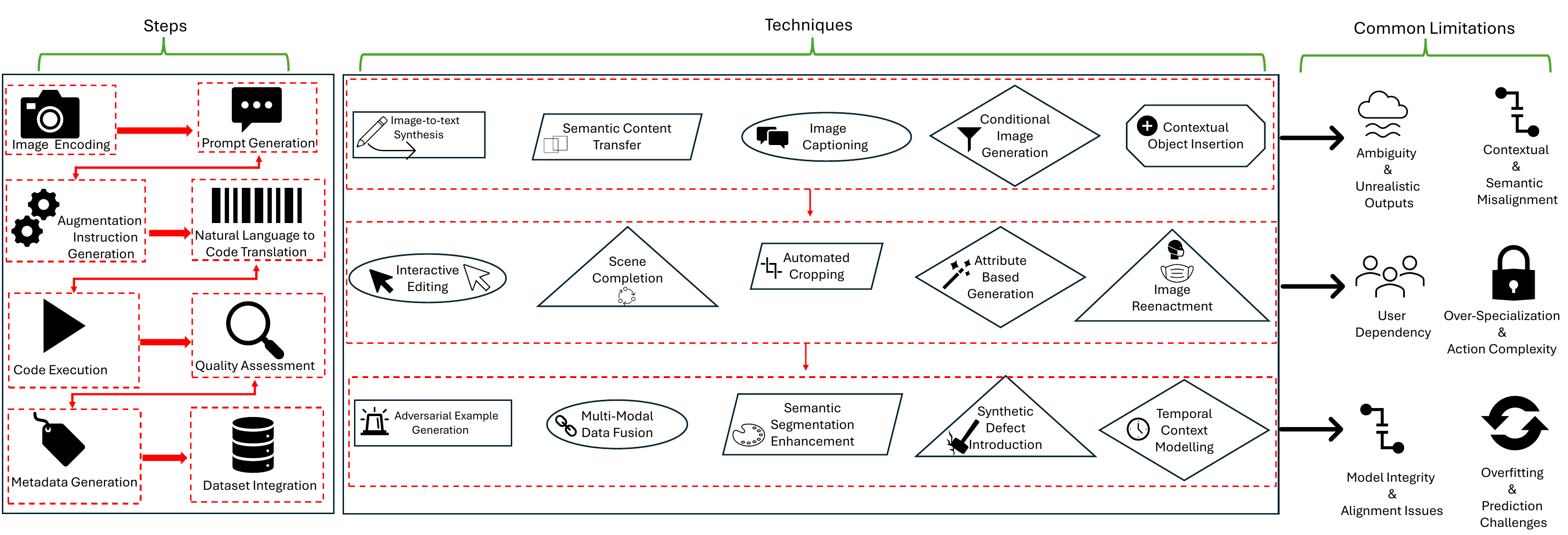}
    \caption{LLM based image data augmentation : showing the technical aspect of how image augmentation using LLM is performed, the techniques of augmenting images using LLM and their limitations }
    \label{fig:ImageLLMResults}
\end{figure*}

Finally, Dataset Integration incorporates these enriched images and their metadata into larger datasets. These datasets provide a robust training environment that exposes machine learning models to a wide array of visual scenarios, significantly improving their real-world applicability. This integration process is particularly emphasized in studies like those by \cite{li2024integrated}, where augmented datasets help enhance diabetes management. Each of these steps, depicted in Figure \ref{fig:ImageLLMResults} (right side as steps), highlights the technical process of LLMs in image data augmentation for advancing AI capabilities in various domains. For example, the integration of LLMs in medical image analysis by \cite{li2024integrated} and in agricultural applications by \cite{li2024itimca} showcases the transformative impact of these advanced data augmentation methods, enhancing both the diversity and the quality of datasets for complex visual recognition tasks.

In the domain of medical diagnostics, LLM-based image data augmentation has shown remarkable utility, particularly in enhancing the accuracy and efficiency of disease detection and treatment monitoring. For example, the Med-MLLM by Liu et al. \cite{liu2023medical} leverages unlabelled data for radiograph representation learning, significantly boosting diagnostic capabilities in visual and textual analyses. Similarly, the DeepDR-LLM model introduced by Li et al. \cite{li2024integrated} incorporates advanced lesion segmentation and diabetic retinopathy grading techniques to improve primary diabetes care. Additionally, MISTRA by Jindal et al. \cite{jindal2024mistra} utilizes variational autoencoders combined with multimodal fusion techniques to detect and classify medical images with high precision, demonstrating the potential of LLMs to revolutionize medical imaging and diagnostics. Table \ref{tab:surveyImageAug} provides a detailed overview of multimodal LLMs in image data augmentation from th erecent published publications, illustrating the diverse applications across different domains. Additionally, similar detailed information on LLM-based image data augmentation, focusing on the innovations and outcomes from recent peer-reviewed work, is systematically tabulated in Table \ref{tab:surveyTableImagePreprint}(Refer to Appendix section), further enriching the understanding of how these advanced models are applied in practice.

\begin{table*}[ht]
\centering
\caption{Survey of Multi-modal LLMs in Image Data Augmentation (Peer Reviewed Papers)}
\label{tab:surveyImageAug}
\scriptsize  
\renewcommand{\arraystretch}{1.3} 
\begin{tabular}{|p{2cm}|p{2cm}|p{4.5cm}|p{5cm}|}
\hline
\textbf{LLM Name} & \textbf{Domain} & \textbf{Data Augmentation Method} & \textbf{Key Insights} \\ \hline
DALL-E \cite{sapkota2024synthetic, sapkota2024zero, sapkota2025improved, sapkota2024comprehensive} & Agriculture & Synthetic image generation & Enhanced apple detection with YOLO models; challenges in differences arising from generated images with real-world conditions. \\ \hline
LLM-PTM \cite{yuan2023large} & Healthcare & Text generation for patient-trial matching & Improved trial-patient matching accuracy; privacy concerns with sensitive healthcare data. \\ \hline
ITIMCA \cite{li2024itimca} & Agriculture & Image-text fusion, contrastive learning & Improved cassava disease detection under limited data; dependent on high-quality multimodal inputs. \\ \hline
LLM SAM \cite{liu2024enhanced} & Healthcare & Dual contrast learning, mask generation & Enhanced breast cancer diagnosis; reliant on high-quality histopathological images. \\ \hline
Generative Models \cite{kirilenko2024generative} & Robotics & Neural networks for pathfinding & Outperformed traditional pathfinding methods; requires comprehensive training data. \\ \hline
MISTRA \cite{jindal2024mistra} & Social Media & Text-image fusion, multimodal detection & Detected misogynous memes with high Macro-F1 scores; challenges in aligning multimodal data inputs. \\ \hline
DeepDR-LLM \cite{li2024integrated} & Healthcare & DR screening with transformers & Improved diabetes care and DR screening in low-resource settings; integration with clinical workflows is complex. \\ \hline
Med-MLLM \cite{liu2023medical} & Healthcare & Multimodal representation learning & Enhanced clinical decision-making with unlabeled radiograph data; scalable for pandemics but limited by rare disease datasets. \\ \hline
Kartezio \cite{cortacero2023evolutionary} & Biomedical Imaging & Genetic programming for segmentation & Achieved precision with fewer training instances; complements deep learning but struggles with broader imaging tasks. \\ \hline
ViGPT2 \cite{raminedi2024multi} & Healthcare & Vision Transformer, GPT-2 integration & Improved medical image analysis and reporting; dependent on high-quality images and extensive computational resources. \\ \hline
MLLM4Rec \cite{wang2024mllm4rec} & Recommendation Systems & Multimodal learning with prompts & Boosted recommendation accuracy using image and audio data; computationally intensive fine-tuning. \\ \hline
DALLMi \cite{bețianu2024dallmi} & NLP & Semi-supervised domain adaptation & Addressed domain shifts in multi-label classification; limited by label imbalance and computational demands. \\ \hline
CNN \& GPT-3 \cite{sheik2024neural} & Legal & Pseudo-label generation & Improved legal overruling predictions; reliant on pseudo-label quality and computational resources. \\ \hline
\end{tabular}
\end{table*}

\vspace{0.5 cm}
\subsubsection{Methods and Techniques}
Figure \ref{fig:ImageLLMResults} in the middle section comprehensively outlines fifteen diverse techniques of LLM-based image data augmentation identified in our survey, each leveraging the unique capabilities of LLMs to enhance, manipulate, and generate image data, as detailed below
\begin{itemize}
    \item \textbf{Image-to-text Synthesis :} Recent advancements in image-to-text synthesis have harnessed the capabilities of LLMs to bridge the gap between visual content and textual descriptions, thereby enhancing the contextual richness of generated images. For instance, the DF-GAN model, introduced by \cite{tao2022df}, simplifies the generative process through a streamlined one-stage architecture, focusing on improving text-image semantic consistency. This is achieved through innovative components such as a Target-Aware Discriminator and a deep text-image fusion block, which ensure that generated images closely align with their textual descriptions, offering a marked improvement over traditional multi-stage generative adversarial networks. Similarly, \cite{liang2024rich} employs rich human feedback to refine image generation in response to textual prompts. This method leverages human annotations to identify and correct areas where generated images do not accurately reflect the text, using these insights to train a multimodal transformer that adjusts the generative process, thus enhancing the fidelity and relevance of the output. Additionally, \cite{phung2024grounded} tackles the challenges posed by diffusion models in maintaining fidelity to complex text prompts, by introducing novel loss functions that guide the image synthesis process more precisely, ensuring that the generated visuals are not only high in quality but also contextually appropriate. 

    \item \textbf{Semantic Content Transfer :} Semantic Content Transfer is another pivotal technique in LLM-based image data augmentation, effectively infusing semantically relevant content into images to boost the robustness and generalizability of machine learning models. The SemAug method, discussed by \cite{heisler2022semaug}, exemplifies this approach by dynamically incorporating contextually appropriate objects into existing images, which enhances object recognition models without the need for extensive contextual analysis. This method significantly improves model generalization capabilities as demonstrated on standard benchmarks like Pascal VOC and COCO. In a similar vein, DALDA, introduced by \cite{jung2024dalda}, integrates semantic information through a sophisticated fusion of LLMs and Diffusion Models (DMs), utilizing enhanced text prompts and adjusted guidance weights to maintain semantic integrity while increasing dataset diversity. This method is particularly effective in scenarios with limited data, demonstrating its utility in enhancing image diversity while preserving semantic accuracy. Furthermore, \cite{yu2023scaling} explores the innovative use of Semantic Imagined Experience in robotics, where text-to-image diffusion models are employed to project robots into varied scenarios through augmented visual data, significantly enhancing their operational adaptability in real-world tasks. 

    \item \textbf{Image Captioning :} Image Captioning for Augmentation represents a significant stride in utilizing LLMs to generate detailed, contextually aligned image captions. The FuseCap technique by \cite{rotstein2024fusecap} innovatively combines machine vision insights with LLM outputs to generate enriched captions that capture overlooked details, thereby improving the training accuracy of visual-language models. This method enhances the detail and accuracy of captions, proving particularly effective in complex image-caption tasks. Additionally, \cite{yi2024augment} introduces a novel approach that preserves semantic consistency between image-caption pairs through text-conditioned image modifications and advanced data augmentation techniques, such as pixel-level masking. This method ensures that the augmented pairs maintain their semantic linkage, thereby improving the efficacy of grounding-based vision and language models. The benefits of these advancements are further underscored in the study by \cite{sharifzadeh2024synth}, where image embeddings are synthesized from captions using LLMs and image generation models, showcasing a novel pathway to enriching training datasets without relying heavily on human-labeled data.

    \item \textbf{Conditional Image Generation :} Conditional Image Generation using LLMs has significantly advanced, allowing for the generation of contextually rich and detailed images to enhance the capabilities of generative models. For example, Koh et al. \cite{koh2024generating} introduced a method that combines pre-trained image encoders and decoders with text-only LLMs by mapping their embedding spaces, enabling the generation of images conditioned on complex and interleaved text and image inputs. This innovation surpasses traditional models in tasks requiring detailed language understanding. Another noteworthy contribution is from Li et al. \cite{li2024unimo} with UNIMO-G, a multimodal conditional diffusion framework that excels in generating images from both textual and visual prompts. This framework distinguishes itself by using a Multimodal LLM that encodes mixed inputs and a denoising diffusion process tailored to these complex inputs, demonstrating effectiveness in generating detailed images that purely text-driven models often overlook. Furthermore, Jung et al. \cite{jung2024dalda} discussed DALDA, a framework that combines LLMs with diffusion models to enhance data augmentation under data-scarce conditions, ensuring the generation of semantically consistent images that remain relevant to the training needs. These studies collectively underscore the significant progress in conditional image generation, offering robust solutions for diverse and challenging scenarios. 
    
    \item \textbf{ Contextual Object Insertion :} Contextual Object Insertion in image data augmentation has seen innovative applications through LLMs, enhancing object detection and generalization capabilities of models. SemAug, introduced by Heisler et al. \cite{heisler2022semaug}, calculates and places new, contextually relevant objects into images, improving model generalization without the need for a context network and demonstrating significant mAP improvements on benchmarks like Pascal VOC and COCO. Similarly, DALDA by Jung et al. \cite{jung2024dalda} integrates LLMs with Diffusion Models to embed novel semantic information into text prompts, maintaining target distribution fidelity while enhancing image diversity. This method proves particularly effective in data-scarce scenarios. Moreover, Yu et al. \cite{yu2023scaling} explore Semantic Imagined Experience in robot learning, using text-to-image diffusion models for data augmentation to robustly perform manipulation tasks in novel scenarios

    \item \textbf{Interactive Editing :} Interactive Editing with LLMs in image augmentation represents a burgeoning field that combines user input with advanced AI techniques to refine and personalize image outputs. The Visual Editing GPT 3.5 by Sultan et al. \cite{sultan2024visual} utilizes a distillation approach with data augmentation to improve fine-tuning in low-data regimes by 25\%, applied effectively in real-time visual editing tasks like color grading. This method demonstrates cost and latency reduction while maintaining high performance. Additionally, ForgeryGPT by Li et al. \cite{li2024forgerygpt} introduces a novel multimodal LLM for image forgery detection and localization, integrating high-order forensics knowledge with explainable AI capabilities. This approach develops innovative training strategies and architecture enhancements, such as the Mask-Aware Forgery Extractor for precise tampering detection and localization, advancing the field towards robust, explainable image forgery analysis.

    \item \textbf{Scene Completion :} Scene completion using LLMs has become a focal area in enhancing the contextual understanding of images through the seamless integration of missing or incomplete elements. For instance, the Image Augmentation Agent (IAA) developed by Wu et al. \cite{wu2024image} uses LLMs and diffusion models to generate high-quality, diverse training images, particularly for semantic segmentation, which includes filling in missing parts of scenes to improve dataset comprehensiveness and quality. This method has shown significant improvements on standard datasets like PASCAL VOC 2012 and MS COCO 2014. Furthermore, the LaB-RAG system by Song et al. \cite{song2024lab} enhances radiology report generation by integrating image-derived labels with retrieval-augmented generation, effectively completing scenes with medically relevant data without the need for deep learning model retraining. 

    \item \textbf{Automated Cropping :} Automated cropping leverages LLMs to intelligently crop images, focusing on enhancing the salient parts of an image while maintaining its aesthetic and informational value. This technique is particularly useful in applications where the key features within images vary widely in their placement. For instance, DIAGen introduced by Lingenberg et al. \cite{lingenberg2024diagen} employs Gaussian noise and class-specific text prompts from LLMs to refine the focus of images, ensuring that the most relevant features are emphasized and less important backgrounds are minimized. Similarly, the T2Vid system by Yin et al. \cite{yin2024t2vid} applies synthesized video-like samples to improve video understanding, which includes strategic cropping of frames to enhance the model’s focus on relevant actions or objects within a dynamic scene. 

    \item \textbf{Attribute-Based Generation :} Attribute-based generation using LLMs focuses on enhancing images by modifying specific attributes or incorporating new ones, thereby increasing the diversity and specificity of image datasets. This technique is crucial for applications requiring high customization or specific attribute alterations. For example, the work by DIAGen \cite{lingenberg2024diagen} not only focuses on diversifying image presentations but also specifically enhances attributes based on class-specific demands, improving classification performance and handling of out-of-distribution samples. Additionally, ForgeryGPT by Li et al. \cite{li2024forgerygpt} uses multimodal LLMs to detect and localize image forgeries by adjusting image attributes to match forensic profiles, enhancing the detection capabilities. These methods showcase the potential of LLMs to not only generate visually appealing images but also to tailor them to fit specific training or operational need. 

    \item \textbf{Image Enhancement :} Image Enhancement through LLMs involves refining image quality and detail to improve the performance of machine learning models, especially in fields like medical imaging and autonomous driving. For instance, Med-MLLM by Liu et al. \cite{liu2023medical} develops a multimodal large language model that enhances radiograph representations, aiding in more accurate disease reporting and diagnosis. Similarly, the DeepDR-LLM model by Li et al. \cite{li2024integrated} integrates LLMs with deep learning for diabetes management by enhancing lesion segmentation and diabetic retinopathy grading. This method significantly improves the clarity and diagnostic utility of medical images in low-resource settings. Additionally, the MISTRA framework by Jindal et al. \cite{jindal2024mistra} employs variational autoencoders alongside CLIP and BLIP models for image enhancement, particularly improving the detection and classification of misogynous memes by refining visual cues and text-image congruence.

    \item \textbf{Adversarial Examples Generation :} Adversarial Examples Generation using LLMs serves as a robust method to test and improve the resilience of AI models against potential exploits. This method involves creating images that are visually similar to original images but have been subtly altered to deceive AI models. For example, T2Vid by Yin et al. \cite{yin2024t2vid} explores the generation of synthesized video-like samples that include adversarial examples to enhance video understanding and model robustness against attacks. Similarly, the DIAGen approach by Lingenberg et al. \cite{lingenberg2024diagen} uses Gaussian noise and targeted attribute alterations to produce images that challenge the model's classification abilities, thereby improving its resistance to adversarial attacks.

    \item \textbf{Multimodal Data Fusion :}Multimodal Data Fusion in image data augmentation involves integrating information from various modalities, such as text, audio, and images, to create richer and more informative training datasets. This approach is exemplified by the MISTRA system by Jindal et al. \cite{jindal2024mistra}, which fuses image and text data to improve the accuracy of misogynous meme detection. Additionally, the MM-Instruct by Liu et al. \cite{liu2024mm} leverages LLMs to generate diverse visual instruction data from image captioning datasets, enhancing the instructional capabilities of large multimodal models. Furthermore, Kartezio by Cortacero et al. \cite{cortacero2023evolutionary} introduces a Cartesian Genetic Programming-based strategy for generating interpretable image processing pipelines that effectively integrate multimodal data for biomedical image segmentation.

    \item \textbf{Semantic Segmentation Enhancement :} Semantic Segmentation Enhancement using LLMs focuses on refining the accuracy with which models delineate and categorize different regions of an image, crucial for applications such as autonomous driving and medical imaging. The Image Augmentation Agent (IAA) by Wu et al. \cite{wu2024image} exemplifies this by utilizing LLMs and diffusion models to generate diverse training images, particularly enhancing the model’s ability to perform semantic segmentation under weak supervision. This approach notably improves segmentation accuracy on benchmark datasets like PASCAL VOC 2012 and MS COCO 2014 by introducing high-quality, varied training examples that help models better understand complex scene compositions. Additionally, the LaB-RAG system by Song et al. \cite{song2024lab} integrates image-derived labels with retrieval-augmented generation to improve radiology report generation, indirectly enhancing semantic segmentation by providing richer context and more detailed labels for medical imaging.

    \item \textbf{Synthetic Defect Introduction :} Synthetic Defect Introduction through LLMs is a method aimed at creating images with intentionally introduced defects to train models for quality control and defect detection tasks. This method is particularly valuable in manufacturing and quality assurance where detecting subtle defects can be crucial. For instance, DIAGen by Lingenberg et al. \cite{lingenberg2024diagen} uses Gaussian noise and class-specific text prompts to simulate defects in images that help models learn to identify and categorize these defects effectively. Another example is ForgeryGPT by Li et al. \cite{li2024forgerygpt}, which not only detects but also localizes image forgeries by introducing synthetic tampering in a controlled manner. This method enhances the capability of models to discern and react to complex forgery patterns, which are akin to defects in digital media.

    \item \textbf{Temporal Context Modeling :} Temporal Context Modeling in image data augmentation leverages LLMs to understand and incorporate the temporal dynamics within image sequences, which is essential for tasks such as video analysis and activity recognition. The T2Vid method by Yin et al. \cite{yin2024t2vid} enhances video understanding by using synthesized video-like samples that model temporal contexts, thus allowing models to better predict and interpret sequences of actions or events. This method reduces reliance on extensive real video datasets by providing rich, synthesized alternatives that capture essential temporal variations. Additionally, the DALL-M system by Hsieh et al. \cite{hsieh2024dall} uses LLMs to generate synthetic clinical data that includes temporal progressions of medical conditions, improving the predictive capabilities of models in medical diagnostics by understanding disease evolution over time.

\end{itemize}

\subsubsection{Limitations and Potential Solutions}
The integration of multimodal LLMs in image augmentation introduces several challenges and limitations, which are itemized below and illustrated on the rightmost side of Figure \ref{fig:ImageLLMResults}:

\begin{itemize}
    \item \textbf{Ambiguity and Unrealistic Outputs:} LLM-based image augmentation can face issues with ambiguity and unrealistic outputs due to several underlying scientific reasons. Primarily, LLMs rely heavily on textual prompts for generating images, and if these prompts lack specificity, the resulting images may not capture essential details, leading to generalized or contextually inaccurate representations \cite{tao2022df}. Additionally, the inherent limitations of LLMs in understanding complex visual semantics can further exacerbate this problem, as the models might not fully grasp subtle dynamics required for accurate visual depiction \cite{liang2024rich}. Consequently, this can lead to the generation of images that, while plausible at a superficial level, fail to accurately reflect detailed or scenario-specific characteristics, reducing their applicability in tasks requiring high fidelity to real-world contexts.

    To address ambiguity and unrealistic outputs in LLM-based image data augmentation, future solutions could involve enhancing textual prompts with more detailed and context-specific descriptions. Implementing a multi-modal training approach, where LLMs are trained not just on text but also on richly annotated visual datasets, could improve the models' understanding of complex visual contexts \cite{kumar2024image,fayaz2024advancements}. Additionally, integrating feedback loops where outputs are evaluated and corrected by humans could refine the models' generative capabilities. Advanced algorithms for semantic parsing could also be employed to better interpret and execute nuanced textual prompts, ensuring that the generated images maintain high fidelity to specified details.

    \item \textbf{Contextual and Semantic Misalignment:} Contextual and semantic misalignment presents a significant challenge in LLM-based image augmentation due to the complexity inherent in accurately interpreting and integrating contextual cues within images \cite{rotstein2024fusecap}. LLMs, despite their advanced capabilities, can falter in understanding the intricate relationships and subtleties that define a coherent visual context. This misalignment often results in object placements and enhancements that do not logically fit within the existing scene structure, making them appear out of context or blatantly irrelevant \cite{ hsieh2024dall}. Such inaccuracies can critically undermine the realism and practical utility of the augmented images, rendering them less effective or even unusable for tasks that require high levels of contextual accuracy, such as in training datasets for AI-driven visual recognition systems. The lack of precise semantic understanding in LLMs thus poses a barrier to creating believable and contextually appropriate visual content, which is essential for applications across various domains that rely on visual data \cite{zang2024contextual}.

    Addressing the limitation of contextual and semantic misalignment in LLM-based image augmentation can be approached by enhancing the contextual awareness and semantic understanding of LLMs. One scientific method involves training LLMs on a broader and more diverse set of context-rich image-text pairs, which can help the models learn more dynamic interpretations of visual contexts and their corresponding textual descriptions \cite{fayaz2024advancements, mumuni2024survey}. Additionally, integrating attention mechanisms can enable LLMs to focus on relevant parts of an image in relation to textual cues, improving alignment accuracy. Employing advanced techniques such as contrastive learning could also refine the models’ ability to distinguish between contextually appropriate and inappropriate augmentations.
    
    \item \textbf{User Dependency:} User dependency poses a significant limitation in the current state of image augmentation using multimodal LLMs due to the crucial role that the quality of input prompts and training data plays in determining the effectiveness of the augmentation process. When these inputs are inconsistent or of poor quality, they can result in augmentations that vary widely in quality, thereby adversely affecting the performance and generalizability of the models \cite{hsieh2024dall}. This dependence on user-provided data means that LLM-based augmentation systems are only as good as the information they are given. Substandard or contextually inaccurate prompts may lead LLMs to generate irrelevant or misleading outputs, which not only diminish the utility of the augmented images but also hinder the training of robust AI systems capable of operating effectively in diverse real-world scenarios \cite{zang2024contextual}.

    To mitigate the limitation of user dependency in LLM-based image augmentation, it is essential to enhance the robustness and contextual understanding of LLMs. One scientific approach is to implement adaptive learning algorithms that can refine their performance over time based on feedback loops. These feedback mechanisms allow LLMs to learn from the outcomes of their augmentations and adjust future outputs accordingly, reducing reliance on initial input quality \cite{kumar2024image}. Additionally, employing sophisticated preprocessing techniques to standardize and enrich input data can help normalize input quality variations \cite{mumuni2024survey, hsieh2024dall}. Advanced natural language understanding (NLU) capabilities can be integrated to better interpret and clarify ambiguous or poorly defined prompts. Finally, augmenting training datasets with a wider range of high-quality, annotated examples can teach LLMs to generate more accurate and contextually appropriate augmentations, even from suboptimal inputs.

    \item \textbf{Over-Specialization and Action Complexity:} Over-specialization and action complexity in LLM-based image augmentation arise primarily from the substantial computational demands associated with processing large datasets or complex multimodal inputs. LLMs, by design, integrate and analyze vast amounts of data to generate high-quality augmentations. This process often involves deep neural networks and sophisticated algorithms that are computationally intensive and energy-consuming \cite{zang2024contextual}. For applications that necessitate real-time processing, such as interactive media or live surveillance analysis, the latency introduced by these computational requirements can render LLM-based solutions impractical. Moreover, the complexity of actions, which refers to the multitude of steps and computations the LLM must perform to produce a single output, further complicates deployment in constrained or low-resource environments. This complexity not only affects scalability but also limits the potential for broader application of LLM technologies in fields where immediate response and agility are crucial \cite{zang2024contextual,hsieh2024dall}.

    To overcome the limitation of over-specialization and action complexity in LLM-based image augmentation, it is crucial to optimize the computational efficiency of these models. One approach involves refining model architecture to reduce complexity without compromising performance, such as using lighter neural network layers or pruning redundant parameters . Implementing more efficient algorithms for processing, like quantization or knowledge distillation, can also decrease computational load, enabling faster processing with fewer resources. Additionally, leveraging edge computing can distribute the computational tasks closer to the data source, reducing latency for real-time applications. Furthe, adopting adaptive computation techniques, which dynamically adjust the processing power based on the complexity of the task, can ensure optimal resource utilization and enhance scalability.
    
    \item \textbf{Model Integrity and Alignment Issues:} Model integrity and alignment issues present significant limitations in LLM-based image data augmentation due to the challenges of integrating these advanced models seamlessly with existing technological frameworks. LLMs are typically designed to generate or manipulate data based on very complex and often domain-specific training. When integrated into broader systems that were not originally designed with these models in mind, discrepancies can arise between the output expectations and the actual performance of the LLMs \cite{zang2024contextual}. Ensuring that the augmented outputs align with the specific requirements of practical applications involves not only technical adjustments to harmonize the interface between the LLMs and existing systems but also continuous calibration to maintain the fidelity of the outputs. The complexity of these tasks often requires substantial computational resources and sophisticated algorithms to manage the dynamic nature of LLM behaviors, making integration both technically demanding and resource-intensive. This need for high-level expertise and significant computational investment to ensure alignment complicates the adoption of LLMs in diverse operational environments. 

    Overcoming model integrity and alignment issues in LLM-based image data augmentation can be achieved through several scientific approaches. Firstly, employing modular integration strategies can allow for more flexible adaptation of LLMs within existing frameworks. This involves designing interfaces that can dynamically adjust to the outputs of LLMs, ensuring smoother integration and alignment with application-specific requirements. Secondly, continuous training and fine-tuning of the models on domain-specific datasets can enhance their understanding and performance, thereby improving the consistency and relevance of their outputs. Thirdly, implementing rigorous validation and testing protocols that simulate real-world scenarios can help identify and correct misalignments before full deployment. Finally, leveraging advanced machine learning techniques such as transfer learning can aid in adapting pre-trained models to new tasks or environments more effectively, enhancing their generalizability and utility across various applications.

    \item \textbf{Overfitting and Prediction Challenges:} Overfitting and prediction challenges are significant limitations in LLM-based image data augmentation, largely due to the models' propensity to reinforce existing biases and learn specific patterns too well. LLMs, when trained on datasets that lack diversity or fail to capture the full complexity of the target domain, might generate augmented data that is too narrowly tailored to the specific examples they have been exposed to. This results in a model that performs well on training data but poorly generalizes to new, unseen data. Scientifically, this issue arises because LLMs, particularly those with vast parameter spaces, are highly capable of memorizing rather than generalizing from input data. The absence of diverse and comprehensive training examples restricts the model's ability to develop robust predictive capabilities, leading it to make decisions based on limited or skewed perspectives, thereby significantly undermining its effectiveness in real-world applications \cite{hsieh2024dall}.

    To address overfitting and prediction challenges in LLM-based image data augmentation, several scientific strategies can be employed. Firstly, enhancing the diversity of the training dataset is crucial. Incorporating a wide range of images from various contexts and conditions can provide a broader base of examples for the LLM to learn from, promoting better generalization to new data. Secondly, implementing regularization techniques such as dropout, L2 regularization, or early stopping during the training process can prevent the model from fitting too closely to the training data. Additionally, employing cross-validation methods can help in assessing how the model will perform on independent data sets. Furthermore, using ensemble methods, which combine multiple models to make predictions, can reduce variance and improve the robustness of the predictions, further mitigating overfitting issues.

\end{itemize}

\begin{figure*}[htbp]
    \centering
    \includegraphics[width= 0.98\textwidth]{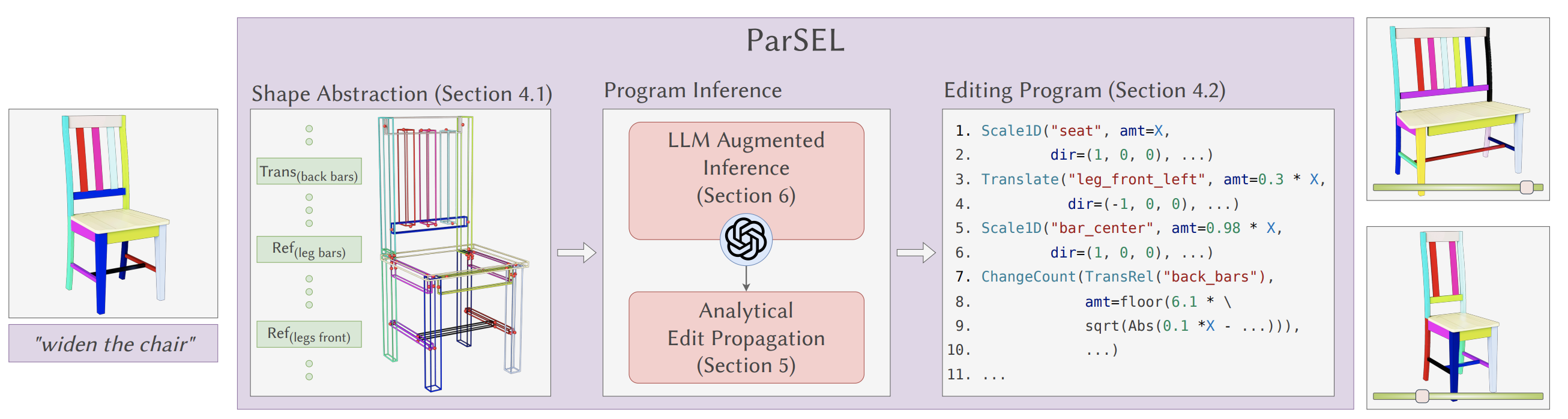}
    \caption{Conceptual example of LLM-based 3D augmentation. Given a base object, a text prompt could specify modifications like scaling. Adopted from \cite{ganeshan2024parselparameterizedshapeediting} }
    \label{fig:pointcloud_aug_llm}
\end{figure*}

\textbf{Point Cloud Data Augmentation}
While point cloud augmentation has traditionally relied on geometric, noise-based, or generative methods in ML/DL frameworks, an emerging research direction involves leveraging LLMs for more context-driven 3D shape manipulation as presented in \ref{fig:pointcloud_aug_llm}. These new approaches typically combine text-driven prompts or semantic information with 3D generation pipelines, offering a richer way to produce and edit point clouds.

\begin{itemize}
    \item \textbf{Moving Beyond Traditional and ML/DL Methods:} Traditional data augmentation strategies focus on generic geometric transformations such as rotation, scaling, translation, and flipping~\cite{qi2017pointnet2,wang2019dynamic}, random dropout or noise injection~\cite{achlioptas2018learning}, and region-level editing~\cite{chen2021deep,fan2021sgpr}. More advanced deep learning methods use generative models like 3D-GANs~\cite{wu2016learning}, VAEs~\cite{valsesia2020learning}, or diffusion models~\cite{zhou20213dgenerative,luo2023scorebased} to produce synthetic shapes. These approaches have proven effective, but typically require handcrafted rules or high-level neural networks that do not incorporate natural language feedback. Consequently, the range of possible edits remains constrained by predefined transformations or learned latent spaces.
    \item \textbf{Text-Guided Shape Editing and Augmentation:} One of the most direct ways LLMs contribute to point cloud augmentation is via text-guided shape editing, where a user or system provides natural language instructions that drive the modifications. Kim et al.~\cite{kim2022text2mesh} introduced a text-to-mesh stylization framework that uses CLIP - an early precursor to multimodal LLMs to align a given 3D surface with text prompts (for example, “make it look like coral”). Although focused on meshes, similar text-driven approaches for point clouds have begun to appear. These systems typically rely on LLM-based encoders to parse text instructions and then invoke a generative or editing module to reshape the geometry accordingly, effectively bridging language and 3D shape manipulation.  
    \item \textbf{ Point-E and Shap-E: Diffusion Models with LLM Guidance:} OpenAI’s Point-E~\cite{nichol2022pointe} and Shap-E~\cite{jun2023shape} highlight a diffusion-like pipeline for generating 3D point clouds or meshes from textual descriptions. These models often use large text encoders (e.g., from GPT or CLIP families) to interpret prompts such as “a chair with curved legs and a tall backrest,” then sample a 3D object consistent with that description. While these systems primarily aim to synthesize 3D shapes from scratch, they also open the door to augmenting existing datasets by generating infinite shape variations with natural language refinements. In principle, one could prompt, “Generate 50 more chairs that have slightly thicker legs and smaller backrests,” thereby expanding the range of training samples for a 3D classifier or detector.
    \item \textbf{DreamFusion and Zero-1-to-3: Text-Conditioned Neural Fields:}  A parallel branch of work employs NeRF-like or implicit field representations for text-guided 3D reconstruction. DreamFusion~\cite{poole2022dreamfusion} and Zero-1-to-3~\cite{liu2023zero1to3} both incorporate LLM-based text encoders (often CLIP or a derivative multimodal module) and an optimization process that iteratively refines a neural field to match the text prompt. Although many of these methods yield volumetric or implicit representations, the resulting geometry can be sampled to produce point clouds. For data augmentation, researchers can systematically vary prompts (e.g., “a slightly longer couch,” “a couch with no legs,” “a couch with accent pillows”) to create a diverse set of shapes without manual 3D modeling.
    \item \textbf{Semantic Part Replacement Driven by LLMs:} Earlier methods for part swapping in point clouds relied on manual labels or specialized networks that recognized object parts (leg, seat, handle, etc.)~\cite{li2021shapepart,wu2021pq}. LLMs now offer a more semantic approach, where the model can parse descriptions (“replace the top of this mug with a small dome”) and translate them into structured part edits~\cite{su2023semanticpc}. The system might first map “top of a mug” to a region of a known shape, then retrieve the requested “small dome” geometry from another shape library, assembling them with appropriate scaling. This process can significantly enrich the object varieties seen during training. 
    \item \textbf{Context-Aware Augmentation for 3D Scenes:} Autonomous vehicles and robotics often work in 3D worlds with multiple objects and complex layouts. LLM-driven augmentation can add or modify objects in scene point clouds according to textual commands specifying the scenario. For instance, a user might say, “Add a parked car next to the curb,” and the system, guided by an LLM, would select a car model, position it near the curb, and align it with the coordinate system~\cite{wu2023det3d}. This approach yields more realistic, scenario-specific data, which helps domain adaptation and robustness. By leveraging an LLM’s language-based reasoning, one can generate many scene variations with minimal manual labeling.
    \item \textbf{Potential Challenges and Limitations :} Although LLM-based approaches promise unprecedented flexibility, they also face unique hurdles. Generating fully realistic and collision-free object placements can be challenging if the LLM’s world model is incomplete or if the generative modules are poorly trained. Text prompts can be ambiguous, leading to geometry that misaligns with the intended meaning, particularly if the system lacks strong 3D priors. Another issue involves the computational overhead of large-scale text-driven generation, as well as possible biases in the textual data used to train LLMs. If the model has not been exposed to certain geometric or stylistic concepts, it may struggle to produce them accurately.
\end{itemize}
 In the next few years, researchers are likely to develop more refined pipelines where a language model, a shape retrieval system, and a 3D generative model interact seamlessly. This could involve iterative “conversation-like” steps (e.g., “Make the table thinner,” “Shorten the legs,” “Move it closer to the wall”) akin to prompt engineering, ensuring each new shape version remains semantically valid. There is also a growing push toward multimodal pretraining that unifies 2D images, textual captions, and 3D geometry~\cite{zhao2023foundations3d}, enabling richer cross-domain data augmentation. Over time, these tools could democratize 3D dataset creation, allowing even non-experts to tailor large point cloud datasets for any target application.

\subsection{LLM-based Text Data Augmentation}
\subsubsection{Process Overview} The process of text data augmmentation using multimodal LLMs can be summarized into eight key steps as illustrated in Figure \ref{fig:ImageLLMResultsTextAug}(left side). It begins with text encoding, where raw text data is transformed into a machine-readable format through techniques like tokenization and embedding. Tokenization breaks down the text into manageable units, while embedding assigns these units into a high-dimensional space that reflects semantic relationships \cite{hu2024llm}. This encoded form serves as the foundation for the LLM to understand and interact with the text. Following encoding, prompt generation occurs where the LLM utilizes the encoded data to generate prompts that direct the augmentation. These prompts define the type of transformations needed, such as paraphrasing or stylistic changes, which guide the subsequent augmentation steps. Next, augmentation instruction generation takes the prompts to develop specific directives for altering the text \cite{wu2024improving}. These instructions dictate exact changes like synonym replacement or sentence rephrasing, ensuring the modifications align with the augmentation goals. The process then moves into natural language to task-specific transformations, where these instructions are interpreted and applied to the text, adapting it for specific tasks while maintaining linguistic accuracy \cite{hu2024llm}. This stage is crucial for ensuring the relevance and applicability of the transformations to the intended training objectives. 

Following this, text transformation execution is carried out, where the actual modifications are implemented on the text. This involves sophisticated linguistic manipulation techniques that adjust the text according to the predefined instructions, ensuring the new versions are both varied and contextually appropriate \cite{zhao2024improving}. The sixth step, quality assessment, evaluates the augmented texts against quality standards such as grammaticality, coherence, and task relevance. This evaluation often employs both automated metrics and manual reviews to ensure only high-quality augmentations are retained \cite{dos2024identifying}. Upon successful quality verification, dataset integration occurs, where the augmented texts are systematically compiled into the training dataset. Alongside, metadata generation provides a comprehensive annotation of the modifications applied, including the type and scope of transformations. This metadata is essential for tracking the augmentation impacts and refining future augmentation strategies, thereby enhancing the training datasets' utility and effectiveness in developing robust NLP models\cite{hu2024llm, dos2024identifying}.

\begin{table*}[ht]
\centering
\caption{Survey of Multi-modal LLMs in Text Data Augmentation (Peer Reviewed Papers)}
\label{tab:survey}
\scriptsize
\renewcommand{\arraystretch}{1.3} 
\begin{tabular}{|p{2cm}|p{2cm}|p{4.5cm}|p{4cm}|}
\hline
\textbf{LLM Name/Method \& Reference} & \textbf{Domain} & \textbf{Text Augmentation Method} & \textbf{Key Insights} \\ \hline
BERT-based TTEC \cite{hua2023multimodal} & Fake News Detection & Back-translation and contrastive learning & Enhances detection by 3.1\% on Mac. F1 scores; requires larger datasets. \\ \hline
RumorLLM \cite{lai2024rumorllm} & Fake News Detection & Rumor-specific writing style finetuning & Improves F1 score and AUC-ROC on BuzzFeed and PolitiFact; ethical concerns with textual bias. \\ \hline
LLM-Enhanced Personality Detection \cite{hu2024llm} & Personality Detection & Semantic, sentiment, and linguistic augmentations & Enhances performance but lacks continuous learning post-deployment. \\ \hline
LLM Mix-Up AAC \cite{wu2024improving} & Captioning & Caption mix-up augmentation with ChatGPT & Achieves 32.6 SPIDEr-FL score; high computational cost during training. \\ \hline
LLM-Based Civic Issues Detection \cite{dos2024identifying} & Social Media Analytics & Text diversity and imbalance correction & Identifies issues from tweets with 90.9\% accuracy; dependent on manual labeling. \\ \hline
ChatGPT-Based DA Enhancement \cite{zhao2024improving} & Text Classification & Rewriting and data generation with ChatGPT & Enhances classification but effectiveness plateaus beyond certain data size. \\ \hline
LA-UCL \cite{zhang2024ucl} & Few-shot Learning & Unsupervised contrastive learning with LLM augmentation & Exceeds baseline models but risks overfitting without diverse data. \\ \hline
LLM-Assisted DLP \cite{zhang2024llm} & Dependency Parsing & Multi-level augmentations (word, syntax, discourse) & Boosts Chinese dialogue parsing but struggles to generalize without diverse training data. \\ \hline
TnT-LLM \cite{wan2024tnt} & Text Mining & End-to-end label generation and assignment & Optimizes user intent detection but relies on consistent LLM performance. \\ \hline
LLM-Based Equity Enhancement \cite{cai2023resolving} & Reviewer Assignment & Hierarchical data augmentation & Improves reviewer assignment but may generate incorrect domain-specific text. \\ \hline
LLM-PTM \cite{yuan2023llm} & Healthcare & Privacy-aware data augmentation & Enhances patient-trial matching by 7.32\%; potential privacy risks. \\ \hline
LLM-PTM \cite{latif2024evaluation} & Healthcare & Privacy-aware data augmentation & Increases matching performance but limits direct data usage. \\ \hline
ASM-LMP \cite{ahmed2024automatic} & Code Summarization & Semantic augmentation for LLM prompts & Surpasses 30 BLEU1 on PHP but lacks generalization across all languages. \\ \hline
CEAN \cite{meng2024cean} & Event Extraction & Multi-pattern rephrasing with LLM & Achieves state-of-the-art results but faces data noise challenges. \\ \hline
LAMBADA \cite{ccataltacs2023comparison} & Sentiment Analysis & Sampling-based data augmentation & Improves robustness on SST-2 but limited by augmentation ratio. \\ \hline
LLM Oversampling \cite{cloutier2023fine} & Multiclass Classification & Generative LLM oversampling & Enhances performance on imbalanced classes but less effective for binary classification. \\ \hline
SK-TOD LLM \cite{jung2023enhancing} & Dialogue Systems & Knowledge-grounded data augmentation & Improves task-oriented systems but may not generalize across all subtasks. \\ \hline
Forged-GAN-BERT \cite{silva2024forged} & Authorship Attribution & Dual augmentation with GAN and Forged Novels Generator & Achieves high F1 scores (0.97, 0.71) but depends on data quality. \\ \hline
OphGLM \cite{deng2024ophglm} & Healthcare (Ophthalmology) & Multimodal data integration with FundusTuning-CN dataset & Superior performance in fundus disease classification but dependent on dataset quality and integration complexity. \\ \hline
\end{tabular}
\end{table*}

\subsubsection{Methods and Techniques}
Figure \ref{fig:ImageLLMResultsTextAug} in the middle section comprehensively outlines fifteen diverse techniques of LLM-based text data augmentation identified in our survey as detailed below:
\begin{figure*}[htbp]
    \centering
    \includegraphics[width=0.98\linewidth]{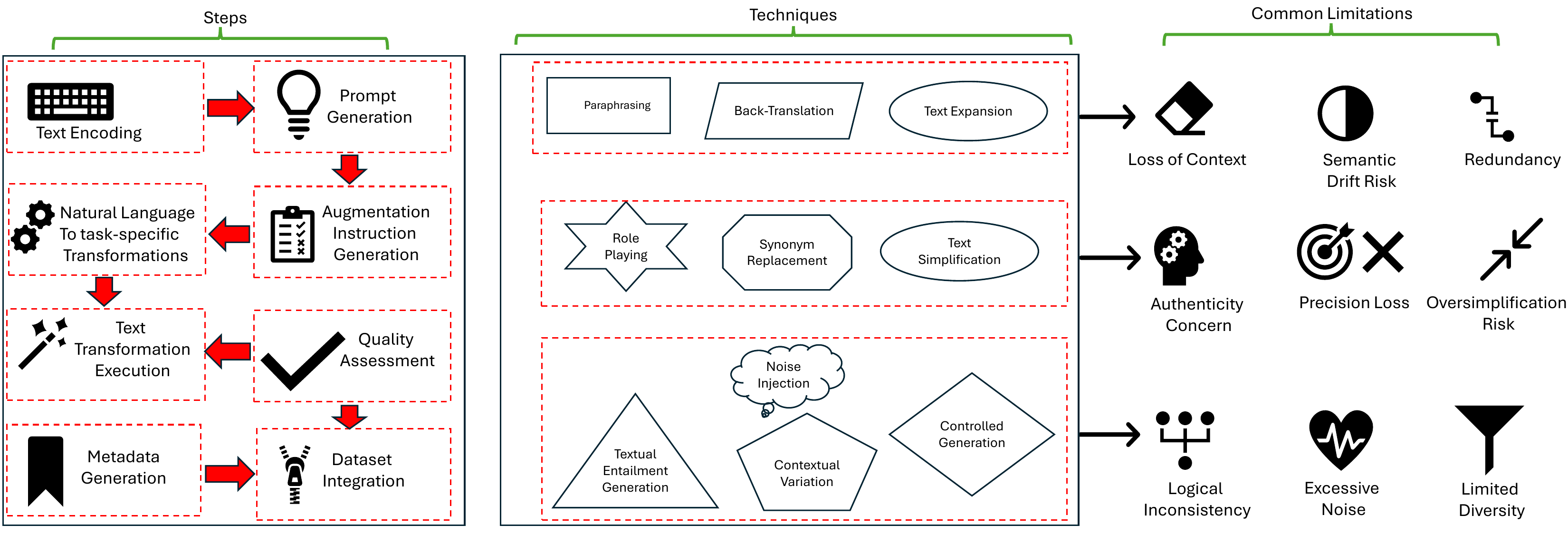}
    \caption{LLM based text data augmentation : showing the technical aspect of how text augmentation using LLM is performed, the techniques of augmenting text data using LLM and their limitations}
    \label{fig:ImageLLMResultsTextAug}
\end{figure*}
\begin{itemize}
    \item \textbf{Paraphrasing:}  Paraphrasing is a critical technique in text data augmentation where the semantics of the original text are preserved while the phrasing and structure are altered. This is achieved through sophisticated LLMs, which analyze and reconstruct text to enhance the richness and diversity of datasets. The method is particularly beneficial in applications requiring nuanced expression alterations without changing underlying meanings, such as personality detection \cite{hu2024llm} and text classification \cite{zhao2024improving}. For instance, by leveraging semantic and linguistic enrichments, LLMs effectively tailor text data to fit various contexts and styles, thereby improving the robustness and generalization of machine learning models. This approach not only aids in the training of NLP systems but also addresses the challenge of dataset scarcity and bias \cite{song2024lab}, \cite{glazkova2024evaluating}.

    \item \textbf{Back-Translation:}Back translation involves translating text to a different language and then translating it back to the original language, serving as a powerful method for text augmentation. This technique enriches the linguistic diversity and syntactic variety of text data, crucial for training more resilient NLP models. Studies like \cite{fischer2024swissadt} utilize back translation to enhance the accuracy of audio description translations in multilingual settings, ensuring the translations retain their original semantic richness across different languages. Moreover, this method supports the improvement of text classification systems by expanding the training dataset with varied syntactical structures, which helps in reducing model overfitting on specific language patterns \cite{dai2023auggpt}, \cite{lee2024llm2llm}.

    \item \textbf{Text Expansion:} Text expansion involves elaborating or extending existing text data to create new content that maintains the original context but adds additional descriptive detail. This technique is particularly useful in scenarios where detailed explanatory content enhances model training, such as in medical report generation \cite{song2024lab} or environmental informatics \cite{glazkova2024evaluating}. By using LLMs to systematically expand text data, researchers can generate more comprehensive training examples that simulate a wider range of real-world applications. This method not only improves the depth and detail of the datasets but also aids models in developing a better understanding of nuanced contexts and complex scenarios \cite{wen2024aidbench}, \cite{kang2024synthetic}. Additionally, text expansion has proven effective in increasing the robustness of models against overfitting by providing them with a broader array of expressions and contexts, thereby enhancing the generalization capabilities of NLP systems.

    \item \textbf{Role Playing:}Role playing in text data augmentation involves simulating different perspectives or personas using LLMs to generate diverse responses based on specified roles. This method effectively broadens the scope of conversational models and enhances their ability to handle varied interaction styles, crucial for applications like customer service bots or interactive AI agents. By adopting multiple character roles, LLMs such as those discussed in \cite{song2024lab} and \cite{glazkova2024evaluating} can produce more natural and contextually appropriate dialogues, significantly improving user engagement and satisfaction. Role playing also aids in training AI to understand and replicate human emotional nuances, thereby making interactions more realistic and empathetic \cite{kang2024synthetic}, \cite{dos2024identifying}.

    \item \textbf{Synonym Replacement:}Synonym replacement is a straightforward yet effective text augmentation technique where words or phrases within a text are replaced with their synonyms to introduce lexical diversity without altering the original meaning. This method is instrumental in training NLP models to recognize and process a broader range of lexical variations, enhancing their linguistic adaptability and understanding. Synonym replacement is particularly valuable in tasks like sentiment analysis and text classification, where the robustness of the model against different expressions of the same sentiment or topic is crucial \cite{hu2024llm}, \cite{zhao2024improving}. By integrating synonym replacement, studies such as \cite{cegin2024llms} and \cite{lee2024llm2llm} demonstrate improved model performance across various text-based tasks by effectively expanding the dataset's lexical field without compromising data quality or contextual relevance.

    \item \textbf{Text Simplification :}Text simplification involves reducing the linguistic complexity of text while maintaining its essential information and meaning, making it accessible to broader audiences, including those with limited language skills or cognitive impairments. This augmentation technique is crucial in educational technologies and readability enhancements, where complex information needs to be conveyed in simpler terms. Simplification can help in training models to generate more user-friendly content, which is especially important in medical or legal information dissemination \cite{song2024lab}, \cite{fischer2024swissadt}. Moreover, by simplifying text, LLMs such as those utilized in \cite{wen2024aidbench} and \cite{glazkova2024evaluating} can produce data that supports models in achieving better accuracy in understanding and processing simplified narratives, thus enhancing the overall effectiveness of AI systems in real-world applications.

    \item \textbf{Textual Entailment Generation :} Textual entailment generation involves using LLMs to create pairs of texts where one text (the premise) logically implies the other (the hypothesis). This form of augmentation is crucial for training models in natural language understanding tasks, such as question answering and information retrieval, where understanding the relationship between text segments can dramatically improve performance. Recent studies like those by \cite{wen2024aidbench} and \cite{lee2024llm2llm} have utilized LLMs to automatically generate large datasets of entailment pairs, which help in improving the inference capabilities of NLP models under various contexts. \cite{song2024lab} and \cite{kang2024synthetic} also highlighted the use of entailment generation to refine models' ability to process and understand nuanced human language, thus enhancing their applicability in real-world scenarios. 

    \item \textbf{Noise Injection :} Noise injection in text data augmentation involves intentionally adding errors or variations to text data, such as misspellings, grammatical errors, or shuffled word orders, to simulate real-world inaccuracies that models may encounter. This technique is crucial for building robust NLP systems that can effectively handle imperfect inputs. Studies by \cite{meng2024cean} and \cite{cloutier2023fine} have shown that training models with noised data can significantly improve their performance in tasks like speech recognition and optical character recognition where input data often contains errors. \cite{zhang2024ucl} and \cite{cai2023resolving} further demonstrate that noise injection helps in enhancing the model's resilience against overfitting and improving its ability to generalize from training to real-world application scenarios. By exposing models to a wider array of linguistic variations, noise injection ensures that NLP systems are not only accurate but also adaptable to the imperfect nature of human-generated text.

    \item \textbf{Contextual Variation :} Contextual variation in text augmentation leverages LLMs to modify a given text's context or to extend its narrative, thus enriching the dataset with diverse linguistic structures and themes. This method is particularly effective in enhancing the model's ability to understand and generate text that varies significantly in style, tone, or context. By integrating studies like \cite{dos2024identifying} and \cite{latif2024evaluation}, researchers can produce text variations that mimic different dialects, cultural nuances, or domain-specific jargon, broadening the training data's scope and depth. Additionally, works by \cite{hu2024llm} and \cite{zhao2024improving} have utilized contextual variations to improve the performance of models in tasks that require a deep understanding of context-specific language use, such as sentiment analysis and personalized content generation. This approach not only improves the linguistic versatility of the models but also their applicability across various domains and user groups.

    \item \textbf{Controlled Generation :} Controlled generation using LLMs focuses on generating text based on specific guidelines or constraints, such as maintaining a certain tone, style, or adhering to predefined content themes. This form of text data augmentation is essential for applications requiring high levels of precision and customization, like marketing content creation or legal document preparation. Controlled generation techniques have been explored in studies like \cite{fischer2024swissadt} and \cite{glazkova2024evaluating}, where LLMs are used to ensure that the generated text meets the specific needs of multilingual translation or tailored content creation. Further, \cite{song2024lab} and \cite{alyafeai2024arabic} demonstrate how controlled generation can be applied to create highly targeted training data that enhances the performance of NLP systems within specific functional parameters. This method not only ensures the relevance and applicability of the generated text but also enhances the effectiveness of the training process by aligning it closely with the end-use cases.
\end{itemize}

\subsubsection{Limitations and Potential Solutions}
\begin{itemize}
    \item \textbf{Loss of context:} Zhao et al. (2024) observed context loss in their study on improving text classification with LLM-based data augmentation \cite{zhao2024improving}. They found that when generating entirely new samples using ChatGPT, the model sometimes produced text that lacked the necessary context for accurate classification. This was particularly evident in domain-specific datasets, where the generated samples failed to capture the detai context of the original data. The loss of context in LLM-based text augmentation can be mitigated through the implementation of fine-tuning strategies that incorporate a mix of relevant and irrelevant contexts, as demonstrated by Yoran et al. (2024) who achieved robustness to irrelevant information with as few as 1,000 training examples \cite{yoran2023making}.

    To improve the loss of context in LLM-based text augmentation, it is essential to adopt strategies that enhance the model's ability to retain and understand relevant contexts. One effective approach is through fine-tuning the LLMs on domain-specific datasets. This method enables the models to learn the intricacies and typical patterns within specialized contexts, thereby generating text that maintains the depth and relevance of the original data. Additionally, implementing mixed-context training,  can further strengthen the model's robustness against irrelevant information \cite{lee2024unlocking}. By incorporating a controlled mix of relevant and irrelevant contexts into the training set, LLMs can learn to discern and prioritize information that is crucial for accurate text generation and classification.
    
    \item \textbf{Semantic Drift Risk} The issue of semantic drift was noted by Whitehouse et al. (2023) in their work on LLM-powered data augmentation for multilingual commonsense reasoning \cite{whitehouse2023llm}. They observed that when generating Tamil text using GPT-4, the model often inserted "uncommon and out-of-context words." For example, the generated text would include phrases that were semantically incorrect or inappropriate for the given context, demonstrating a drift from the intended meaning.  Semantic drift can be addressed by employing filtering methods, such as using natural language inference models to assess the relevance of generated content to the original task or domain \cite{yoran2023making, min2023recent}. 

    To minimize the risk of semantic drift in LLM-based text data augmentation for deep learning applications, it is essential to employ advanced filtering and training strategies. Integrating natural language inference (NLI) models can be highly effective. These models assess the logical and contextual relevance of the generated text, ensuring it aligns with the original input’s semantic boundaries. This acts as a critical check against the insertion of out-of-context or inappropriate content. Expanding and diversifying the training datasets is another crucial strategy. By incorporating a wide range of linguistic dynamics and diverse contexts, especially in multilingual settings, the LLM can develop a more robust understanding of language patterns. This extensive exposure helps the model better grasp and reproduce the contextual subtleties required for accurate text augmentation. Additionally, implementing adaptive training methods that include feedback loops to continuously refine the model’s outputs can further enhance accuracy. This approach allows for real-time adjustments to the model’s parameters, correcting deviations and aligning generated content more closely with desired semantic and contextual standards.

    \item \textbf{Redundancies and Authenticity Concern:} in LLM-generated text were also identified by Whitehouse et al. (2023) \cite{whitehouse2023llm}. They found that the augmented data often contained "redundant words with the same meaning." An example they provided was the phrase "I will retry to try it again," where the concepts of retrying and trying again are unnecessarily repeated, creating redundancy in the generated text. Redundancies in augmented text can be reduced through the application of model compression techniques like pruning and quantization, which optimize the model's efficiency without significant performance loss, thereby improving the quality and conciseness of generated content \cite{gupta2022compression, dantas2024comprehensive}.  LLM-based text augmentation, while powerful, faces significant authenticity concerns as a key limitation. Research by Silva et al. (2024) \cite{silva2024forged} and Wu et al. (2023) \cite{wu2025survey}  highlights this issue. Silva's team found that LLMs can produce convincingly human-like texts, creating challenges in authorship attribution, especially in literary contexts. They observed LLMs could generate novels falsely attributed to famous authors or mimic the style of well-known works. Wu's study revealed LLMs' tendency to fabricate information, rely on outdated data, and be overly sensitive to prompts, potentially spreading misinformation and undermining expertise. Both studies emphasize the difficulty in distinguishing LLM-generated content from human-authored text, with human detection methods proving unreliable

    Future research in LLM-based text augmentation could focus on developing advanced filtering mechanisms that leverage natural language inference models and semantic similarity metrics to ensure generated content maintains contextual relevance, semantic coherence, and authenticity, while simultaneously implementing robust verification systems that employ multi-modal analysis to detect and mitigate potential fabrications or misattributions in augmented text

    \item \textbf{Oversimplification Risk: } Oversimplification Risk in LLM-based text augmentation refers to the potential for generated data to lack the complexity and nuance of real-world examples, potentially leading to reduced model performance on more intricate tasks. This limitation was identified in a study published in August 2024 titled "LLMs vs Established Text Augmentation Techniques for Classification \cite{cegin2024llms}. The researchers found that while LLM-based augmentation can improve downstream classifier accuracy, it may not always outperform established methods significantly. They observed that LLM-generated samples might oversimplify the original text, especially when dealing with complex topics or specialized domains. This oversimplification can result in a loss of important contextual information or subtle features that are crucial for certain classification tasks. The study emphasized the need for careful consideration when using LLM-based augmentation, particularly in domains where detailed understanding is critical for accurate classification. Additionally, the risk of LLMs providing overly simplistic or reductionist assessments, especially in the context of economic policy is highlighted in a recent study \cite{atashbar2024reinforcement}. The authors warned that LLMs might generate advice that overlooks important contextual factors, unintended consequences, or distributional effects, potentially leading to suboptimal or harmful decisions if relied upon too heavily. 

    \item \textbf{Precision Loss and Logical Inconsistency:} LLMs often struggle to maintain the semantic integrity of the original text during augmentation. This can result in data that, while varied, might not always be semantically consistent with the source or contextually appropriate, affecting the quality and usability of the augmented data in training models for specific task\cite{ye2024llm}.The effectiveness of data augmentation heavily relies on the quality of the augmentation instructions provided to the LLM. Poorly defined or ambiguous instructions can lead to augmented data that does not meet the desired criteria or varies significantly in quality across different tasks. A study on ChatGPT-based data augmentation for text classification found that for labels with already sufficient training samples and high accuracy, adding augmented data may introduce noise and decrease performance \cite{zhao2024improving, sapkota2025comprehensive}. Likewise, logical inconsistency in LLM-based text data augmentation poses a significant limitation, impacting the utility and reliability of augmented datasets. Despite LLMs' remarkable capabilities in generating diverse and voluminous textual content, their tendency to produce logically inconsistent outputs when faced with complex input queries can severely undermine the quality of augmented text \cite{ghosh2024logical}. This inconsistency often manifests as a divergence in the logical coherence of the text relative to the original material, which is especially detrimental in tasks requiring precise logical structures, such as legal reasoning, academic research, and technical documentation\cite{huang2023survey, ghosh2024logical}. When employing LLMs for text augmentation, especially in sophisticated domains that require high fidelity to original content meaning and structure, the challenge lies in ensuring that augmented outputs maintain the same logical and factual integrity as their sources \cite{liu2024aligning}. The issue is exacerbated by LLMs’ susceptibility to generating responses that, while superficially plausible, may harbor subtle logical errors or misrepresentations known as "hallucinations" in AI parlance . Such errors can mislead downstream applications, lead to the propagation of inaccuracies, and reduce the overall effectiveness of models trained on these datasets. 

    To overcome precision loss and logical inconsistency in LLM-based text augmentation, it is essential to refine the quality control mechanisms and provide precise, context-specific augmentation guidelines. Enhancing training data with well-annotated, high-quality examples that demonstrate desired outcomes can help LLMs learn to generate more accurate and logically consistent augmentations. Implementing post-generation validation steps, such as semantic and logical coherence checks using rule-based systems or secondary models trained to identify and correct inconsistencies, can further mitigate errors. Additionally, fine-tuning LLMs on a domain-specific corpus before deployment can align their outputs more closely with the contextual and factual demands of specific tasks \cite{zhao2024improving, ghosh2024logical}.

    \item \textbf{Excessive Noise: } Excessive noise in text data augmentation using LLMs refers to the introduction of too much irrelevant or incorrect information during the augmentation process, potentially degrading the quality of the generated data. This can lead to reduced model performance and increased overfitting \cite{ye2024llm}. Recent studies have highlighted this limitation. Ye et al. (2024) demonstrated that excessive noise in LLM-generated augmented data can compromise the semantic integrity of the original text, leading to decreased name entity recognition (NER) model performance \cite{ye2024llm} . Similarly, Bolding et al. (2023) \cite{bolding2023ask}observed that while LLMs can effectively clean noisy translation data, they may introduce excessive noise when generating entirely new samples, necessitating careful balance in data augmentation strategies.

    To address excessive noise in text augmentation using LLMs, researchers have proposed implementing noise control mechanisms and leveraging LLMs' capabilities for targeted data cleaning \cite{yin2024leveraging}. Such approaches include using LLMs to select appropriate noise types for fine-tuning, employing curriculum learning to gradually increase data complexity, and utilizing LLMs to remove specific types of noise while preserving semantic integrity \cite{yin2024leveraging, bolding2023ask}. 

    \item \textbf{Limited Diversity}: Limited diversity in LLM-based text data augmentation refers to the challenge of generating sufficiently varied and unique text samples while maintaining semantic consistency with the original data \cite{ding2024data}. This can lead to reduced model generalization and reinforcement of existing biases, as the augmented datasets lack the breadth of linguistic expressions, structures, or content necessary to improve model performance effectively \cite{yu2024large}. Recent studies have extensively explored this limitation and proposed various strategies to address it, emphasizing the critical need to enhance diversity in LLM-generated text for data augmentation \cite{kim2025tardis}.    
    
    To address this, recent research has focused on developing strategies to increase the diversity in LLM-generated text. For instance, Cigen et al. (2024) \cite{cegin2024effects} explored how diversity incentives, such as the use of taboo words, chaining techniques, and hints, can enhance lexical diversity in generated paraphrases, ultimately improving downstream model performance. Likewise, future strategies should include diversifying the initial training data, implementing controlled augmentation techniques, and utilizing hybrid models that blend LLMs with adaptive learning methods. Fine-tuning on specialized datasets and applying regularization techniques can help generate more context-specific and varied outputs. Development of diversity-specific evaluation metrics and post-generation refinement processes are also vital. Additionally, monitoring and mitigating biases is crucial to ensure that the augmented data is not only linguistically diverse but also inclusively represents different perspectives, thereby enhancing model robustness and applicability across various domains.
\end{itemize}

\subsection{LLM-Based Speech Data Augmentation}      

\subsubsection{Process Overview}
\begin{figure*}[htbp]
    \centering
    \includegraphics[width=0.98\linewidth]{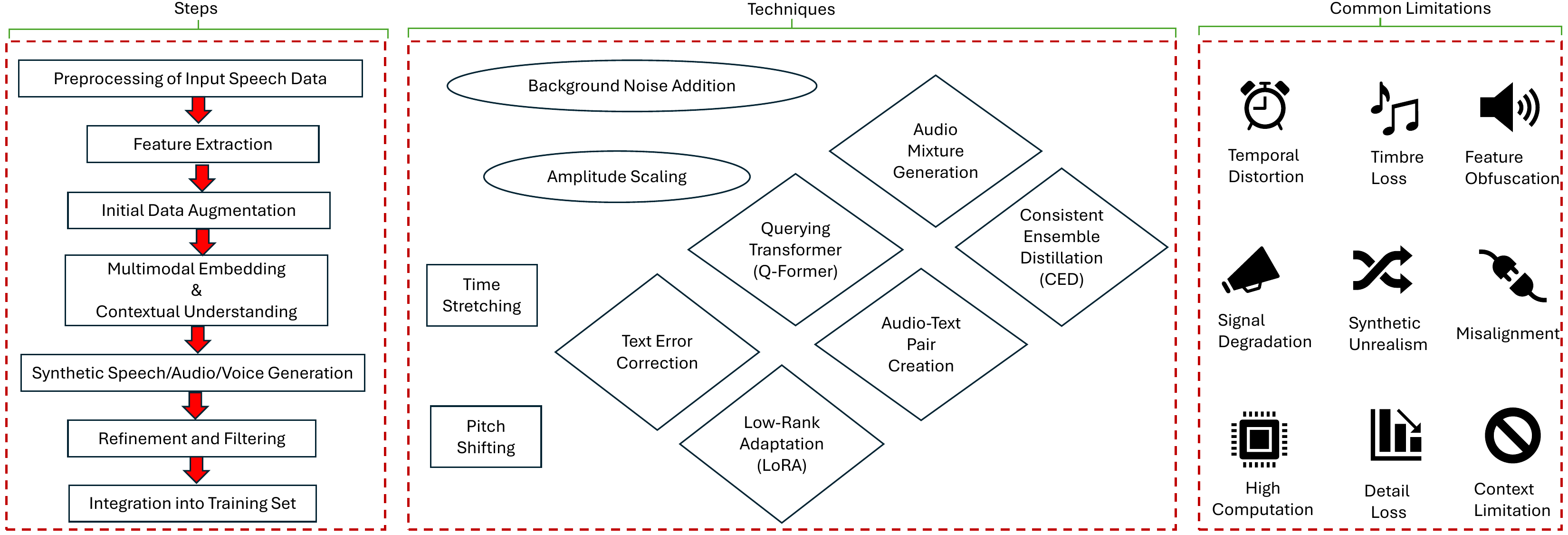}
    \caption{ LLM based speech data augmentation : showing the technical aspect of how audio/speech augmentation using LLM is performed, the techniques of augmenting speech data using LLM and their limitations}
    \label{fig:ImageLLMResultsAudioAug}
\end{figure*}

Multimodal LLM-based speech data augmentation can be explained in 7 steps as depicted in Figure \ref{fig:ImageLLMResultsAudioAug} (left side). The initial phase of LLM-based speech data augmentation starts with the meticulous preprocessing of raw audio data. This process involves sampling the audio at a consistent rate, normalizing volume levels to a standard decibel range, and segmenting the speech into shorter clips to facilitate processing efficiency. These steps are critical for ensuring uniformity in data quality, which is vital for effective feature extraction. The feature extraction step converts this standardized audio into a set of features more amenable to machine learning algorithms. Techniques such as extracting Mel-frequency cepstral coefficients (MFCCs) and generating spectrograms are common, as they distill the essential characteristics of speech necessary for recognizing phonetic elements. This extraction process is foundational for enhancing model training and accuracy in diverse acoustic conditions, as demonstrated by Cai et al. (\cite{cai2024av}) and reinforced by Dhingra et al. (\cite{dhingra2024speech}) in their work on audio-visual deepfake detection and speech de-identification.

The third step involves applying a variety of traditional audio augmentation techniques such as noise injection, time stretching, pitch shifting, and volume adjustment. These manipulations introduce critical variability into the audio data, mimicking real-world audio environments and thus enhancing the robustness of the subsequent models. This approach is crucial for training models to perform reliably across different speaking conditions and noise levels. Studies by Dhingra et al. (\cite{dhingra2024enhancing}) and Ma et al. (\cite{ma2024leveraging}) provide evidence that such diversity in training data significantly improves the resilience and accuracy of speech recognition and emotion recognition systems, respectively. The next step is embedding with Multimodal Context and Synthetic Speech Generation, where embeddings are generated using multimodal Large Language Models that integrate both textual and acoustic information, creating a nuanced understanding of the speech context. These embeddings are pivotal for generating synthetic speech through advanced neural network architectures like variational autoencoders or Generative Adversarial Networks (GANs). This process synthesizes new speech samples that retain the linguistic properties of the original data while varying in vocal intonations and environmental sounds, thereby enriching the training dataset with realistic, diverse samples. The integration of LLMs in this process, as explored by Xu (\cite{xu2024audiosetmix}) and Wu et al. (\cite{wu2024improving}), enables the generation of high-quality synthetic speech that significantly enhances training data diversity and model performance.

The final step is refinement, filteringa nd integration into DL training set which involve the refinement and filtering of the generated synthetic speech to ensure it meets quality standards without unwanted noise or artifacts. This step is critical to maintaining the usability of the synthetic samples in training robust speech recognition systems. Once refined, these samples are integrated into the main training dataset, where they contribute to a comprehensive and varied training environment. This enriched dataset is then used to train and continually refine deep learning models, ensuring they are well-adapted to a variety of speech patterns and acoustic environments. The effectiveness of this comprehensive training approach is supported by the work of Xu (\cite{xu2024audiosetmix}) and Ghosh et al. (\cite{ghosh2024synthio}), who have demonstrated how diversified training sets can significantly enhance the generalizability and accuracy of speech recognition systems.

Table \ref{tab:surveyLLMaudioTable} shows the detailed survey of audio data augmentation performed using multimodal LLMs in the last 5 (2020 till) years:

\begin{table*}[ht!]
\centering
\caption{Survey of Multi-modal LLMs in Audio/Voice/Speech Data Augmentation (Peer Reviewed Papers, 2020–present)}
\label{tab:surveyLLMaudioTable}
\scriptsize
\renewcommand{\arraystretch}{1.3} 
\begin{tabular}{|p{2cm}|p{2cm}|p{4.5cm}|p{4cm}|}
\hline
\textbf{LLM Name} & \textbf{Domain} & \textbf{Audio Augmentation Method} & \textbf{Key Insights} \\ \hline
Instructor LLM \& ChatGPT \cite{wu2024improving} & Audio Captioning & Fine-grained audio feature extraction and caption mix-ups & Achieved new SPIDEr-FL score record; won the 2023 DCASE AAC challenge. Mix-ups risk overfitting due to data complexity. \\ \hline
LLM-Assisted \cite{xu2024audiosetmix} & Audio-Text Pairing & Text embeddings and signal processing for audio-caption datasets & Improved performance on benchmarks; addressed lack of modifiers in datasets. Relies on alignment quality of text with audio. \\ \hline
LLM-Assisted \cite{dhingra2024enhancing} & Speech De-identification & Automated augmentation for robust PII detection & Enhanced accuracy for speech de-identification; reliant on LLM-generated data quality. \\ \hline
AV-Deepfake1M \cite{cai2024av} & Deepfake Detection & LLM-driven audio-visual manipulations & Generated 1M+ videos; tested state-of-the-art methods for deepfake detection. Performance challenges persist. \\ \hline
ER-PTM-LLM-TTS \cite{ma2024leveraging} & Speech Emotion Recognition & Emotional TTS, Speech PTM, Text LLM integration & Boosted SER performance with cutting-edge methods (data2vec, GPT-4, Azure TTS); integration complexity is high. \\ \hline
Speech De-Id NER \cite{dhingra2024speech} & Named Entity Recognition (NER) & Synthetic speech-style text generation for semi-automatic PII annotation & Enhanced speech de-identification NER; semi-automatic process requires manual review. \\ \hline
ArzEn-LLM \cite{heakl2024arzen} & Code-Switched ASR & LLM integration for Egyptian Arabic-English ASR and MT systems & Improved code-switching translation and ASR handling; dependent on high-quality data integration. \\ \hline
Nor-BERT \cite{hashmi2024self} & Hate Speech Detection & Lexical and semantic augmentations with Barlow Twins & Enhanced hate speech detection in Norwegian; reliant on dataset quality and scale. \\ \hline
AR-GPT-4 \cite{xu2024integrating} & Emergency Response & AR-LLM integration for task efficiency & Improved situational awareness and task performance; dependent on real-time integration. \\ \hline
LLM-Commentator \cite{cook2024llm} & Sports Commentary & LLM for real-time football commentary generation & Generated accurate football commentary in real-time; requires high-quality domain-specific training. \\ \hline
LLM-HRC \cite{gkournelos2024llm} & Human-Robot Collaboration & LLM for natural language programming in manufacturing & Improved HRC efficiency with intuitive interface; requires integration with Digital Twins and real-time control. \\ \hline
LAMB \cite{alier2025lamb} & Education & LLM for AI-powered educational assistants & Streamlined AI integration into LMS; focused on privacy and customizability. Dependent on robust LMS integration and educational standards. \\ \hline
LaMini-Flan-T5 \cite{senthilselvi2024abstractive} & Video Summarization & Abstractive summarization for YouTube videos & Provided concise summaries to improve accessibility; relies on accurate video transcripts. \\ \hline
ReSLM \cite{wang2024retrieval} & Spoken Dialog Systems & Retrieval-augmented dialog state tracking & Improved accuracy in domain-specific entity recognition; dependent on retrieval component quality. \\ \hline
MMed-Llama 3 \cite{qiu2024towards} & Multilingual Healthcare & Multilingual medical corpus for domain-specific adaptation & Facilitated multilingual medical understanding with benchmarks; corpus diversity is critical for performance. \\ \hline
GRBAS-Net \cite{hasebe2024effect} & Voice Classification & Noise-resilient classification for pathological voices & Enhanced GRBAS scale classification; performance deteriorates with increased noise. \\ \hline
\end{tabular}
\end{table*}

\subsubsection{Methods and Techniques}
\begin{itemize}
    \item \textbf{Background Noise Addition :} Background noise addition is a prevalent method in speech data augmentation aimed at enhancing the robustness of speech recognition systems under varied acoustic environments. By introducing different types of ambient noises, such as traffic, crowd, or white noise, during the training phase, models learn to maintain accuracy despite auditory distractions. This method has been pivotal in scenarios where speech recognition must perform reliably in non-ideal and noisy conditions. For instance, the works by \cite{xu2024audiosetmix} and \cite{wu2024improving} demonstrate how integrating complex audio backgrounds can prepare systems for real-world applications, significantly enhancing the environmental adaptability of speech recognition models. Additionally, \cite{dhingra2024enhancing} illustrates the effectiveness of this approach in the context of speech de-identification, where noise can mask personally identifiable information, enhancing privacy. \cite{cai2024av} also utilizes background noise to challenge the robustness of audio-visual deepfake detection models, ensuring they can perform under diverse conditions.

    \item \textbf{Amplititude Scaling :} Amplitude scaling involves adjusting the volume levels of speech recordings to simulate various speaking and listening conditions. This technique is essential for training models to recognize speech across a spectrum of vocal intensities and distances from the sound source, from whispers to shouts. Amplitude scaling helps in normalizing the loudness of audio inputs, which is crucial for consistent model performance. \cite{ma2024leveraging} applies this method to enhance the emotion recognition capabilities of systems by training them to detect subtle variations in speech dynamics that may indicate emotional states. Similarly, \cite{wu2024improving} and \cite{xu2024audiosetmix} have demonstrated how varying amplitude can prepare models for real-life scenarios where audio levels are not constant. \cite{ghosh2024synthio} uses amplitude scaling to test the resilience of audio classification systems, ensuring they remain effective regardless of volume variations.

    \item \textbf{Time Stretching: }Time stretching modifies the speed of speech playback without altering the pitch, effectively simulating different speech rates. This augmentation is crucial for developing ASR systems that can accurately transcribe inputs from speakers who articulate at varying speeds. By expanding or compressing the temporal duration of audio samples, models are trained to handle slow and fast speech effectively. Studies such as those by \cite{ma2024leveraging} and \cite{dhingra2024speech} have utilized time stretching to adapt models for emotional and secure speech recognition tasks, where delivery speed can vary significantly based on the speaker's emotional state or privacy concerns. Furthermore, \cite{xu2024audiosetmix} and \cite{cai2024av} incorporate time stretching to enhance the adaptability of models in multimedia environments where speech tempo can change contextually.

    \item \textbf{Pitch Shifting :} Pitch shifting adjusts the pitch or tone of the speech while keeping the speed constant. This method is instrumental in training models to recognize voices across different age groups, genders, and vocal characteristics. By simulating a wide range of vocal pitches, speech recognition systems can better identify and understand speakers from diverse demographic backgrounds. \cite{wu2024improving} and \cite{xu2024audiosetmix} employ pitch shifting to diversify the acoustic properties of training datasets, improving the system's ability to generalize across different speakers. \cite{ghosh2024synthio} and \cite{ma2024leveraging} also highlight the use of pitch variation to train more resilient and inclusive speech recognition and emotion detection models, ensuring effectiveness across varied vocal registers.

    \item \textbf{Text Error Correction :} Text error correction in the context of speech data augmentation often involves using LLMs to identify and correct transcription errors that arise from speech-to-text processes. This technique enhances the accuracy of ASR (Automatic Speech Recognition) systems by refining the textual output using contextual understanding and language models. The LLMs, such as those mentioned in \cite{dhingra2024speech} and \cite{ghosh2024synthio}, are trained on vast datasets, allowing them to predict and replace erroneous words or phrases accurately. Furthermore, the integration of these models in systems like the one described by \cite{xu2024audiosetmix} helps in adjusting text outputs to better match the spoken input, significantly reducing the error rates in transcription. \cite{sridhar2024enhancing} also highlights the application of LLMs in enhancing comprehension and response accuracy by correcting semantic errors, thus improving the overall reliability of speech-driven applications.

    \item \textbf{Querying Transformer (Q-former) :} The Querying Transformer, or Q-former, is an advanced model architecture that incorporates querying mechanisms into transformers for better handling and understanding of speech queries. This method is particularly useful in speech recognition systems where the context and intent behind spoken words are critical for accurate interpretation. By embedding querying capabilities, such as those discussed in \cite{ma2024leveraging} and \cite{whitehouse2023llm}, Q-formers can dynamically adjust their processing strategies based on the query's complexity and specificity. This adaptability is crucial in applications like customer service bots and interactive AI, where \cite{gkournelos2024llm} and \cite{lei2024contextualization} have demonstrated significant improvements in user interaction and satisfaction by employing Q-formers to accurately address user inquiries.

    \item \textbf{Audio Mixture Generation :} Audio Mixture Generation involves creating composite audio tracks from multiple sources, which is pivotal in training robust ASR systems to handle overlapping speech and background noise. This technique uses LLMs to intelligently blend different audio signals, maintaining a balance that mimics real-world scenarios. Studies such as \cite{xu2024audiosetmix} and \cite{ghosh2024synthio} utilize this method to generate training data that helps models learn to isolate relevant speech from noise. Additionally, \cite{goel2024audio} and \cite{yang2023uniaudio} explore the effectiveness of audio mixture generation in enhancing the diversity and complexity of audio datasets, thus preparing ASR systems for more challenging acoustic environments.

    \item \textbf{Low-Rank Adaptation (LoRA) :} LoRA is a technique used to fine-tune large LLMs efficiently by modifying only a small subset of model parameters, thus reducing computational costs while maintaining performance. In speech augmentation, LoRA can be applied to adapt pre-trained LLMs to specific audio tasks without extensive retraining. For instance, \cite{ma2024leveraging} and \cite{dhingra2024speech} apply LoRA to adjust models for emotional recognition and PII masking, respectively, by focusing on layers directly involved in these tasks. This method ensures that the LLMs remain lightweight and agile, as supported by \cite{whitehouse2023llm} and \cite{wang2024audio}, which highlight the practicality of LoRA in deploying advanced speech technologies in resource-constrained environments.

    \item \textbf{Audio-Text Pair Creation :} Audio-Text Pair Creation is central to training multimodal LLMs, where synchronization between spoken audio and corresponding textual data is crucial. This process involves generating or curating matched pairs of text and audio to train models in tasks like audio captioning and speech recognition. The effectiveness of this method is evident in works like \cite{xu2024audiosetmix} and \cite{lei2024contextualization}, where aligned audio-text pairs improve the model's ability to learn contextual nuances. Additionally, \cite{ghosh2024synthio} and \cite{cook2024llm} utilize this technique to enhance the accuracy and contextual relevance of generated captions, providing a richer training dataset that mimics real-world interactions.

    \item \textbf{Consistent Ensemble Distillation (CED) :} CED) is a method that leverages the strengths of multiple models to create a single, more robust model. In the context of speech data augmentation, CED can be used to combine the outputs of various LLMs to enhance speech recognition accuracy under diverse conditions. This approach, as discussed in \cite{ma2024leveraging} and \cite{dhingra2024speech}, involves distilling knowledge from multiple specialized models into a unified system that exhibits improved performance and resilience. \cite{gkournelos2024llm} and \cite{lei2024contextualization} further illustrate how CED facilitates the integration of diverse linguistic and acoustic features, enhancing the system's ability to handle complex speech patterns and ambient sounds effectively.
\end{itemize}

\subsubsection{Limitations and Potential Solutions}

\begin{itemize}
    \item \textbf{Temporal Distortion:}Temporal distortion is a significant limitation in LLM-based audio data augmentation due to the constraints of large language models (LLMs) in handling the temporal dynamics of speech \cite{cuskley2024limitations, lee2024performance}. LLMs, primarily developed for processing text, lack the specialized architecture required to effectively manage time-dependent elements in audio signals \cite{ma2024leveraging}. This deficiency leads to challenges in maintaining the naturalness and rhythm of speech when the speed is altered through temporal distortion. Such modifications can make the speech sound unnatural or robotic, potentially degrading the training quality of speech recognition systems. Moreover, the alteration of speech speed can disrupt the acoustic cues essential for models to learn effectively, impacting their ability to perform accurately in real-world scenarios where speech tempo varies \cite{srivastava2024unvoiced, dhingra2024enhancing}. Additionally, the modified timing can complicate the comprehension of speech by language models, as the distorted temporal features may hinder their ability to interpret syntactic and semantic information correctly.  

    To address the temporal distortion in LLM-based audio augmentation effectively, future advancements should involve refining LLM architectures to accommodate audio-specific temporal dynamics, potentially through the integration of temporal attention mechanisms or recurrent neural network layers \cite{ding2024data}. Additionally, developing hybrid models that merge LLMs with digital signal processing techniques could enhance the handling of variable speech rates, thereby preserving the naturalness of speech \cite{kheddar2024automatic}. Furthermore, diversifying training datasets to include audio samples with varied speech tempos can improve model robustness, ensuring better performance across different real-world speech scenarios \cite{li2024groundinggpt, ding2024data}.

    \item \textbf{Timbre Loss:} Timbre loss represents another significant technical challenge in LLM-based audio augmentation, primarily because LLM architectures, traditionally optimized for text, do not effectively preserve the complex spectral qualities that define an audio signal's unique timbre \cite{zhu2024llm,ma2024leveraging}. This limitation is critical because timbre imparts the distinct character to voices or musical sounds, which is essential for applications like speech synthesis and music generation \cite{heidemann2016system}. The alteration of timbre during processes such as pitch shifting or speed modification can result in audio outputs that fail to replicate the original sound's perceptual attributes, thereby diminishing the authenticity and effectiveness of the augmented audio in practical applications \cite{schneider2018perception}. 
    
    To address timbre loss in LLM-based audio augmentation, several approaches have been proposed  such as Joint modeling of timbre and content \cite{chen2024takin}, where the system employs a joint modeling approach that integrates timbre features with both supervised and self-supervised content representations. This enhances speaker similarity and intelligibility, allowing for more accurate reproduction detailed speaker characteristics. Likewise, Conditional flow matching-based decoder can optimize the alignment between timbre and content features. This leads to more natural and accurate voice conversions, preserving timbre qualities \cite{chen2024takin}. Additionally,  semantic-acoustic integration approach integrates both semantic and acoustic features into a unified tokenization framework. By employing a distinctive "X-shaped" structure with two inputs and two outputs, it enables simultaneous embedding learning of semantic richness and acoustic fidelity for every token. This design helps in preserving timbre information alongside semantic content \cite{ye2024codec}. To further address timbre loss in LLM-based audio augmentation, additional strategies could include the implementation of advanced deep learning architectures such as GANs are tailored specifically for audio tasks. These networks can learn to generate highly realistic audio samples by training on a competitive dynamic between the generative model producing audio and the discriminative model evaluating it \cite{qian2019data, barreto2023generative}. Additionally, integrating multimodal data during the training phase can enhance the model’s ability to understand and replicate timbre by exposing it to a broader range of audio characteristics \cite{zhang2024multimodal}. Techniques like transfer learning could also be employed, where models pre-trained on vast, diverse datasets are fine-tuned with specific timbre data, improving the model's generalization capabilities to new voices or sounds while maintaining the unique timbral qualities \cite{mariya2023data} .

    \item \textbf{Feature Obfuscation :} Feature obfuscation is a limitation of LLM-based speech augmentation due to the challenges it presents in preserving important speech characteristics while attempting to protect privacy or modify content. This limitation manifests in several ways such as  Loss of acoustic fidelity \cite{dhingra2024speech} and difficulty in preserving prosody \cite{peng2024voicetextblender}. When obfuscating features for privacy reasons, such as in speech de-identification, there's a risk of degrading the overall acoustic quality of the augmented speech \cite{dhingra2024speech}. This can impact the naturalness and intelligibility of the output. LLMs may struggle to maintain the original prosodic features of speech when augmenting or obfuscating content, potentially leading to unnatural-sounding output \cite{peng2024voicetextblender}. Additionally, feature obfuscation inadvertently risks obscuring critical acoustic features. For example, methods aiming to anonymize speaker identity might modify pitch or tone, potentially leading to a loss in the naturalness and expressiveness of speech \cite{nautsch2019preserving}. This degradation not only affects the audio's perceptual quality but can also impair the effectiveness of speech recognition systems that rely on distinct acoustic signals for accurate processing. Furthermore, when LLMs attempt to balance feature obfuscation with maintaining intelligibility, the resulting speech might not faithfully represent the original speaker's emotional or prosodic cues, thus diminishing the listener's experience and the speech's emotional conveyance.

    Addressing feature obfuscation in LLM-based speech augmentation requires a balanced approach that preserves essential speech characteristics while maintaining privacy. One effective method is the implementation of differential privacy techniques during the training phase of speech synthesis models \cite{nautsch2019preserving}. This involves adding controlled noise to training data, which helps obscure individual speech features without significantly degrading overall audio quality. Additionally, advanced machine learning strategies like adversarial training can be employed, where models are trained to resist attempts at extracting specific features, thereby enhancing privacy \cite{hu2019adversarial, chen2022towards}.

    \item \textbf{Signal Degradation :} Signal degradation represents a substantial challenge and limitation in LLM-based audio or speech augmentation due to its adverse effects on the audio quality and perceptual integrity of the output. In augmentation processes that involve LLMs, signal degradation typically arises from operations such as compression, format conversion, or even the introduction of synthetic elements that can diminish the audio's natural qualities \cite{han2024review, ogof2024enhancing}. This degradation can crucially impair the audio's clarity and fidelity, making it less suitable for applications reliant on high-quality audio inputs, such as speech recognition technologies and auditory user interfaces \cite{kheddar2024automatic}. Furthermore, the authenticity and expressiveness of speech are often compromised, leading to outputs that might not accurately convey the intended emotions\cite{srivastava2024unvoiced}. 

    To address the issue of signal degradation in LLM-based speech augmentation, it's essential to integrate advanced audio processing techniques that enhance signal integrity while maintaining or improving the natural characteristics of speech \cite{kheddar2024automatic, ding2024data}. One effective approach involves the use of sophisticated denoising algorithms that can remove unwanted noise without affecting the core audio signal. Additionally, implementing dynamic range compression can help maintain audio quality by balancing the volume levels within a speech, thus preventing distortions caused by volume peaks or lows. Further advancements can be made by utilizing high-resolution signal processing techniques during the augmentation process to preserve the fine details of the speech waveform. Machine learning models, particularly those trained on a diverse set of high-quality audio samples, can be employed to learn and predict optimal signal characteristics that minimize loss during transformations.

    \item \textbf{Synthetic Unrealism :} Synthetic Unrealism poses a significant challenge in audio data augmentation using multimodal LLMs due to the potential introduction of unrealistic or inconsistent audio features \cite{dekel2024speak}. This phenomenon can lead to the generation of augmented data that does not accurately represent real-world audio characteristics, potentially compromising the model's ability to generalize to authentic scenarios \cite{zheng2025teaching}. The issue is exacerbated by the complex nature of audio data, which encompasses various dimensions such as pitch, tempo, and timbre. LLMs, while adept at generating diverse content, may struggle to maintain the intricate relationships between these audio components, resulting in synthetic samples that lack coherence or natural acoustic properties \cite{ghiuruau2024distinguishing, dekel2024speak}. Furthermore, the risk of overfitting to artificially generated patterns increases, potentially leading to reduced model performance on genuine audio inputs.

    To tackle the issue of synthetic unrealism in LLM-based audio data augmentation, a multifaceted approach focusing on enhancing the model's capability to generate more realistic and coherent audio samples is essential. First, integrating more comprehensive and high-fidelity training datasets can provide a richer foundation for learning authentic audio characteristics \cite{bakir2024comprehensive}. This can involve using recordings from diverse environments and contexts to cover a broader spectrum of real-world audio features. Secondly, applying advanced machine learning techniques such as conditional  GANs (cGANs) could enforce more stringent checks on the realism of generated audio \cite{wali2022generative, sheng2018data}. These networks can learn to distinguish between real and synthetic audio samples, pushing the generative model to produce outputs that are indistinguishable from genuine recordings. Finally, regularizing the training process to prevent overfitting is crucial. Techniques like dropout, data augmentation (ironically), and cross-validation can be employed to ensure the LLM does not learn to replicate only the training data's patterns but can generalize well across unseen audio inputs \cite{mushtaq2020environmental}.

    \item \textbf{Misalignment:} Misalignment in LLM-based audio or speech data augmentation refers to discrepancies between augmented audio data and the original audio's characteristics, such as emotional tone or  linguistic details \cite{shen2023mingling}. This issue typically arises because LLMs, originally designed for text processing, may not adequately capture complex audio features like pitch, timbre, or rhythm \cite{fathullah2024audiochatllama, shen2023mingling}. Such misalignment is problematic for deep learning applications reliant on precise, high-quality audio data for training. For example, in speech recognition or voice-assisted AI, misaligned data can lead to systems that misinterpret user commands or fail in real applications, reducing the reliability and effectiveness of AI interactions with human speech \cite{vu2024gptvoicetasker}.

    To address misalignment in LLM-based audio augmentation for deep learning applications, future solutions should focus on enhancing model sensitivity. One effective approach is to integrate audio-specific adaptations into the LLM architectures, such as using convolutional layers that can better capture the temporal and spectral dynamics of sound \cite{panagopoulou2025x}. Advanced techniques like attention mechanisms could also be employed to focus on critical features of the audio signal, improving the model's ability to retain important acoustic properties during augmentation \cite{majhi2024automatic, rossenbach2020generating}. Moreover, incorporating multimodal training data that includes both audio and corresponding textual annotations can help improve the alignment between the generated audio and its intended meaning \cite{fan2024transformer}. This would enable the LLM to learn more comprehensive representations of audio features and their contextual significance. Additionally, continuous evaluation and feedback loops involving real-world testing and user input should be established to iteratively refine the models, ensuring that the audio outputs remain aligned with practical applications and user expectations.  

    \item \textbf{High Computation}: LLM-based data augmentation for audio or speech is computationally intensive primarily due to the sheer size and complexity of these models. Large language models contain billions of parameters, demanding extensive memory and processing power for operations \cite{raiaan2024review}. This computational requirement becomes a significant challenge because it necessitates the use of specialized hardware like GPUs or TPUs, which can be costly and less accessible for many researchers and developers \cite{kachris2025survey}. Moreover, the energy-intensive nature of training and deploying these models adds to their operational costs and environmental impact \cite{wen2024generative, luccioni2024power}. This high computational demand limits the scalability of deploying LLMs for audio data augmentation, especially in environments with limited hardware resources. Additionally, the requirement for significant computational resources can impede rapid testing and iteration of models, slowing down the development cycle and innovation in audio processing applications.

    To address the high computational demands of LLM-based audio augmentation, several efficiency-enhancing strategies can be implemented. Model pruning and quantization are key techniques that streamline the model by reducing parameters and computation precision, respectively, without significantly impacting performance \cite{tan2021towards}. Knowledge distillation transfers expertise from a large model to a smaller, more efficient one, maintaining accuracy while decreasing computational load \cite{cai2024leveraging}. Efficient hardware utilization, such as specialized neural network processors or hardware accelerators like FPGAs, can boost processing speed and reduce power consumption \cite{kim2021fpga}. Likewise,  Adaptive computing techniques adjust model complexity based on task requirements, conserving resources for simpler tasks \cite{bell2020adaptation}.

    \item \textbf{Detail Loss:} Detail loss in LLM-based audio data augmentation significantly hampers the utility of this technology due to several inherent limitations associated with audio complexity \cite{xu2024audiosetmix}. Audio data consists of dynamic acoustic features such as pitch, tempo, timbre, and emotional inflections, which are crucial for maintaining the authenticity of the sound \cite{bhattarai2023comprehensive}. LLMs, although capable of generating varied audio content, often struggle with preserving these intricate details. This limitation arises partly because of the inadequate representation of fine-grained features in the training datasets, which tend not to include high-quality text-audio pairs that capture detailed acoustic properties and object relationships. For instance, in speech augmentation, LLM might modify a voice's fundamental frequency but fail to retain the speaker’s distinct vocal characteristics or the subtleties of their emotional tone \cite{srivastava2024unvoiced}. Similarly, in music augmentation, critical attributes like the attack and decay phases of notes played by different instruments might be inaccurately rendered or completely lost \cite{meerza2024harmonycloak}. Such detail loss not only diminishes the fidelity of the augmented audio compared to real-world sounds but also limits the practical applications of LLMs in fields requiring high precision in audio reproduction, such as virtual reality, filmmaking, and advanced user interfaces. 

    To reduce detail loss in LLM-based audio augmentation, enhancing training datasets with high-quality, diverse recordings and employing specialized neural network architectures and multi-modal learning approaches can significantly improve the model's ability to accurately generate complex audio data.

    \item \textbf{Context Limitation:} Context limitation in LLM-based speech augmentation refers to the challenge these models face in accurately understanding and generating contextually appropriate audio responses \cite{peng2024survey}. Unlike simple text generation, speech involves complex elements such as tone, inflection, and rhythm, which are deeply embedded in the specific contexts of conversation or narration \cite{satish2024voice}. Scientifically, the main issue arises because LLMs typically generate outputs based on patterns recognized in training data, without a genuine understanding of human emotions, intentions, or the subtleties of spoken language dynamics \cite{chen2024evolution}. This becomes a significant limitation in applications where speech needs to convey more than just words, such as in emotional speech synthesis for virtual assistants, interactive learning environments, or customer service bots. The inability to incorporate real-time context and emotional undertones often results in responses that, while grammatically correct, can seem out of place, emotionally flat, or inappropriate to human listeners. 

    To overcome context limitation in LLM-based speech data augmentation, future strategies should aim to enhance models' contextual and emotional understanding through advanced training methodologies and architectural improvements. Implementing context-aware training, where models are exposed to a diverse range of speech scenarios with varying emotional and situational contexts, can enrich their understanding. For example, training on datasets from theatrical dialogues or emotionally charged speeches could help models better recognize and replicate context-specific speech patterns. Additionally, integrating Emotional Neural Networks (EmoNNs), which are specially trained to recognize and replicate human emotions in speech, can significantly enhance the emotional intelligence of LLMs. Multi-modal learning is another promising approach, where combining inputs like text, audio cues, and visual data such as facial expressions or gestures can deepen models’ contextual comprehension. For instance, training a model using both the audio and video of a speaker can teach it to correlate speech patterns with corresponding physical expressions. Furthermore, developing adaptive models that adjust their outputs based on real-time feedback can also help. Interactive voice response systems that modify their tone in response to user emotions like frustration or happiness exemplify this approach.
\end{itemize}

\section{Discussion}

 This survey provides an exhaustive review of the latest advancements in LLM-based data augmentation, covering a broad range of applications across image, text, and audio data modalities. It includes both 2D and 3D image data augmentation techniques, ensuring a holistic understanding of the field.
The survey delves into the technical processes, methods, and techniques of LLM-based data augmentation, offering detailed insights that are crucial for researchers and practitioners. This depth extends to a critical examination of the limitations and potential solutions, enhancing the practical value of the survey.
By focusing on publications from the last five years, the survey ensures that the findings are relevant and up-to-date, reflecting the cutting-edge developments in LLM applications and data augmentation strategies.

\subsection{Impact}

\textbf{Theoretical Impact} This survey contributes significantly to the theoretical understanding of LLM-based data augmentation by identifying methodologies that enhance the robustness and generalization of AI models. These insights are essential for enabling reliable and adaptive performance in diverse real-world scenarios. By critically evaluating current limitations and gaps in the literature, the survey not only enriches the foundational knowledge of the field but also provides a roadmap for addressing these challenges. Additionally, the proposed research directions stimulate further innovation, guiding researchers toward developing advanced, more effective AI systems.

\textbf{Practical Impact} 
The practical implications of this survey are far-reaching for both academia and industry. For practitioners, it offers a detailed analysis of state-of-the-art techniques that can be directly applied to improve model performance across various applications, including computer vision, natural language processing, and audio processing. The survey also identifies actionable solutions to common challenges, equipping developers and engineers with tools to implement efficient and scalable data augmentation pipelines. Stakeholders can use this knowledge to harness the potential of LLM-based data augmentation for innovation in their respective domains.

\subsection{Future Perspectives}
\begin{figure}[htbp]
    \centering
    \includegraphics[width=0.78\linewidth]{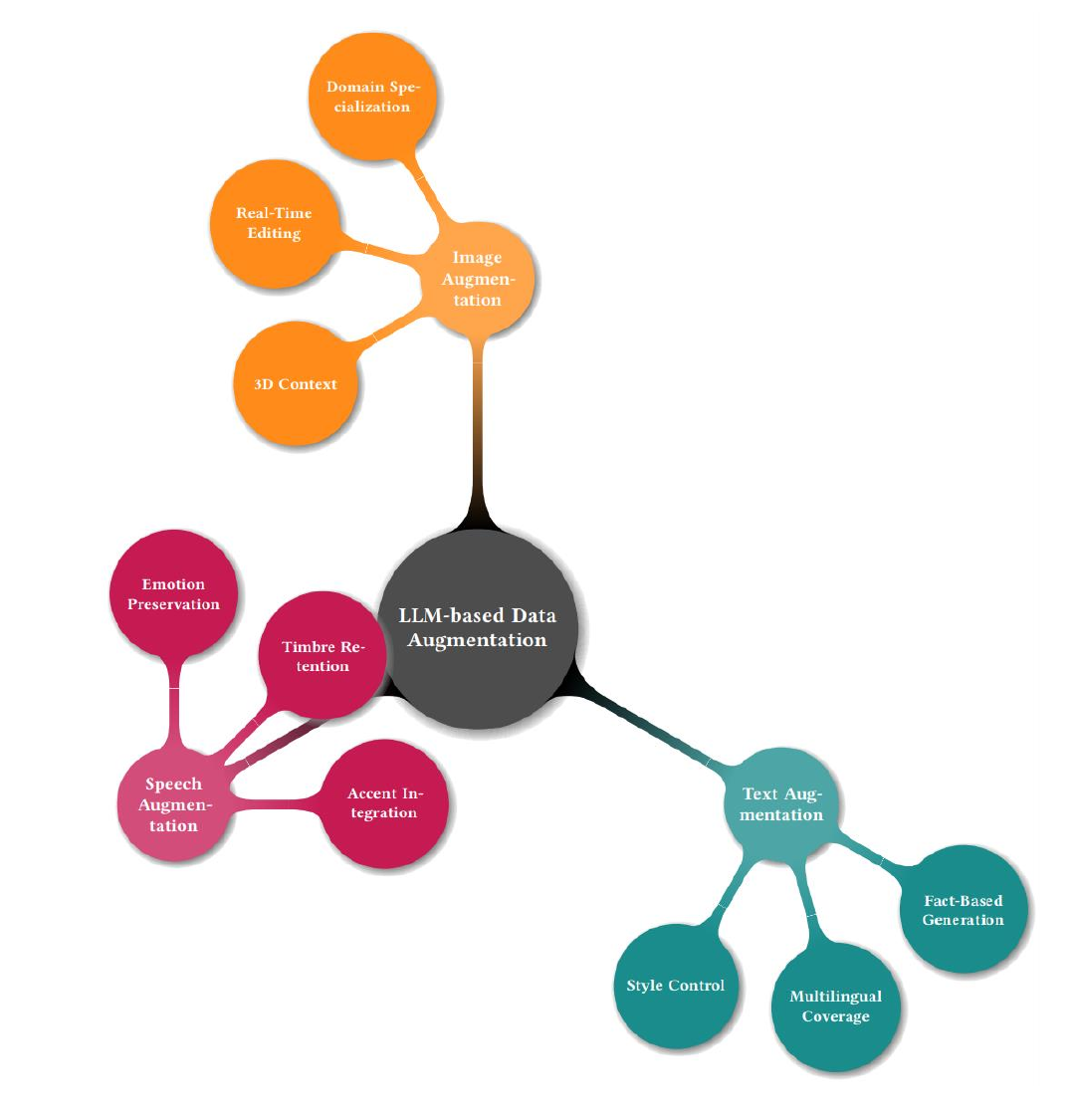}
    \caption{ A mind map illustrating future perspectives of LLM-based data augmentation for image, text, and
speech domains.}
    \label{fig:mindmap-llm-aug}
\end{figure}

The future of LLM-based data augmentation spans a diverse range of innovations, as illustrated in Figure \ref{fig:mindmap-llm-aug}.  Here, image augmentation techniques are projected to emphasize 3D context and multi-view data processing, supporting more realistic scene representations and advanced real-time editing pipelines. At the same time, enhanced domain specialization will drive the adaptation of augmentation strategies to particular fields such as healthcare, manufacturing, or robotics, all while maintaining semantic consistency through robust validation mechanisms. Similarly, text augmentation is set to benefit from fact-based, knowledge-grounded generation, improving the reliability and interpretability of synthetic text. This is especially pertinent in multilingual low-resource contexts, where LLMs could fill data gaps for underrepresented languages, all the while supporting fine-grained style and emotion control to produce more expressive outputs and on-the-fly language adaptation in real-world communication systems.

In the realm of speech augmentation, emotion and prosody preservation will be key to delivering authentic, human-like audio data, making voice-based AI more natural and empathetic. High-fidelity timbre retention likewise remains an open challenge, especially for sensitive applications such as assistive speech devices and high-end multimedia production. Augmentation tailored to context-aware accent integration promises a more inclusive approach, broadening AI’s reach across diverse linguistic communities. Finally, robust streaming augmentation methods could enable real-time transformations of audio for live scenarios, such as simultaneous interpretation or interactive voice assistants. 

Additionally, the advent of reinforcement learning-based methodologies such as DeepSeek R1 heralds a transformative shift in LLM based data augmentation \cite{guo2025deepseek}. Traditionally, LLMs have been dependent on large, meticulously annotated datasets, which can restrict their adaptability and scope for innovation. However, DeepSeek R1 shifts away from this reliance by using reinforcement learning (RL) to enable models to continuously learn and evolve autonomously by interacting with their environment and adjusting based on feedback. This shift is particularly transformative for data augmentation strategies. Reinforcement learning allows models like DeepSeek R1 to not only generate their own data but also to continuously refine this data in response to new challenges and feedback. This dynamic approach to data generation represents a significant move away from static, pre-labeled datasets, reducing costs and dependency on human annotation. The implications for data augmentation are profound. As models become capable of self-augmentation, they can explore a broader range of scenarios and adapt to complex, unforeseen challenges more effectively than models restricted by their initial training data. This not only enhances the flexibility and depth of the models' knowledge but also opens up new possibilities for efficiency in model training. As reinforcement learning continues to advance, it is set to reshape the landscape of data augmentation, promising a future where AI systems are more robust, context-aware, and adaptable. This development indicates a new era where LLMs can surpass traditional capabilities, driven by their ability to learn independently and innovate beyond the boundaries of their initial programming.

\section{Conclusion}
\label{conclusion}
In this comprehensive survey, we have explored the role of multimodal LLMs within the domain of  AI for data augmentation practice. We focusing specifically on three fundamental data modalities: image, text, and speech. The findings of this survey shows that over the past five years( 2020 onwards), there has been a significant increase in the publication of peer-reviewed and preprint papers that utilize LLMs for data augmentation. Our survey methodically assessed the state-of-the-art techniques reported in recent literature, extracting key details about the technical processes employed in LLM-based data augmentation. We delved into the diverse methods and techniques utilized to augment data through LLMs, providing a clear overview of how these models are applied to enhance the quality and effectiveness of datasets across different applications. This included a thorough analysis of augmentation strategies for 2D and 3D images, the enhancement of textual data for better language model training, and the augmentation of audio data to improve the robustness and accuracy of speech recognition systems. Moreover, our study identified limitations inherent in current LLM-based data augmentation strategies. These issues range from the potential for introducing bias and reducing model generalizability to challenges in maintaining the semantic integrity and contextual relevance of augmented data. To address these limitations, we discussed potential solutions that could pave the way for more effective and reliable data augmentation practices in the future. The findings of this survey can be used to throry for next generation data augmentation.

Ranjan Sapkota led research conceptualization, methodology, and writing; Shaina Raza, Maged Shoman, and Achyut Paudel contributed to review and methodology; Manoj Karkee provided overall supervision and funding.


\bibliographystyle{unsrtnat}  
\bibliography{sample-acmsmall.bib}

\begin{thebibliography}{293}
\providecommand{\natexlab}[1]{#1}
\providecommand{\url}[1]{\texttt{#1}}
\expandafter\ifx\csname urlstyle\endcsname\relax
  \providecommand{\doi}[1]{doi: #1}\else
  \providecommand{\doi}{doi: \begingroup \urlstyle{rm}\Url}\fi

\bibitem[Iglesias et~al.(2023)Iglesias, Talavera, Gonz{\'a}lez-Prieto, Mozo, and G{\'o}mez-Canaval]{iglesias2023data}
Guillermo Iglesias, Edgar Talavera, {\'A}ngel Gonz{\'a}lez-Prieto, Alberto Mozo, and Sandra G{\'o}mez-Canaval.
\newblock Data augmentation techniques in time series domain: a survey and taxonomy.
\newblock \emph{Neural Computing and Applications}, 35\penalty0 (14):\penalty0 10123--10145, 2023.

\bibitem[Shorten and Khoshgoftaar(2019)]{shorten2019survey}
Connor Shorten and Taghi~M Khoshgoftaar.
\newblock A survey on image data augmentation for deep learning.
\newblock \emph{Journal of big data}, 6\penalty0 (1):\penalty0 1--48, 2019.

\bibitem[Ratner et~al.(2017)Ratner, Ehrenberg, Hussain, Dunnmon, and R{\'e}]{ratner2017learning}
Alexander~J Ratner, Henry Ehrenberg, Zeshan Hussain, Jared Dunnmon, and Christopher R{\'e}.
\newblock Learning to compose domain-specific transformations for data augmentation.
\newblock \emph{Advances in neural information processing systems}, 30, 2017.

\bibitem[Mumuni and Mumuni(2022)]{mumuni2022data}
Alhassan Mumuni and Fuseini Mumuni.
\newblock Data augmentation: A comprehensive survey of modern approaches.
\newblock \emph{Array}, 16:\penalty0 100258, 2022.

\bibitem[Iwana and Uchida(2021)]{iwana2021empirical}
Brian~Kenji Iwana and Seiichi Uchida.
\newblock An empirical survey of data augmentation for time series classification with neural networks.
\newblock \emph{Plos one}, 16\penalty0 (7):\penalty0 e0254841, 2021.

\bibitem[Maharana et~al.(2022)Maharana, Mondal, and Nemade]{maharana2022review}
Kiran Maharana, Surajit Mondal, and Bhushankumar Nemade.
\newblock A review: Data pre-processing and data augmentation techniques.
\newblock \emph{Global Transitions Proceedings}, 3\penalty0 (1):\penalty0 91--99, 2022.

\bibitem[Bayer et~al.(2023)Bayer, Kaufhold, Buchhold, Keller, Dallmeyer, and Reuter]{bayer2023data}
Markus Bayer, Marc-Andr{\'e} Kaufhold, Bj{\"o}rn Buchhold, Marcel Keller, J{\"o}rg Dallmeyer, and Christian Reuter.
\newblock Data augmentation in natural language processing: a novel text generation approach for long and short text classifiers.
\newblock \emph{International journal of machine learning and cybernetics}, 14\penalty0 (1):\penalty0 135--150, 2023.

\bibitem[Kaur et~al.(2021)Kaur, Khehra, and Mavi]{kaur2021data}
Parvinder Kaur, Baljit~Singh Khehra, and Er~Bhupinder~Singh Mavi.
\newblock Data augmentation for object detection: A review.
\newblock In \emph{2021 IEEE International Midwest Symposium on Circuits and Systems (MWSCAS)}, pages 537--543. IEEE, 2021.

\bibitem[Abonizio et~al.(2021)Abonizio, Paraiso, and Barbon]{abonizio2021toward}
Hugo~Queiroz Abonizio, Emerson~Cabrera Paraiso, and Sylvio Barbon.
\newblock Toward text data augmentation for sentiment analysis.
\newblock \emph{IEEE Transactions on Artificial Intelligence}, 3\penalty0 (5):\penalty0 657--668, 2021.

\bibitem[Pervaiz et~al.(2020)Pervaiz, Hussain, Israr, Tahir, Raja, Baloch, Ishmanov, and Zikria]{pervaiz2020incorporating}
Ayesha Pervaiz, Fawad Hussain, Huma Israr, Muhammad~Ali Tahir, Fawad~Riasat Raja, Naveed~Khan Baloch, Farruh Ishmanov, and Yousaf~Bin Zikria.
\newblock Incorporating noise robustness in speech command recognition by noise augmentation of training data.
\newblock \emph{Sensors}, 20\penalty0 (8):\penalty0 2326, 2020.

\bibitem[Veluri et~al.(2024)Veluri, Itani, Chen, Yoshioka, and Gollakota]{veluri2024look}
Bandhav Veluri, Malek Itani, Tuochao Chen, Takuya Yoshioka, and Shyamnath Gollakota.
\newblock Look once to hear: Target speech hearing with noisy examples.
\newblock In \emph{Proceedings of the CHI Conference on Human Factors in Computing Systems}, pages 1--16, 2024.

\bibitem[Zhang et~al.(2023)Zhang, Carballo, Yang, and Takeda]{zhang2023perception}
Yuxiao Zhang, Alexander Carballo, Hanting Yang, and Kazuya Takeda.
\newblock Perception and sensing for autonomous vehicles under adverse weather conditions: A survey.
\newblock \emph{ISPRS Journal of Photogrammetry and Remote Sensing}, 196:\penalty0 146--177, 2023.

\bibitem[J{\"o}ckel et~al.(2019)J{\"o}ckel, Kl{\"a}s, and Mart{\'\i}nez-Fern{\'a}ndez]{jockel2019safe}
Lisa J{\"o}ckel, Michael Kl{\"a}s, and Silverio Mart{\'\i}nez-Fern{\'a}ndez.
\newblock Safe traffic sign recognition through data augmentation for autonomous vehicles software.
\newblock In \emph{2019 IEEE 19th International Conference on Software Quality, Reliability and Security Companion (QRS-C)}, pages 540--541. IEEE, 2019.

\bibitem[Shin et~al.(2018)Shin, Tenenholtz, Rogers, Schwarz, Senjem, Gunter, Andriole, and Michalski]{shin2018medical}
Hoo-Chang Shin, Neil~A Tenenholtz, Jameson~K Rogers, Christopher~G Schwarz, Matthew~L Senjem, Jeffrey~L Gunter, Katherine~P Andriole, and Mark Michalski.
\newblock Medical image synthesis for data augmentation and anonymization using generative adversarial networks.
\newblock In \emph{Simulation and Synthesis in Medical Imaging: Third International Workshop, SASHIMI 2018, Held in Conjunction with MICCAI 2018, Granada, Spain, September 16, 2018, Proceedings 3}, pages 1--11. Springer, 2018.

\bibitem[Chlap et~al.(2021)Chlap, Min, Vandenberg, Dowling, Holloway, and Haworth]{chlap2021review}
Phillip Chlap, Hang Min, Nym Vandenberg, Jason Dowling, Lois Holloway, and Annette Haworth.
\newblock A review of medical image data augmentation techniques for deep learning applications.
\newblock \emph{Journal of Medical Imaging and Radiation Oncology}, 65\penalty0 (5):\penalty0 545--563, 2021.

\bibitem[{\c{S}}ahin(2022)]{csahin2022augment}
G{\"o}zde~G{\"u}l {\c{S}}ahin.
\newblock To augment or not to augment? a comparative study on text augmentation techniques for low-resource nlp.
\newblock \emph{Computational Linguistics}, 48\penalty0 (1):\penalty0 5--42, 2022.

\bibitem[Chen et~al.(2023)Chen, Tam, Raffel, Bansal, and Yang]{chen2023empirical}
Jiaao Chen, Derek Tam, Colin Raffel, Mohit Bansal, and Diyi Yang.
\newblock An empirical survey of data augmentation for limited data learning in nlp.
\newblock \emph{Transactions of the Association for Computational Linguistics}, 11:\penalty0 191--211, 2023.

\bibitem[Wang et~al.(2020{\natexlab{a}})Wang, Wang, and Lian]{wang2020survey}
Xiang Wang, Kai Wang, and Shiguo Lian.
\newblock A survey on face data augmentation for the training of deep neural networks.
\newblock \emph{Neural computing and applications}, 32\penalty0 (19):\penalty0 15503--15531, 2020{\natexlab{a}}.

\bibitem[Ding et~al.(2024)Ding, Qin, Zhao, Luo, Li, Chen, Xia, Hu, Tuan, and Joty]{ding2024data}
Bosheng Ding, Chengwei Qin, Ruochen Zhao, Tianze Luo, Xinze Li, Guizhen Chen, Wenhan Xia, Junjie Hu, Luu~Anh Tuan, and Shafiq Joty.
\newblock Data augmentation using llms: Data perspectives, learning paradigms and challenges.
\newblock In \emph{Findings of the Association for Computational Linguistics ACL 2024}, pages 1679--1705, 2024.

\bibitem[Sufi(2024)]{sufi2024generative}
Fahim Sufi.
\newblock Generative pre-trained transformer (gpt) in research: A systematic review on data augmentation.
\newblock \emph{Information}, 15\penalty0 (2):\penalty0 99, 2024.

\bibitem[Hu et~al.(2023)Hu, Liu, Zhao, Hou, Nie, and Li]{hu2023survey}
Linmei Hu, Zeyi Liu, Ziwang Zhao, Lei Hou, Liqiang Nie, and Juanzi Li.
\newblock A survey of knowledge enhanced pre-trained language models.
\newblock \emph{IEEE Transactions on Knowledge and Data Engineering}, 2023.

\bibitem[Gracia~Mois{\'e}s et~al.(2023)Gracia~Mois{\'e}s, Vitoria~Pascual, Imas~Gonz{\'a}lez, and Ruiz~Zamarre{\~n}o]{gracia2023data}
Ander Gracia~Mois{\'e}s, Ignacio Vitoria~Pascual, Jos{\'e}~Javier Imas~Gonz{\'a}lez, and Carlos Ruiz~Zamarre{\~n}o.
\newblock Data augmentation techniques for machine learning applied to optical spectroscopy datasets in agrifood applications: A comprehensive review.
\newblock \emph{Sensors}, 23\penalty0 (20):\penalty0 8562, 2023.

\bibitem[Al-Fraihat et~al.(2024)Al-Fraihat, Sharrab, Alzyoud, Qahmash, Tarawneh, and Maaita]{al2024speech}
Dimah Al-Fraihat, Yousef Sharrab, Faisal Alzyoud, Ayman Qahmash, Monther Tarawneh, and Adi Maaita.
\newblock Speech recognition utilizing deep learning: A systematic review of the latest developments.
\newblock \emph{Human-centric Computing and Information Sciences}, 14, 2024.

\bibitem[Abayomi-Alli et~al.(2022)Abayomi-Alli, Dama{\v{s}}evi{\v{c}}ius, Qazi, Adedoyin-Olowe, and Misra]{abayomi2022data}
Olusola~O Abayomi-Alli, Robertas Dama{\v{s}}evi{\v{c}}ius, Atika Qazi, Mariam Adedoyin-Olowe, and Sanjay Misra.
\newblock Data augmentation and deep learning methods in sound classification: A systematic review.
\newblock \emph{Electronics}, 11\penalty0 (22):\penalty0 3795, 2022.

\bibitem[Liu et~al.(2020)Liu, Wang, Xiang, and Meng]{liu2020survey}
Pei Liu, Xuemin Wang, Chao Xiang, and Weiye Meng.
\newblock A survey of text data augmentation.
\newblock In \emph{2020 International Conference on Computer Communication and Network Security (CCNS)}, pages 191--195. IEEE, 2020.

\bibitem[Cauli and Reforgiato~Recupero(2022)]{cauli2022survey}
Nino Cauli and Diego Reforgiato~Recupero.
\newblock Survey on videos data augmentation for deep learning models.
\newblock \emph{Future Internet}, 14\penalty0 (3):\penalty0 93, 2022.

\bibitem[Bayer et~al.(2022)Bayer, Kaufhold, and Reuter]{bayer2022survey}
Markus Bayer, Marc-Andr{\'e} Kaufhold, and Christian Reuter.
\newblock A survey on data augmentation for text classification.
\newblock \emph{ACM Computing Surveys}, 55\penalty0 (7):\penalty0 1--39, 2022.

\bibitem[Kumar et~al.(2024)Kumar, Brennan, Mileo, and Bendechache]{kumar2024image}
Teerath Kumar, Rob Brennan, Alessandra Mileo, and Malika Bendechache.
\newblock Image data augmentation approaches: A comprehensive survey and future directions.
\newblock \emph{IEEE Access}, 2024.

\bibitem[Fayaz et~al.(2024)Fayaz, Ahmad~Shah, ud~din, Gul, and Assad]{fayaz2024advancements}
Salma Fayaz, Syed~Zubair Ahmad~Shah, Nusrat~Mohi ud~din, Naillah Gul, and Assif Assad.
\newblock Advancements in data augmentation and transfer learning: A comprehensive survey to address data scarcity challenges.
\newblock \emph{Recent Advances in Computer Science and Communications (Formerly: Recent Patents on Computer Science)}, 17\penalty0 (8):\penalty0 14--35, 2024.

\bibitem[Mumuni et~al.(2024)Mumuni, Mumuni, and Gerrar]{mumuni2024survey}
Alhassan Mumuni, Fuseini Mumuni, and Nana~Kobina Gerrar.
\newblock A survey of synthetic data augmentation methods in machine vision.
\newblock \emph{Machine Intelligence Research}, pages 1--39, 2024.

\bibitem[Sheng et~al.(2018)Sheng, Yang, Hu, Tan, and Qian]{sheng2018data}
Peiyao Sheng, Zhuolin Yang, Hu~Hu, Tian Tan, and Yanmin Qian.
\newblock Data augmentation using conditional generative adversarial networks for robust speech recognition.
\newblock In \emph{2018 11th international symposium on Chinese spoken language processing (ISCSLP)}, pages 121--125. IEEE, 2018.

\bibitem[Qian et~al.(2019)Qian, Hu, and Tan]{qian2019data}
Yanmin Qian, Hu~Hu, and Tian Tan.
\newblock Data augmentation using generative adversarial networks for robust speech recognition.
\newblock \emph{Speech Communication}, 114:\penalty0 1--9, 2019.

\bibitem[Wali et~al.(2022)Wali, Alamgir, Karim, Fawaz, Ali, Adan, and Mujtaba]{wali2022generative}
Aamir Wali, Zareen Alamgir, Saira Karim, Ather Fawaz, Mubariz~Barkat Ali, Muhammad Adan, and Malik Mujtaba.
\newblock Generative adversarial networks for speech processing: A review.
\newblock \emph{Computer Speech \& Language}, 72:\penalty0 101308, 2022.

\bibitem[Carrera-Rivera et~al.(2022)Carrera-Rivera, Ochoa, Larrinaga, and Lasa]{carrera2022conduct}
Angela Carrera-Rivera, William Ochoa, Felix Larrinaga, and Ganix Lasa.
\newblock How-to conduct a systematic literature review: A quick guide for computer science research.
\newblock \emph{MethodsX}, 9:\penalty0 101895, 2022.

\bibitem[{National Institutes of Health}(n.d.)]{nihguidingprinciples}
{National Institutes of Health}.
\newblock Guiding principles for ethical research, n.d.
\newblock URL \url{https://www.nih.gov/health-information/nih-clinical-research-trials-you/guiding-principles-ethical-research}.
\newblock Accessed: 2025-01-25.

\bibitem[Joshi and Sivaswamy(2008)]{joshi2008colour}
Gopal~Datt Joshi and Jayanthi Sivaswamy.
\newblock Colour retinal image enhancement based on domain knowledge.
\newblock In \emph{2008 Sixth Indian Conference on Computer Vision, Graphics \& Image Processing}, pages 591--598. IEEE, 2008.

\bibitem[Starck et~al.(2003)Starck, Murtagh, Cand{\`e}s, and Donoho]{starck2003gray}
J-L Starck, Fionn Murtagh, Emmanuel~J Cand{\`e}s, and David~L Donoho.
\newblock Gray and color image contrast enhancement by the curvelet transform.
\newblock \emph{IEEE Transactions on image processing}, 12\penalty0 (6):\penalty0 706--717, 2003.

\bibitem[Frei(1977)]{frei1977image}
Werner Frei.
\newblock Image enhancement by histogram hyperbolization.
\newblock \emph{Computer Graphics and Image Processing}, 6\penalty0 (3):\penalty0 286--294, 1977.

\bibitem[Lee(1980)]{lee1980digital}
Jong-Sen Lee.
\newblock Digital image enhancement and noise filtering by use of local statistics.
\newblock \emph{IEEE transactions on pattern analysis and machine intelligence}, \penalty0 (2):\penalty0 165--168, 1980.

\bibitem[Ketcham(1976)]{ketcham1976real}
David~J Ketcham.
\newblock Real-time image enhancement techniques.
\newblock In \emph{Image processing}, volume~74, pages 120--125. SPIE, 1976.

\bibitem[Osher and Rudin(1990)]{osher1990feature}
Stanley Osher and Leonid~I Rudin.
\newblock Feature-oriented image enhancement using shock filters.
\newblock \emph{SIAM Journal on numerical analysis}, 27\penalty0 (4):\penalty0 919--940, 1990.

\bibitem[Pal and King(1980)]{pal1980image}
Sankar~K Pal and Robert~A King.
\newblock Image enhancement using fuzzy set.
\newblock \emph{Electronics letters}, 16\penalty0 (10):\penalty0 376--378, 1980.

\bibitem[Park et~al.(2003)Park, Park, and Kang]{park2003super}
Sung~Cheol Park, Min~Kyu Park, and Moon~Gi Kang.
\newblock Super-resolution image reconstruction: a technical overview.
\newblock \emph{IEEE signal processing magazine}, 20\penalty0 (3):\penalty0 21--36, 2003.

\bibitem[Gibson et~al.(2002)Gibson, Cook, Howard, Hubbold, and Oram]{gibson2002accurate}
Simon Gibson, Jonathan Cook, Toby Howard, Roger Hubbold, and Daniel Oram.
\newblock Accurate camera calibration for off-line, video-based augmented reality.
\newblock In \emph{Proceedings. International Symposium on Mixed and Augmented Reality}, pages 37--46. IEEE, 2002.

\bibitem[Varma and Zisserman(2008)]{varma2008statistical}
Manik Varma and Andrew Zisserman.
\newblock A statistical approach to material classification using image patch exemplars.
\newblock \emph{IEEE transactions on pattern analysis and machine intelligence}, 31\penalty0 (11):\penalty0 2032--2047, 2008.

\bibitem[Rubio et~al.(2006)Rubio, Quintana, P{\'e}rez-Ros{\'e}s, Quir{\'o}s, and Camahort]{rubio2006jittering}
Monica Rubio, Arturo Quintana, Hebert P{\'e}rez-Ros{\'e}s, Ricardo Quir{\'o}s, and Emilio Camahort.
\newblock Jittering reduction in marker-based augmented reality systems.
\newblock In \emph{Computational Science and Its Applications-ICCSA 2006: International Conference, Glasgow, UK, May 8-11, 2006. Proceedings, Part I 6}, pages 510--517. Springer, 2006.

\bibitem[Aghagolzadeh and Ersoy(1992)]{aghagolzadeh1992transform}
Sabzali Aghagolzadeh and Okan~K Ersoy.
\newblock Transform image enhancement.
\newblock \emph{Optical Engineering}, 31\penalty0 (3):\penalty0 614--626, 1992.

\bibitem[Goljan and Fridrich(2008)]{goljan2008camera}
Miroslav Goljan and Jessica Fridrich.
\newblock Camera identification from cropped and scaled images.
\newblock In \emph{Security, Forensics, Steganography, and Watermarking of Multimedia Contents X}, volume 6819, pages 154--166. SPIE, 2008.

\bibitem[Greenspan et~al.(2000)Greenspan, Anderson, and Akber]{greenspan2000image}
Hayit Greenspan, Charles~H Anderson, and Sofia Akber.
\newblock Image enhancement by nonlinear extrapolation in frequency space.
\newblock \emph{IEEE Transactions on Image Processing}, 9\penalty0 (6):\penalty0 1035--1048, 2000.

\bibitem[Zhu et~al.(2007)Zhu, Goldberg, Eldawy, Dyer, and Strock]{zhu2007text}
Xiaojin Zhu, Andrew~B Goldberg, Mohamed Eldawy, Charles~R Dyer, and Bradley Strock.
\newblock A text-to-picture synthesis system for augmenting communication.
\newblock In \emph{AAAI}, volume~7, pages 1590--1595, 2007.

\bibitem[Edmonds(1999)]{edmonds1999semantic}
Philip~Glenny Edmonds.
\newblock \emph{Semantic representations of near-synonyms for automatic lexical choice.}
\newblock University of Toronto, 1999.

\bibitem[Barnard and Johnson(2005)]{barnard2005word}
Kobus Barnard and Matthew Johnson.
\newblock Word sense disambiguation with pictures.
\newblock \emph{Artificial Intelligence}, 167\penalty0 (1-2):\penalty0 13--30, 2005.

\bibitem[Boulis and Ostendorf(2005)]{boulis2005text}
Constantinos Boulis and Mari Ostendorf.
\newblock Text classification by augmenting the bag-of-words representation with redundancy-compensated bigrams.
\newblock In \emph{Proc. of the International Workshop in Feature Selection in Data Mining}, pages 9--16. Citeseer, 2005.

\bibitem[Ben-Naim and Krapivsky(2007)]{ben2007addition}
E~Ben-Naim and PL~Krapivsky.
\newblock Addition--deletion networks.
\newblock \emph{Journal of Physics A: Mathematical and theoretical}, 40\penalty0 (30):\penalty0 8607, 2007.

\bibitem[McNamara et~al.(1996)McNamara, Kintsch, Songer, and Kintsch]{mcnamara1996good}
Danielle~S McNamara, Eileen Kintsch, Nancy~Butler Songer, and Walter Kintsch.
\newblock Are good texts always better? interactions of text coherence, background knowledge, and levels of understanding in learning from text.
\newblock \emph{Cognition and instruction}, 14\penalty0 (1):\penalty0 1--43, 1996.

\bibitem[Nakagawa and Freckleton(2008)]{nakagawa2008missing}
Shinichi Nakagawa and Robert~P Freckleton.
\newblock Missing inaction: the dangers of ignoring missing data.
\newblock \emph{Trends in ecology \& evolution}, 23\penalty0 (11):\penalty0 592--596, 2008.

\bibitem[Kukich(1992)]{kukich1992techniques}
Karen Kukich.
\newblock Techniques for automatically correcting words in text.
\newblock \emph{ACM computing surveys (CSUR)}, 24\penalty0 (4):\penalty0 377--439, 1992.

\bibitem[McRoy et~al.(2003)McRoy, Channarukul, and Ali]{mcroy2003augmented}
Susan~W McRoy, Songsak Channarukul, and Syed~S Ali.
\newblock An augmented template-based approach to text realization.
\newblock \emph{Natural Language Engineering}, 9\penalty0 (4):\penalty0 381--420, 2003.

\bibitem[Haritaoglu(2001)]{haritaoglu2001scene}
Ismail Haritaoglu.
\newblock Scene text extraction and translation for handheld devices.
\newblock In \emph{Proceedings of the 2001 IEEE Computer Society Conference on Computer Vision and Pattern Recognition. CVPR 2001}, volume~2, pages II--II. IEEE, 2001.

\bibitem[Yan et~al.(2008)Yan, Grosky, and Fotouhi]{yan2008augmenting}
Hua Yan, William~I Grosky, and Farshad Fotouhi.
\newblock Augmenting the power of lsi in text retrieval: Singular value rescaling.
\newblock \emph{Data \& Knowledge Engineering}, 65\penalty0 (1):\penalty0 108--125, 2008.

\bibitem[Kay(1997)]{kay1997proper}
Martin Kay.
\newblock The proper place of men and machines in language translation.
\newblock \emph{machine translation}, 12:\penalty0 3--23, 1997.

\bibitem[Kraaij et~al.(2003)Kraaij, Nie, and Simard]{kraaij2003embedding}
Wessel Kraaij, Jian-Yun Nie, and Michel Simard.
\newblock Embedding web-based statistical translation models in cross-language information retrieval.
\newblock \emph{Computational Linguistics}, 29\penalty0 (3):\penalty0 381--419, 2003.

\bibitem[Fung and McKeown(1997)]{fung1997technical}
Pascale Fung and Kathleen McKeown.
\newblock A technical word-and term-translation aid using noisy parallel corpora across language groups.
\newblock \emph{Machine translation}, 12:\penalty0 53--87, 1997.

\bibitem[Dhillon et~al.(2002)Dhillon, Guan, and Kogan]{dhillon2002iterative}
Inderjit~S Dhillon, Yuqiang Guan, and Jacob Kogan.
\newblock Iterative clustering of high dimensional text data augmented by local search.
\newblock In \emph{2002 IEEE International Conference on Data Mining, 2002. Proceedings.}, pages 131--138. IEEE, 2002.

\bibitem[Cohen et~al.(1993)Cohen, Aoki, and Koizumi]{cohen1993augmented}
Michael Cohen, Shigeaki Aoki, and Nobuo Koizumi.
\newblock Augmented audio reality: Telepresence/vr hybrid acoustic environments.
\newblock In \emph{Proceedings of 1993 2nd IEEE International Workshop on Robot and Human Communication}, pages 361--364. IEEE, 1993.

\bibitem[Brandenburg et~al.(1992)Brandenburg, Kirkham, and Koschkee]{brandenburg1992vocal}
James~H Brandenburg, Wayne Kirkham, and Danna Koschkee.
\newblock Vocal cord augmentation with autogenous fat.
\newblock \emph{The Laryngoscope}, 102\penalty0 (5):\penalty0 495--500, 1992.

\bibitem[Watanabe et~al.(1985)Watanabe, Murakami, Ishikawa, and Kamae]{watanabe1985audio}
Kazuhisa Watanabe, S~Murakami, HIROSHI Ishikawa, and TAKAHlKO Kamae.
\newblock Audio and visually augmented teleconferencing.
\newblock \emph{Proceedings of the IEEE}, 73\penalty0 (4):\penalty0 656--670, 1985.

\bibitem[Schmandt et~al.(1990)Schmandt, Ackerman, and Hindus]{schmandt1990augmenting}
Chris Schmandt, Mark~S. Ackerman, and Debby Hindus.
\newblock Augmenting a window system with speech input.
\newblock \emph{Computer}, 23\penalty0 (8):\penalty0 50--56, 1990.

\bibitem[Adams and Lang(1992)]{adams1992can}
Scott~G Adams and Anthony~E Lang.
\newblock Can the lombard effect be used to improve low voice intensity in parkinson's disease?
\newblock \emph{European Journal of Disorders of Communication}, 27\penalty0 (2):\penalty0 121--127, 1992.

\bibitem[van Hoesel and Tyler(2003)]{van2003speech}
Richard~JM van Hoesel and Richard~S Tyler.
\newblock Speech perception, localization, and lateralization with bilateral cochlear implants.
\newblock \emph{The Journal of the Acoustical Society of America}, 113\penalty0 (3):\penalty0 1617--1630, 2003.

\bibitem[Won et~al.(2008)Won, Schimmel, Drennan, Souza, Atlas, and Rubinstein]{won2008improving}
Jong~Ho Won, Steven~M Schimmel, Ward~R Drennan, Pamela~E Souza, Les Atlas, and Jay~T Rubinstein.
\newblock Improving performance in noise for hearing aids and cochlear implants using coherent modulation filtering.
\newblock \emph{Hearing research}, 239\penalty0 (1-2):\penalty0 1--11, 2008.

\bibitem[Nishimura et~al.(2006)Nishimura, Ishizuka, Nakadai, Nakano, and Tsujino]{nishimura2006speech}
Yoshitaka Nishimura, Mitsuru Ishizuka, Kazuhiro Nakadai, Mikio Nakano, and Hiroshi Tsujino.
\newblock Speech recognition for a humanoid with motor noise utilizing missing feature theory.
\newblock In \emph{2006 6th IEEE-RAS International Conference on Humanoid Robots}, pages 26--33. IEEE, 2006.

\bibitem[H{\"a}rm{\"a} et~al.(2004)H{\"a}rm{\"a}, Jakka, Tikander, Karjalainen, Lokki, Hiipakka, and Lorho]{harma2004augmented}
Aki H{\"a}rm{\"a}, Julia Jakka, Miikka Tikander, Matti Karjalainen, Tapio Lokki, Jarmo Hiipakka, and Ga{\"e}tan Lorho.
\newblock Augmented reality audio for mobile and wearable appliances.
\newblock \emph{Journal of the Audio Engineering Society}, 52\penalty0 (6):\penalty0 618--639, 2004.

\bibitem[Shu and Chang(2008)]{shu2008power}
Wei Shu and Joseph~S Chang.
\newblock Power supply noise in analog audio class d amplifiers.
\newblock \emph{IEEE Transactions on Circuits and Systems I: Regular Papers}, 56\penalty0 (1):\penalty0 84--96, 2008.

\bibitem[Yadav and Jadhav(2019)]{yadav2019deep}
Samir~S Yadav and Shivajirao~M Jadhav.
\newblock Deep convolutional neural network based medical image classification for disease diagnosis.
\newblock \emph{Journal of Big data}, 6\penalty0 (1):\penalty0 1--18, 2019.

\bibitem[Altaf et~al.(2019)Altaf, Islam, Akhtar, and Janjua]{altaf2019going}
Fouzia Altaf, Syed~MS Islam, Naveed Akhtar, and Naeem~Khalid Janjua.
\newblock Going deep in medical image analysis: concepts, methods, challenges, and future directions.
\newblock \emph{IEEE Access}, 7:\penalty0 99540--99572, 2019.

\bibitem[Shu et~al.(2021)Shu, Shen, Lin, and Goldstein]{shu2021adversarial}
Manli Shu, Yu~Shen, Ming~C Lin, and Tom Goldstein.
\newblock Adversarial differentiable data augmentation for autonomous systems.
\newblock In \emph{2021 IEEE International Conference on Robotics and Automation (ICRA)}, pages 14069--14075. IEEE, 2021.

\bibitem[Su et~al.(2021)Su, Kong, Qiao, and Sukkarieh]{su2021data}
Daobilige Su, He~Kong, Yongliang Qiao, and Salah Sukkarieh.
\newblock Data augmentation for deep learning based semantic segmentation and crop-weed classification in agricultural robotics.
\newblock \emph{Computers and Electronics in Agriculture}, 190:\penalty0 106418, 2021.

\bibitem[Jo et~al.(2022)Jo, Heo, Park, Yoo, Cho, and Kim]{jo2022dagam}
Byeong-Cheol Jo, Tak-Sung Heo, Yeongjoon Park, Yongmin Yoo, Won~Ik Cho, and Kyungsun Kim.
\newblock Dagam: data augmentation with generation and modification.
\newblock \emph{arXiv preprint arXiv:2204.02633}, 2022.

\bibitem[Sapkota and Karkee(2023)]{sapkota2023creating}
Ranjan Sapkota and Manoj Karkee.
\newblock Generative ai in agriculture: Creating image datasets using dall.e's advanced large language model capabilities.
\newblock \emph{arXiv preprint arXiv:2307.08789}, 2023.

\bibitem[Hernandez et~al.(2022)Hernandez, Epelde, Alberdi, Cilla, and Rankin]{hernandez2022synthetic}
Mikel Hernandez, Gorka Epelde, Ane Alberdi, Rodrigo Cilla, and Debbie Rankin.
\newblock Synthetic data generation for tabular health records: A systematic review.
\newblock \emph{Neurocomputing}, 493:\penalty0 28--45, 2022.

\bibitem[Garcea et~al.(2023)Garcea, Serra, Lamberti, and Morra]{garcea2023data}
Fabio Garcea, Alessio Serra, Fabrizio Lamberti, and Lia Morra.
\newblock Data augmentation for medical imaging: A systematic literature review.
\newblock \emph{Computers in Biology and Medicine}, 152:\penalty0 106391, 2023.

\bibitem[Sapkota et~al.(2024{\natexlab{a}})Sapkota, Qureshi, Hassan, Shutske, Shoman, Sajjad, Dharejo, Paudel, Li, Meng, et~al.]{sapkota2024multi}
Ranjan Sapkota, Rizwan Qureshi, Syed~Zohaib Hassan, John Shutske, Maged Shoman, Muhammad Sajjad, Fayaz~Ali Dharejo, Achyut Paudel, Jiajia Li, Zhichao Meng, et~al.
\newblock Multi-modal llms in agriculture: A comprehensive review.
\newblock \emph{Authorea Preprints}, 2024{\natexlab{a}}.

\bibitem[Shijie et~al.(2017)Shijie, Ping, Peiyi, and Siping]{shijie2017research}
Jia Shijie, Wang Ping, Jia Peiyi, and Hu~Siping.
\newblock Research on data augmentation for image classification based on convolution neural networks.
\newblock In \emph{2017 Chinese automation congress (CAC)}, pages 4165--4170. IEEE, 2017.

\bibitem[Vyas et~al.(2024)Vyas, Ragothaman, Chauhan, and Rimal]{vyas2024data}
Piyush Vyas, Kaushik~Muthusamy Ragothaman, Akhilesh Chauhan, and Bhaskar Rimal.
\newblock Data augmentation and generative machine learning on the cloud platform.
\newblock \emph{International Journal of Information Technology}, 16\penalty0 (8):\penalty0 4833--4843, 2024.

\bibitem[Wang et~al.(2020{\natexlab{b}})Wang, Yang, Wu, Qian, and Yu]{wang2020data}
Shuai Wang, Yexin Yang, Zhanghao Wu, Yanmin Qian, and Kai Yu.
\newblock Data augmentation using deep generative models for embedding based speaker recognition.
\newblock \emph{IEEE/ACM Transactions on Audio, Speech, and Language Processing}, 28:\penalty0 2598--2609, 2020{\natexlab{b}}.

\bibitem[Fung and Wu(1994)]{fung1994statistical}
Pascale Fung and Dekai Wu.
\newblock Statistical augmentation of a chinese machine-readable dictionary.
\newblock \emph{arXiv preprint cmp-lg/9406015}, 1994.

\bibitem[Dietterich et~al.(1995)Dietterich, Hild, and Bakiri]{dietterich1995comparison}
Thomas~G Dietterich, Hermann Hild, and Ghulum Bakiri.
\newblock A comparison of id3 and backpropagation for english text-to-speech mapping.
\newblock \emph{Machine Learning}, 18:\penalty0 51--80, 1995.

\bibitem[Rubin(1987)]{rubin1987comment}
Donald~B Rubin.
\newblock Comment: A noniterative sampling/importance resampling alternative to the data augmentation algorithm for creating a few imputations when fractions of missing information are modest: The sir algorithm.
\newblock \emph{Journal of the American Statistical Association}, 82\penalty0 (398):\penalty0 542--543, 1987.

\bibitem[Onan(2023)]{onan2023srl}
Aytu{\u{g}} Onan.
\newblock Srl-aco: A text augmentation framework based on semantic role labeling and ant colony optimization.
\newblock \emph{Journal of King Saud University-Computer and Information Sciences}, 35\penalty0 (7):\penalty0 101611, 2023.

\bibitem[Troiano et~al.(2023)Troiano, Velutharambath, and Klinger]{troiano2023theories}
Enrica Troiano, Aswathy Velutharambath, and Roman Klinger.
\newblock From theories on styles to their transfer in text: Bridging the gap with a hierarchical survey.
\newblock \emph{Natural Language Engineering}, 29\penalty0 (4):\penalty0 849--908, 2023.

\bibitem[Liu et~al.(2023{\natexlab{a}})Liu, Han, Ma, Zhang, Yang, Tian, He, Li, He, Liu, et~al.]{liu2023summary}
Yiheng Liu, Tianle Han, Siyuan Ma, Jiayue Zhang, Yuanyuan Yang, Jiaming Tian, Hao He, Antong Li, Mengshen He, Zhengliang Liu, et~al.
\newblock Summary of chatgpt-related research and perspective towards the future of large language models.
\newblock \emph{Meta-Radiology}, page 100017, 2023{\natexlab{a}}.

\bibitem[Alam et~al.(2020)Alam, Samad, Vidyaratne, Glandon, and Iftekharuddin]{alam2020survey}
Mahbubul Alam, Manar~D Samad, Lasitha Vidyaratne, Alexander Glandon, and Khan~M Iftekharuddin.
\newblock Survey on deep neural networks in speech and vision systems.
\newblock \emph{Neurocomputing}, 417:\penalty0 302--321, 2020.

\bibitem[Khan et~al.(2023)Khan, Daud, Khan, Muhammad, and Haq]{khan2023exploring}
Wahab Khan, Ali Daud, Khairullah Khan, Shakoor Muhammad, and Rafiul Haq.
\newblock Exploring the frontiers of deep learning and natural language processing: A comprehensive overview of key challenges and emerging trends.
\newblock \emph{Natural Language Processing Journal}, page 100026, 2023.

\bibitem[Braik and Koliou(2024)]{braik2024automated}
Abdullah~M Braik and Maria Koliou.
\newblock Automated building damage assessment and large-scale mapping by integrating satellite imagery, gis, and deep learning.
\newblock \emph{Computer-Aided Civil and Infrastructure Engineering}, 2024.

\bibitem[Zuo et~al.(2024)Zuo, Niu, Li, Fu, and Zhou]{zuo2024machine}
Zitu Zuo, Yongjie Niu, Jiale Li, Hongpeng Fu, and Mengjie Zhou.
\newblock Machine learning for advanced emission monitoring and reduction strategies in fossil fuel power plants.
\newblock \emph{Applied Sciences}, 14\penalty0 (18):\penalty0 8442, 2024.

\bibitem[Raiaan et~al.(2024)Raiaan, Mukta, Fatema, Fahad, Sakib, Mim, Ahmad, Ali, and Azam]{raiaan2024review}
Mohaimenul Azam~Khan Raiaan, Md~Saddam~Hossain Mukta, Kaniz Fatema, Nur~Mohammad Fahad, Sadman Sakib, Most Marufatul~Jannat Mim, Jubaer Ahmad, Mohammed~Eunus Ali, and Sami Azam.
\newblock A review on large language models: Architectures, applications, taxonomies, open issues and challenges.
\newblock \emph{IEEE Access}, 2024.

\bibitem[Wei et~al.(2020)Wei, Zou, Liao, et~al.]{wei2020comparison}
Shengyun Wei, Shun Zou, Feifan Liao, et~al.
\newblock A comparison on data augmentation methods based on deep learning for audio classification.
\newblock In \emph{Journal of physics: Conference series}, volume 1453, page 012085. IOP Publishing, 2020.

\bibitem[Cohen et~al.(2022)Cohen, Rimon, Aflalo, and Permuter]{cohen2022study}
Ariel Cohen, Inbal Rimon, Eran Aflalo, and Haim~H Permuter.
\newblock A study on data augmentation in voice anti-spoofing.
\newblock \emph{Speech Communication}, 141:\penalty0 56--67, 2022.

\bibitem[Zhou et~al.(2022)Zhou, Chen, and Chien]{zhou2022analysis}
George Zhou, Yunchan Chen, and Candace Chien.
\newblock On the analysis of data augmentation methods for spectral imaged based heart sound classification using convolutional neural networks.
\newblock \emph{BMC medical informatics and decision making}, 22\penalty0 (1):\penalty0 226, 2022.

\bibitem[Rayavarapu et~al.(2024)Rayavarapu, Prasanthi, Gottapu, and Singam]{rayavarapu2024comprehensive}
Swarajya~Madhuri Rayavarapu, Tammineni~Shanmukha Prasanthi, Sasibhushana~Rao Gottapu, and Aruna Singam.
\newblock A comprehensive overview on data augmentation techniques for medical images.
\newblock In \emph{2024 5th International Conference on Electronics and Sustainable Communication Systems (ICESC)}, pages 1324--1329. IEEE, 2024.

\bibitem[Li et~al.(2024{\natexlab{a}})Li, Chen, Chen, Li, He, and Yingbiao]{li2024itimca}
Huinian Li, Baoyu Chen, Jingjia Chen, Shuting Li, Feiyong He, and Hu~Yingbiao.
\newblock Itimca: Image-text information and cross-attention for multi-modal cassava leaf disease classification based on a novel multi-modal dataset in natural environments.
\newblock \emph{Crop Protection}, page 106981, 2024{\natexlab{a}}.

\bibitem[Sapkota et~al.(2024{\natexlab{b}})Sapkota, Meng, and Karkee]{sapkota2024synthetic}
Ranjan Sapkota, Zhichao Meng, and Manoj Karkee.
\newblock Synthetic meets authentic: Leveraging llm generated datasets for yolo11 and yolov10-based apple detection through machine vision sensors.
\newblock \emph{Smart Agricultural Technology}, page 100614, 2024{\natexlab{b}}.

\bibitem[Liu et~al.(2024{\natexlab{a}})Liu, Zhu, Gu, Pan, Li, Fan, Li, and Zeng]{liu2024enhanced}
Yang Liu, Yiqi Zhu, Zhehao Gu, Jinshan Pan, Juncheng Li, Ming Fan, Lihua Li, and Tieyong Zeng.
\newblock Enhanced dual contrast representation learning with cell separation and merging for breast cancer diagnosis.
\newblock \emph{Computer Vision and Image Understanding}, 247:\penalty0 104065, 2024{\natexlab{a}}.

\bibitem[Kirilenko et~al.(2024)Kirilenko, Andreychuk, Panov, and Yakovlev]{kirilenko2024generative}
Daniil Kirilenko, Anton Andreychuk, Aleksandr~I Panov, and Konstantin Yakovlev.
\newblock Generative models for grid-based and image-based pathfinding.
\newblock \emph{Artificial Intelligence}, page 104238, 2024.

\bibitem[Li et~al.(2024{\natexlab{b}})Li, Guan, Wang, Cheung, Zheng, Lim, Lim, Ruamviboonsuk, Raman, Corsino, et~al.]{li2024integrated}
Jiajia Li, Zhouyu Guan, Jing Wang, Carol~Y Cheung, Yingfeng Zheng, Lee-Ling Lim, Cynthia~Ciwei Lim, Paisan Ruamviboonsuk, Rajiv Raman, Leonor Corsino, et~al.
\newblock Integrated image-based deep learning and language models for primary diabetes care.
\newblock \emph{Nature medicine}, pages 1--11, 2024{\natexlab{b}}.

\bibitem[Liu et~al.(2023{\natexlab{b}})Liu, Zhu, Wu, Yang, You, Wang, Lu, Liu, Zheng, Sun, et~al.]{liu2023medical}
Fenglin Liu, Tingting Zhu, Xian Wu, Bang Yang, Chenyu You, Chenyang Wang, Lei Lu, Zhangdaihong Liu, Yefeng Zheng, Xu~Sun, et~al.
\newblock A medical multimodal large language model for future pandemics.
\newblock \emph{NPJ Digital Medicine}, 6\penalty0 (1):\penalty0 226, 2023{\natexlab{b}}.

\bibitem[Jindal et~al.(2024)Jindal, Kumaresan, Ponnusamy, Thavareesan, Rajiakodi, and Chakravarthi]{jindal2024mistra}
Nitesh Jindal, Prasanna~Kumar Kumaresan, Rahul Ponnusamy, Sajeetha Thavareesan, Saranya Rajiakodi, and Bharathi~Raja Chakravarthi.
\newblock Mistra: Misogyny detection through text--image fusion and representation analysis.
\newblock \emph{Natural Language Processing Journal}, 7:\penalty0 100073, 2024.

\bibitem[Sapkota et~al.(2024{\natexlab{c}})Sapkota, Paudel, and Karkee]{sapkota2024zero}
Ranjan Sapkota, Achyut Paudel, and Manoj Karkee.
\newblock Zero-shot automatic annotation and instance segmentation using llm-generated datasets: Eliminating field imaging and manual annotation for deep learning model development.
\newblock \emph{arXiv preprint arXiv:2411.11285}, 2024{\natexlab{c}}.

\bibitem[Sapkota and Karkee(2025)]{sapkota2025improved}
Ranjan Sapkota and Manoj Karkee.
\newblock Improved yolov12 with llm-generated synthetic data for enhanced apple detection and benchmarking against yolov11 and yolov10.
\newblock \emph{arXiv preprint arXiv:2503.00057}, 2025.

\bibitem[Sapkota et~al.(2024{\natexlab{d}})Sapkota, Meng, Churuvija, Du, Ma, and Karkee]{sapkota2024comprehensive}
Ranjan Sapkota, Zhichao Meng, Martin Churuvija, Xiaoqiang Du, Zenghong Ma, and Manoj Karkee.
\newblock Comprehensive performance evaluation of yolov12, yolo11, yolov10, yolov9 and yolov8 on detecting and counting fruitlet in complex orchard environments.
\newblock \emph{arXiv preprint arXiv:2407.12040}, 2024{\natexlab{d}}.

\bibitem[Yuan et~al.(2023{\natexlab{a}})Yuan, Tang, Jiang, and Hu]{yuan2023large}
Jiayi Yuan, Ruixiang Tang, Xiaoqian Jiang, and Xia Hu.
\newblock Large language models for healthcare data augmentation: An example on patient-trial matching.
\newblock In \emph{AMIA Annual Symposium Proceedings}, volume 2023, page 1324. American Medical Informatics Association, 2023{\natexlab{a}}.

\bibitem[Cortacero et~al.(2023)Cortacero, McKenzie, M{\"u}ller, Khazen, Lafouresse, Corsaut, Van~Acker, Frenois, Lamant, Meyer, et~al.]{cortacero2023evolutionary}
K{\'e}vin Cortacero, Brienne McKenzie, Sabina M{\"u}ller, Roxana Khazen, Fanny Lafouresse, Ga{\"e}lle Corsaut, Nathalie Van~Acker, Fran{\c{c}}ois-Xavier Frenois, Laurence Lamant, Nicolas Meyer, et~al.
\newblock Evolutionary design of explainable algorithms for biomedical image segmentation.
\newblock \emph{Nature communications}, 14\penalty0 (1):\penalty0 7112, 2023.

\bibitem[Raminedi et~al.(2024)Raminedi, Shridevi, and Won]{raminedi2024multi}
Santhosh Raminedi, S~Shridevi, and Daehan Won.
\newblock Multi-modal transformer architecture for medical image analysis and automated report generation.
\newblock \emph{Scientific Reports}, 14\penalty0 (1):\penalty0 19281, 2024.

\bibitem[Wang et~al.(2024{\natexlab{a}})Wang, Shi, and Zhao]{wang2024mllm4rec}
Yuxiang Wang, Xin Shi, and Xueqing Zhao.
\newblock Mllm4rec: multimodal information enhancing llm for sequential recommendation.
\newblock \emph{Journal of Intelligent Information Systems}, pages 1--17, 2024{\natexlab{a}}.

\bibitem[Bețianu et~al.(2024)Bețianu, M{\u{a}}lan, Aldinucci, Birke, and Chen]{bețianu2024dallmi}
Miruna Bețianu, Abele M{\u{a}}lan, Marco Aldinucci, Robert Birke, and Lydia Chen.
\newblock Dallmi: Domain adaption for llm-based multi-label classifier.
\newblock In \emph{Pacific-Asia Conference on Knowledge Discovery and Data Mining}, pages 277--289. Springer, 2024.

\bibitem[Sheik et~al.(2024)Sheik, Siva~Sundara, and Nirmala]{sheik2024neural}
Reshma Sheik, KP~Siva~Sundara, and S~Jaya Nirmala.
\newblock Neural data augmentation for legal overruling task: Small deep learning models vs. large language models.
\newblock \emph{Neural Processing Letters}, 56\penalty0 (2):\penalty0 121, 2024.

\bibitem[Tao et~al.(2022)Tao, Tang, Wu, Jing, Bao, and Xu]{tao2022df}
Ming Tao, Hao Tang, Fei Wu, Xiao-Yuan Jing, Bing-Kun Bao, and Changsheng Xu.
\newblock Df-gan: A simple and effective baseline for text-to-image synthesis.
\newblock In \emph{Proceedings of the IEEE/CVF conference on computer vision and pattern recognition}, pages 16515--16525, 2022.

\bibitem[Liang et~al.(2024)Liang, He, Li, Li, Klimovskiy, Carolan, Sun, Pont-Tuset, Young, Yang, et~al.]{liang2024rich}
Youwei Liang, Junfeng He, Gang Li, Peizhao Li, Arseniy Klimovskiy, Nicholas Carolan, Jiao Sun, Jordi Pont-Tuset, Sarah Young, Feng Yang, et~al.
\newblock Rich human feedback for text-to-image generation.
\newblock In \emph{Proceedings of the IEEE/CVF Conference on Computer Vision and Pattern Recognition}, pages 19401--19411, 2024.

\bibitem[Phung et~al.(2024)Phung, Ge, and Huang]{phung2024grounded}
Quynh Phung, Songwei Ge, and Jia-Bin Huang.
\newblock Grounded text-to-image synthesis with attention refocusing.
\newblock In \emph{Proceedings of the IEEE/CVF Conference on Computer Vision and Pattern Recognition}, pages 7932--7942, 2024.

\bibitem[Heisler et~al.(2022)Heisler, Banitalebi-Dehkordi, and Zhang]{heisler2022semaug}
Morgan Heisler, Amin Banitalebi-Dehkordi, and Yong Zhang.
\newblock Semaug: Semantically meaningful image augmentations for object detection through language grounding.
\newblock In \emph{European Conference on Computer Vision}, pages 610--626. Springer, 2022.

\bibitem[Jung et~al.(2024)Jung, Seo, Cho, Kim, Min, and Choi]{jung2024dalda}
Kyuheon Jung, Yongdeuk Seo, Seongwoo Cho, Jaeyoung Kim, Hyun-seok Min, and Sungchul Choi.
\newblock Dalda: Data augmentation leveraging diffusion model and llm with adaptive guidance scaling.
\newblock \emph{arXiv preprint arXiv:2409.16949}, 2024.

\bibitem[Yu et~al.(2023)Yu, Xiao, Stone, Tompson, Brohan, Wang, Singh, Tan, Peralta, Ichter, et~al.]{yu2023scaling}
Tianhe Yu, Ted Xiao, Austin Stone, Jonathan Tompson, Anthony Brohan, Su~Wang, Jaspiar Singh, Clayton Tan, Jodilyn Peralta, Brian Ichter, et~al.
\newblock Scaling robot learning with semantically imagined experience.
\newblock \emph{arXiv preprint arXiv:2302.11550}, 2023.

\bibitem[Rotstein et~al.(2024)Rotstein, Bensa{\"\i}d, Brody, Ganz, and Kimmel]{rotstein2024fusecap}
Noam Rotstein, David Bensa{\"\i}d, Shaked Brody, Roy Ganz, and Ron Kimmel.
\newblock Fusecap: Leveraging large language models for enriched fused image captions.
\newblock In \emph{Proceedings of the IEEE/CVF Winter Conference on Applications of Computer Vision}, pages 5689--5700, 2024.

\bibitem[Yi et~al.(2024)Yi, Uzkent, Ignat, Li, Garg, Yu, and Liu]{yi2024augment}
Jingru Yi, Burak Uzkent, Oana Ignat, Zili Li, Amanmeet Garg, Xiang Yu, and Linda Liu.
\newblock Augment the pairs: Semantics-preserving image-caption pair augmentation for grounding-based vision and language models.
\newblock In \emph{Proceedings of the IEEE/CVF Winter Conference on Applications of Computer Vision}, pages 5520--5530, 2024.

\bibitem[Sharifzadeh et~al.(2024)Sharifzadeh, Kaplanis, Pathak, Kumaran, Ilic, Mitrovic, Blundell, and Banino]{sharifzadeh2024synth}
Sahand Sharifzadeh, Christos Kaplanis, Shreya Pathak, Dharshan Kumaran, Anastasija Ilic, Jovana Mitrovic, Charles Blundell, and Andrea Banino.
\newblock Synth2: Boosting visual-language models with synthetic captions and image embeddings.
\newblock \emph{arXiv preprint arXiv:2403.07750}, 2024.

\bibitem[Koh et~al.(2024)Koh, Fried, and Salakhutdinov]{koh2024generating}
Jing~Yu Koh, Daniel Fried, and Russ~R Salakhutdinov.
\newblock Generating images with multimodal language models.
\newblock \emph{Advances in Neural Information Processing Systems}, 36, 2024.

\bibitem[Li et~al.(2024{\natexlab{c}})Li, Xu, Liu, and Xiao]{li2024unimo}
Wei Li, Xue Xu, Jiachen Liu, and Xinyan Xiao.
\newblock Unimo-g: Unified image generation through multimodal conditional diffusion.
\newblock \emph{arXiv preprint arXiv:2401.13388}, 2024{\natexlab{c}}.

\bibitem[Sultan et~al.(2024)Sultan, Khasin, Shiran, Greenstein-Messica, and Shahaf]{sultan2024visual}
Oren Sultan, Alex Khasin, Guy Shiran, Asnat Greenstein-Messica, and Dafna Shahaf.
\newblock Visual editing with llm-based tool chaining: An efficient distillation approach for real-time applications.
\newblock \emph{arXiv preprint arXiv:2410.02952}, 2024.

\bibitem[Li et~al.(2024{\natexlab{d}})Li, Zhang, Zhu, Sun, Zhang, and Zha]{li2024forgerygpt}
Jiawei Li, Fanrui Zhang, Jiaying Zhu, Esther Sun, Qiang Zhang, and Zheng-Jun Zha.
\newblock Forgerygpt: Multimodal large language model for explainable image forgery detection and localization.
\newblock \emph{arXiv preprint arXiv:2410.10238}, 2024{\natexlab{d}}.

\bibitem[Wu et~al.(2024{\natexlab{a}})Wu, Qiu, Song, Chen, Huang, Ma, and Xiao]{wu2024image}
Wangyu Wu, Xianglin Qiu, Siqi Song, Zhenhong Chen, Xiaowei Huang, Fei Ma, and Jimin Xiao.
\newblock Image augmentation agent for weakly supervised semantic segmentation.
\newblock \emph{arXiv preprint arXiv:2412.20439}, 2024{\natexlab{a}}.

\bibitem[Song et~al.(2024{\natexlab{a}})Song, Subramanyam, Madejski, and Grossman]{song2024lab}
Steven Song, Anirudh Subramanyam, Irene Madejski, and Robert~L Grossman.
\newblock Lab-rag: Label boosted retrieval augmented generation for radiology report generation.
\newblock \emph{arXiv preprint arXiv:2411.16523}, 2024{\natexlab{a}}.

\bibitem[Lingenberg et~al.(2024)Lingenberg, Reuter, Sudhakaran, Gojny, Roth, and Schaub-Meyer]{lingenberg2024diagen}
Tobias Lingenberg, Markus Reuter, Gopika Sudhakaran, Dominik Gojny, Stefan Roth, and Simone Schaub-Meyer.
\newblock Diagen: Diverse image augmentation with generative models.
\newblock \emph{arXiv preprint arXiv:2408.14584}, 2024.

\bibitem[Yin et~al.(2024{\natexlab{a}})Yin, Fu, Zhao, Shen, Ge, Yang, Long, Dai, Xu, Sun, et~al.]{yin2024t2vid}
Shukang Yin, Chaoyou Fu, Sirui Zhao, Yunhang Shen, Chunjiang Ge, Yan Yang, Zuwei Long, Yuhan Dai, Tong Xu, Xing Sun, et~al.
\newblock T2vid: Translating long text into multi-image is the catalyst for video-llms.
\newblock \emph{arXiv preprint arXiv:2411.19951}, 2024{\natexlab{a}}.

\bibitem[Liu et~al.(2024{\natexlab{b}})Liu, Huang, Zheng, Liu, Wang, Yoshie, Liu, and Li]{liu2024mm}
Jihao Liu, Xin Huang, Jinliang Zheng, Boxiao Liu, Jia Wang, Osamu Yoshie, Yu~Liu, and Hongsheng Li.
\newblock Mm-instruct: Generated visual instructions for large multimodal model alignment.
\newblock \emph{arXiv preprint arXiv:2406.19736}, 2024{\natexlab{b}}.

\bibitem[Hsieh et~al.(2024)Hsieh, Moreira, Nobre, Sousa, Ouyang, Brereton, Jorge, and Nascimento]{hsieh2024dall}
Chihcheng Hsieh, Catarina Moreira, Isabel~Blanco Nobre, Sandra~Costa Sousa, Chun Ouyang, Margot Brereton, Joaquim Jorge, and Jacinto~C Nascimento.
\newblock Dall-m: Context-aware clinical data augmentation with llms.
\newblock \emph{arXiv preprint arXiv:2407.08227}, 2024.

\bibitem[Zang et~al.(2024)Zang, Li, Han, Zhou, and Loy]{zang2024contextual}
Yuhang Zang, Wei Li, Jun Han, Kaiyang Zhou, and Chen~Change Loy.
\newblock Contextual object detection with multimodal large language models.
\newblock \emph{International Journal of Computer Vision}, pages 1--19, 2024.

\bibitem[Ganeshan et~al.(2024)Ganeshan, Huang, Xu, Jones, and Ritchie]{ganeshan2024parselparameterizedshapeediting}
Aditya Ganeshan, Ryan~Y. Huang, Xianghao Xu, R.~Kenny Jones, and Daniel Ritchie.
\newblock Parsel: Parameterized shape editing with language, 2024.
\newblock URL \url{https://arxiv.org/abs/2405.20319}.

\bibitem[Qi et~al.(2017)Qi, Yi, Su, and Guibas]{qi2017pointnet2}
Charles~R. Qi, Li~Yi, Hao Su, and Leonidas~J. Guibas.
\newblock Pointnet++: Deep hierarchical feature learning on point sets in a metric space.
\newblock In \emph{Advances in Neural Information Processing Systems (NeurIPS)}, pages 5105--5114, 2017.

\bibitem[Wang et~al.(2019)Wang, Sun, Liu, Sarasua, Bronstein, and Solomon]{wang2019dynamic}
Yue Wang, Yongbin Sun, Ziwei Liu, Sanjay~E. Sarasua, Michael~M. Bronstein, and Justin~M. Solomon.
\newblock Dynamic graph cnn for learning on point clouds.
\newblock In \emph{ACM Transactions on Graphics (TOG)}, volume~38, pages 1--12, 2019.

\bibitem[Achlioptas et~al.(2018)Achlioptas, Diamanti, Mitliagkas, and Guibas]{achlioptas2018learning}
Panos Achlioptas, Olga Diamanti, Ioannis Mitliagkas, and Leonidas Guibas.
\newblock Learning representations and generative models for 3d point clouds.
\newblock In \emph{International Conference on Machine Learning (ICML)}, pages 40--49, 2018.

\bibitem[Chen et~al.(2021)Chen, Xu, et~al.]{chen2021deep}
Weikai Chen, Kai Xu, et~al.
\newblock Deep part-aware shape editing for 3d point clouds.
\newblock In \emph{International Conference on 3D Vision (3DV)}, pages 438--447, 2021.

\bibitem[Fan and Wu(2021)]{fan2021sgpr}
Zhihao Fan and Xiaojun Wu.
\newblock Sgpr: Segmentation-guided point cloud reconstruction for rotated objects.
\newblock In \emph{IEEE Winter Conference on Applications of Computer Vision (WACV)}, pages 1465--1474, 2021.

\bibitem[Wu et~al.(2016)Wu, Zhang, Xue, Freeman, and Tenenbaum]{wu2016learning}
Jiajun Wu, Chengkai Zhang, Tianfan Xue, William~T. Freeman, and Joshua~B. Tenenbaum.
\newblock Learning a probabilistic latent space of object shapes via 3d generative-adversarial modeling.
\newblock In \emph{Advances in Neural Information Processing Systems (NeurIPS)}, pages 82--90, 2016.

\bibitem[Valsesia et~al.(2020)Valsesia, Fracastoro, and Magli]{valsesia2020learning}
Diego Valsesia, Giulia Fracastoro, and Enrico Magli.
\newblock Learning localized generative models for 3d point clouds via graph convolution.
\newblock In \emph{International Conference on Learning Representations (ICLR)}, 2020.

\bibitem[Zhou et~al.(2021)Zhou, Li, Li, et~al.]{zhou20213dgenerative}
Bo~Zhou, Yuan Li, Yijun Li, et~al.
\newblock 3d generative adversarial models with diffusion-based methods.
\newblock In \emph{International Conference on 3D Vision (3DV)}, pages 191--202, 2021.

\bibitem[Luo et~al.(2023)Luo, Hu, Markham, and Guibas]{luo2023scorebased}
Jinliang Luo, Qingyong Hu, Andrew Markham, and Leonidas Guibas.
\newblock Score-based generative models for 3d point cloud generation and editing.
\newblock In \emph{IEEE Conference on Computer Vision and Pattern Recognition (CVPR)}, pages 5527--5537, 2023.

\bibitem[Kim et~al.(2022)Kim, Park, Kwon, Park, and Kim]{kim2022text2mesh}
Min-Gyu Kim, Sanghyun Park, Gyeongsik Kwon, Hyunwoo Park, and Kyoung Mu~Lee Kim.
\newblock Text2mesh: Text-driven neural stylization for meshes and point clouds.
\newblock In \emph{ACM SIGGRAPH Asia}, pages 1--10, 2022.

\bibitem[Nichol et~al.(2022)Nichol, Ramesh, and Dhariwal]{nichol2022pointe}
Alex Nichol, Aditya Ramesh, and Prafulla Dhariwal.
\newblock Point-e: A system for generating 3d point clouds from complex prompts.
\newblock OpenAI Technical Report, 2022.
\newblock \url{https://github.com/openai/point-e}.

\bibitem[Jun et~al.(2023)Jun, Rombach, Blattmann, Beyer, Ommer, Salimans, and Sutskever]{jun2023shape}
Heewoo Jun, Robin Rombach, Andreas Blattmann, Lucas Beyer, Bjorn Ommer, Tim Salimans, and Ilya Sutskever.
\newblock Shap-e: Generating conditional 3d implicit functions.
\newblock OpenAI Technical Report, 2023.

\bibitem[Poole et~al.(2022)Poole, Jalal, Barron, et~al.]{poole2022dreamfusion}
Ben Poole, Ajay Jalal, Jonathan~T. Barron, et~al.
\newblock Dreamfusion: Text-to-3d using 2d diffusion.
\newblock Google Research Preprint, 2022.
\newblock \url{https://dreamfusion3d.github.io/}.

\bibitem[Liu et~al.(2023{\natexlab{c}})Liu, Liu, et~al.]{liu2023zero1to3}
Fangzhou Liu, Pengsong Liu, et~al.
\newblock Zero-1-to-3: Zero-shot one image to 3d object.
\newblock arXiv preprint arXiv:2303.11328, 2023{\natexlab{c}}.

\bibitem[Li et~al.(2021)Li, Shao, and Yang]{li2021shapepart}
Yifan Li, Lin Shao, and Bin Yang.
\newblock Shapepart: Learning region-level decompositions of 3d objects via part-aware shape synthesis.
\newblock In \emph{IEEE International Conference on Computer Vision (ICCV)}, pages 1241--1250, 2021.

\bibitem[Wu et~al.(2021)Wu, Xiang, and Li]{wu2021pq}
Wei Wu, Fangfei Xiang, and Zhen Li.
\newblock Pq-net: A generative part quality network for 3d shape composition.
\newblock In \emph{IEEE Conference on Computer Vision and Pattern Recognition (CVPR)}, pages 4902--4911, 2021.

\bibitem[Su et~al.(2023)Su, Liu, and Zhu]{su2023semanticpc}
Yang Su, Qing Liu, and Song-Chun Zhu.
\newblock Semanticpc: Semantic-driven part composition for 3d object generation.
\newblock In \emph{Conference on Computer Vision and Pattern Recognition (CVPR)}, pages 21334--21343, 2023.

\bibitem[Wu et~al.(2023)Wu, Chen, Li, and Sato]{wu2023det3d}
Haoyang Wu, Boqing Chen, Xi~Li, and Takashi Sato.
\newblock Det3d: Multimodal 3d detection with point clouds, images, and language prompts.
\newblock In \emph{IEEE/RSJ International Conference on Intelligent Robots and Systems (IROS)}, 2023.

\bibitem[Zhao et~al.(2023)Zhao, Yi, Guibas, and Jia]{zhao2023foundations3d}
Hengshuang Zhao, Li~Yi, Leonidas Guibas, and Jiaya Jia.
\newblock Foundations3d: Large-scale pretraining of 3d vision-language models via foundational shapes and descriptions.
\newblock In \emph{International Conference on Machine Learning (ICML)}, pages 8487--8501, 2023.

\bibitem[Hu et~al.(2024)Hu, He, Wang, Zhao, Shao, and Nie]{hu2024llm}
Linmei Hu, Hongyu He, Duokang Wang, Ziwang Zhao, Yingxia Shao, and Liqiang Nie.
\newblock Llm vs small model? large language model based text augmentation enhanced personality detection model.
\newblock In \emph{Proceedings of the AAAI Conference on Artificial Intelligence}, volume~38, pages 18234--18242, 2024.

\bibitem[Wu et~al.(2024{\natexlab{b}})Wu, Chang, Wichern, Jung, Germain, Le~Roux, and Watanabe]{wu2024improving}
Shih-Lun Wu, Xuankai Chang, Gordon Wichern, Jee-weon Jung, Fran{\c{c}}ois Germain, Jonathan Le~Roux, and Shinji Watanabe.
\newblock Improving audio captioning models with fine-grained audio features, text embedding supervision, and llm mix-up augmentation.
\newblock In \emph{ICASSP 2024-2024 IEEE International Conference on Acoustics, Speech and Signal Processing (ICASSP)}, pages 316--320. IEEE, 2024{\natexlab{b}}.

\bibitem[Zhao et~al.(2024)Zhao, Chen, Ruggles, Feng, Singh, and Yoon]{zhao2024improving}
Huanhuan Zhao, Haihua Chen, Thomas~A Ruggles, Yunhe Feng, Debjani Singh, and Hong-Jun Yoon.
\newblock Improving text classification with large language model-based data augmentation.
\newblock \emph{Electronics}, 13\penalty0 (13):\penalty0 2535, 2024.

\bibitem[dos Santos et~al.(2024)dos Santos, Santos, Lynn, and Benatallah]{dos2024identifying}
Vitor~Gaboardi dos Santos, Guto~Leoni Santos, Theo Lynn, and Boualem Benatallah.
\newblock Identifying citizen-related issues from social media using llm-based data augmentation.
\newblock In \emph{International Conference on Advanced Information Systems Engineering}, pages 531--546. Springer, 2024.

\bibitem[Hua et~al.(2023)Hua, Cui, Li, Tang, and Zhu]{hua2023multimodal}
Jiaheng Hua, Xiaodong Cui, Xianghua Li, Keke Tang, and Peican Zhu.
\newblock Multimodal fake news detection through data augmentation-based contrastive learning.
\newblock \emph{Applied Soft Computing}, 136:\penalty0 110125, 2023.

\bibitem[Lai et~al.(2024)Lai, Yang, Luo, Zhou, Li, Wang, and Shi]{lai2024rumorllm}
Jianqiao Lai, Xinran Yang, Wenyue Luo, Linjiang Zhou, Langchen Li, Yongqi Wang, and Xiaochuan Shi.
\newblock Rumorllm: A rumor large language model-based fake-news-detection data-augmentation approach.
\newblock \emph{Applied Sciences}, 14\penalty0 (8):\penalty0 3532, 2024.

\bibitem[Zhang et~al.(2024{\natexlab{a}})Zhang, Gao, Zhang, Feng, Deng, and Hou]{zhang2024ucl}
Jing Zhang, Hui Gao, Peng Zhang, Boda Feng, Wenmin Deng, and Yuexian Hou.
\newblock La-ucl: Llm-augmented unsupervised contrastive learning framework for few-shot text classification.
\newblock In \emph{Proceedings of the 2024 Joint International Conference on Computational Linguistics, Language Resources and Evaluation (LREC-COLING 2024)}, pages 10198--10207, 2024{\natexlab{a}}.

\bibitem[Zhang et~al.(2024{\natexlab{b}})Zhang, Jiang, Liu, Chen, and Zhang]{zhang2024llm}
Meishan Zhang, Gongyao Jiang, Shuang Liu, Jing Chen, and Min Zhang.
\newblock Llm--assisted data augmentation for chinese dialogue--level dependency parsing.
\newblock \emph{Computational Linguistics}, pages 1--24, 2024{\natexlab{b}}.

\bibitem[Wan et~al.(2024)Wan, Safavi, Jauhar, Kim, Counts, Neville, Suri, Shah, White, Yang, et~al.]{wan2024tnt}
Mengting Wan, Tara Safavi, Sujay~Kumar Jauhar, Yujin Kim, Scott Counts, Jennifer Neville, Siddharth Suri, Chirag Shah, Ryen~W White, Longqi Yang, et~al.
\newblock Tnt-llm: Text mining at scale with large language models.
\newblock In \emph{Proceedings of the 30th ACM SIGKDD Conference on Knowledge Discovery and Data Mining}, pages 5836--5847, 2024.

\bibitem[Cai et~al.(2023)Cai, Xiao, Ning, and Zhou]{cai2023resolving}
Xunxin Cai, Meng Xiao, Zhiyuan Ning, and Yuanchun Zhou.
\newblock Resolving the imbalance issue in hierarchical disciplinary topic inference via llm-based data augmentation.
\newblock In \emph{2023 IEEE International Conference on Data Mining Workshops (ICDMW)}, pages 1424--1429. IEEE, 2023.

\bibitem[Yuan et~al.(2023{\natexlab{b}})Yuan, Tang, Jiang, and Hu]{yuan2023llm}
Jiayi Yuan, Ruixiang Tang, Xiaoqian Jiang, and Xia Hu.
\newblock Llm for patient-trial matching: Privacy-aware data augmentation towards better performance and generalizability.
\newblock In \emph{American Medical Informatics Association (AMIA) Annual Symposium}, 2023{\natexlab{b}}.

\bibitem[Latif and Kim(2024)]{latif2024evaluation}
Atif Latif and Jihie Kim.
\newblock Evaluation and analysis of large language models for clinical text augmentation and generation.
\newblock \emph{IEEE Access}, 2024.

\bibitem[Ahmed et~al.(2024)Ahmed, Pai, Devanbu, and Barr]{ahmed2024automatic}
Toufique Ahmed, Kunal~Suresh Pai, Premkumar Devanbu, and Earl Barr.
\newblock Automatic semantic augmentation of language model prompts (for code summarization).
\newblock In \emph{Proceedings of the IEEE/ACM 46th International Conference on Software Engineering}, pages 1--13, 2024.

\bibitem[Meng et~al.(2024)Meng, Liu, Zhang, Feng, and Zhao]{meng2024cean}
Zihao Meng, Tao Liu, Heng Zhang, Kai Feng, and Peng Zhao.
\newblock Cean: Contrastive event aggregation network with llm-based augmentation for event extraction.
\newblock In \emph{Proceedings of the 18th Conference of the European Chapter of the Association for Computational Linguistics (Volume 1: Long Papers)}, pages 321--333, 2024.

\bibitem[{\c{C}}atalta{\c{s}} et~al.(2023){\c{C}}atalta{\c{s}}, Baykan, and Cicekli]{ccataltacs2023comparison}
Mustafa {\c{C}}atalta{\c{s}}, Nurdan~Akhan Baykan, and Ilyas Cicekli.
\newblock Comparison of textual data augmentation methods on sst-2 dataset.
\newblock In \emph{International Congress of Electrical and Computer Engineering}, pages 189--201. Springer, 2023.

\bibitem[Cloutier and Japkowicz(2023)]{cloutier2023fine}
Nicolas~Antonio Cloutier and Nathalie Japkowicz.
\newblock Fine-tuned generative llm oversampling can improve performance over traditional techniques on multiclass imbalanced text classification.
\newblock In \emph{2023 IEEE International Conference on Big Data (BigData)}, pages 5181--5186. IEEE, 2023.

\bibitem[Jung et~al.(2023)Jung, Yeen, Lee, Kim, Bang, and Koo]{jung2023enhancing}
Haein Jung, Heuiyeen Yeen, Jeehyun Lee, Minju Kim, Namo Bang, and Myoung-Wan Koo.
\newblock Enhancing task-oriented dialog system with subjective knowledge: A large language model-based data augmentation framework.
\newblock In \emph{Proceedings of The Eleventh Dialog System Technology Challenge}, pages 150--165, 2023.

\bibitem[Silva et~al.(2024)Silva, Frommholz, Can, Blain, Sarwar, and Ugolini]{silva2024forged}
Kanishka Silva, Ingo Frommholz, Burcu Can, Fred Blain, Raheem Sarwar, and Laura Ugolini.
\newblock Forged-gan-bert: Authorship attribution for llm-generated forged novels.
\newblock In \emph{Proceedings of the 18th Conference of the European Chapter of the Association for Computational Linguistics: Student Research Workshop}, pages 325--337, 2024.

\bibitem[Deng et~al.(2024)Deng, Gao, Chen, Niu, Gong, Zhang, Cao, Li, Ma, Wei, et~al.]{deng2024ophglm}
Zhuo Deng, Weihao Gao, Chucheng Chen, Zhiyuan Niu, Zheng Gong, Ruiheng Zhang, Zhenjie Cao, Fang Li, Zhaoyi Ma, Wenbin Wei, et~al.
\newblock Ophglm: An ophthalmology large language-and-vision assistant.
\newblock \emph{Artificial Intelligence in Medicine}, 157:\penalty0 103001, 2024.

\bibitem[Glazkova and Zakharova(2024)]{glazkova2024evaluating}
Anna Glazkova and Olga Zakharova.
\newblock Evaluating llm prompts for data augmentation in multi-label classification of ecological texts.
\newblock \emph{arXiv preprint arXiv:2411.14896}, 2024.

\bibitem[Fischer et~al.(2024)Fischer, Gao, Lintner, and Ebling]{fischer2024swissadt}
Lukas Fischer, Yingqiang Gao, Alexa Lintner, and Sarah Ebling.
\newblock Swissadt: An audio description translation system for swiss languages.
\newblock \emph{arXiv preprint arXiv:2411.14967}, 2024.

\bibitem[Dai et~al.(2023)Dai, Liu, Liao, Huang, Cao, Wu, Zhao, Xu, Liu, Liu, et~al.]{dai2023auggpt}
Haixing Dai, Zhengliang Liu, Wenxiong Liao, Xiaoke Huang, Yihan Cao, Zihao Wu, Lin Zhao, Shaochen Xu, Wei Liu, Ninghao Liu, et~al.
\newblock Auggpt: Leveraging chatgpt for text data augmentation.
\newblock \emph{arXiv preprint arXiv:2302.13007}, 2023.

\bibitem[Lee et~al.(2024{\natexlab{a}})Lee, Wattanawong, Kim, Mangalam, Shen, Anumanchipalli, Mahoney, Keutzer, and Gholami]{lee2024llm2llm}
Nicholas Lee, Thanakul Wattanawong, Sehoon Kim, Karttikeya Mangalam, Sheng Shen, Gopala Anumanchipalli, Michael~W Mahoney, Kurt Keutzer, and Amir Gholami.
\newblock Llm2llm: Boosting llms with novel iterative data enhancement.
\newblock \emph{arXiv preprint arXiv:2403.15042}, 2024{\natexlab{a}}.

\bibitem[Wen et~al.(2024{\natexlab{a}})Wen, Guo, and Zhang]{wen2024aidbench}
Zichen Wen, Dadi Guo, and Huishuai Zhang.
\newblock Aidbench: A benchmark for evaluating the authorship identification capability of large language models.
\newblock \emph{arXiv preprint arXiv:2411.13226}, 2024{\natexlab{a}}.

\bibitem[Kang et~al.(2024)Kang, Chen, Lee-Youngzie, and Fu]{kang2024synthetic}
Andrea Kang, Jun~Yu Chen, Zoe Lee-Youngzie, and Shuhao Fu.
\newblock Synthetic data generation with llm for improved depression prediction.
\newblock \emph{arXiv preprint arXiv:2411.17672}, 2024.

\bibitem[Cegin et~al.(2024{\natexlab{a}})Cegin, Simko, and Brusilovsky]{cegin2024llms}
Jan Cegin, Jakub Simko, and Peter Brusilovsky.
\newblock Llms vs established text augmentation techniques for classification: When do the benefits outweight the costs?
\newblock \emph{arXiv preprint arXiv:2408.16502}, 2024{\natexlab{a}}.

\bibitem[Alyafeai et~al.(2024)Alyafeai, Pieler, Teufel, Tow, Bellagente, Phung, Pinnaparaju, Adithyan, Rocha, Zhuravinskyi, et~al.]{alyafeai2024arabic}
Zaid Alyafeai, Michael Pieler, Hannah Teufel, Jonathan Tow, Marco Bellagente, Duy Phung, Nikhil Pinnaparaju, Reshinth Adithyan, Paulo Rocha, Maksym Zhuravinskyi, et~al.
\newblock Arabic stable lm: Adapting stable lm 2 1.6 b to arabic.
\newblock \emph{arXiv preprint arXiv:2412.04277}, 2024.

\bibitem[Yoran et~al.(2023)Yoran, Wolfson, Ram, and Berant]{yoran2023making}
Ori Yoran, Tomer Wolfson, Ori Ram, and Jonathan Berant.
\newblock Making retrieval-augmented language models robust to irrelevant context.
\newblock \emph{arXiv preprint arXiv:2310.01558}, 2023.

\bibitem[Lee et~al.(2024{\natexlab{b}})Lee, Chen, Chen, Lee, Chen, Wu, and Dai]{lee2024unlocking}
You-Qian Lee, Ching-Tai Chen, Chien-Chang Chen, Chung-Hong Lee, Peitsz Chen, Chi-Shin Wu, and Hong-Jie Dai.
\newblock Unlocking the secrets behind advanced artificial intelligence language models in deidentifying chinese-english mixed clinical text: Development and validation study.
\newblock \emph{Journal of Medical Internet Research}, 26:\penalty0 e48443, 2024{\natexlab{b}}.

\bibitem[Whitehouse et~al.(2023)Whitehouse, Choudhury, and Aji]{whitehouse2023llm}
Chenxi Whitehouse, Monojit Choudhury, and Alham~Fikri Aji.
\newblock Llm-powered data augmentation for enhanced cross-lingual performance.
\newblock \emph{arXiv preprint arXiv:2305.14288}, 2023.

\bibitem[Min et~al.(2023)Min, Ross, Sulem, Veyseh, Nguyen, Sainz, Agirre, Heintz, and Roth]{min2023recent}
Bonan Min, Hayley Ross, Elior Sulem, Amir Pouran~Ben Veyseh, Thien~Huu Nguyen, Oscar Sainz, Eneko Agirre, Ilana Heintz, and Dan Roth.
\newblock Recent advances in natural language processing via large pre-trained language models: A survey.
\newblock \emph{ACM Computing Surveys}, 56\penalty0 (2):\penalty0 1--40, 2023.

\bibitem[Gupta and Agrawal(2022)]{gupta2022compression}
Manish Gupta and Puneet Agrawal.
\newblock Compression of deep learning models for text: A survey.
\newblock \emph{ACM Transactions on Knowledge Discovery from Data (TKDD)}, 16\penalty0 (4):\penalty0 1--55, 2022.

\bibitem[Dantas et~al.(2024)Dantas, Sabino~da Silva~Jr, Cordeiro, and Carvalho]{dantas2024comprehensive}
Pierre~Vilar Dantas, Waldir Sabino~da Silva~Jr, Lucas~Carvalho Cordeiro, and Celso~Barbosa Carvalho.
\newblock A comprehensive review of model compression techniques in machine learning.
\newblock \emph{Applied Intelligence}, 54\penalty0 (22):\penalty0 11804--11844, 2024.

\bibitem[Wu et~al.(2025)Wu, Yang, Zhan, Yuan, Chao, and Wong]{wu2025survey}
Junchao Wu, Shu Yang, Runzhe Zhan, Yulin Yuan, Lidia~Sam Chao, and Derek~Fai Wong.
\newblock A survey on llm-generated text detection: Necessity, methods, and future directions.
\newblock \emph{Computational Linguistics}, pages 1--65, 2025.

\bibitem[Atashbar(2024)]{atashbar2024reinforcement}
Tohid Atashbar.
\newblock Reinforcement learning from experience feedback: Application to economic policy.
\newblock 2024.

\bibitem[Ye et~al.(2024{\natexlab{a}})Ye, Xu, Wang, Zhou, Zhang, Gui, and Huang]{ye2024llm}
Junjie Ye, Nuo Xu, Yikun Wang, Jie Zhou, Qi~Zhang, Tao Gui, and Xuanjing Huang.
\newblock Llm-da: Data augmentation via large language models for few-shot named entity recognition.
\newblock \emph{arXiv preprint arXiv:2402.14568}, 2024{\natexlab{a}}.

\bibitem[Sapkota et~al.(2025)Sapkota, Raza, and Karkee]{sapkota2025comprehensive}
Ranjan Sapkota, Shaina Raza, and Manoj Karkee.
\newblock Comprehensive analysis of transparency and accessibility of chatgpt, deepseek, and other sota large language models.
\newblock \emph{arXiv preprint arXiv:2502.18505}, 2025.

\bibitem[Ghosh et~al.(2024{\natexlab{a}})Ghosh, Hasan, Arafat, and Khan]{ghosh2024logical}
Bishwamittra Ghosh, Sarah Hasan, Naheed~Anjum Arafat, and Arijit Khan.
\newblock Logical consistency of large language models in fact-checking.
\newblock \emph{arXiv preprint arXiv:2412.16100}, 2024{\natexlab{a}}.

\bibitem[Huang et~al.(2023{\natexlab{a}})Huang, Yu, Ma, Zhong, Feng, Wang, Chen, Peng, Feng, Qin, et~al.]{huang2023survey}
Lei Huang, Weijiang Yu, Weitao Ma, Weihong Zhong, Zhangyin Feng, Haotian Wang, Qianglong Chen, Weihua Peng, Xiaocheng Feng, Bing Qin, et~al.
\newblock A survey on hallucination in large language models: Principles, taxonomy, challenges, and open questions.
\newblock \emph{ACM Transactions on Information Systems}, 2023{\natexlab{a}}.

\bibitem[Liu et~al.(2024{\natexlab{c}})Liu, Guo, Liang, Shareghi, Vuli{\'c}, and Collier]{liu2024aligning}
Yinhong Liu, Zhijiang Guo, Tianya Liang, Ehsan Shareghi, Ivan Vuli{\'c}, and Nigel Collier.
\newblock Aligning with logic: Measuring, evaluating and improving logical consistency in large language models.
\newblock \emph{arXiv preprint arXiv:2410.02205}, 2024{\natexlab{c}}.

\bibitem[Bolding et~al.(2023)Bolding, Liao, Denis, Luo, and Monz]{bolding2023ask}
Quinten Bolding, Baohao Liao, Brandon~James Denis, Jun Luo, and Christof Monz.
\newblock Ask language model to clean your noisy translation data.
\newblock \emph{arXiv preprint arXiv:2310.13469}, 2023.

\bibitem[Yin et~al.(2024{\natexlab{b}})Yin, Xiao, Bai, and Das]{yin2024leveraging}
Han Yin, Yang Xiao, Jisheng Bai, and Rohan~Kumar Das.
\newblock Leveraging llm and text-queried separation for noise-robust sound event detection.
\newblock \emph{arXiv preprint arXiv:2411.01174}, 2024{\natexlab{b}}.

\bibitem[Yu et~al.(2024)Yu, Zhuang, Zhang, Meng, Ratner, Krishna, Shen, and Zhang]{yu2024large}
Yue Yu, Yuchen Zhuang, Jieyu Zhang, Yu~Meng, Alexander~J Ratner, Ranjay Krishna, Jiaming Shen, and Chao Zhang.
\newblock Large language model as attributed training data generator: A tale of diversity and bias.
\newblock \emph{Advances in Neural Information Processing Systems}, 36, 2024.

\bibitem[Kim et~al.(2025)Kim, Im, Kim, and Oh]{kim2025tardis}
Kyungmin Kim, SangHun Im, GiBaeg Kim, and Heung-Seon Oh.
\newblock Tardis: Text augmentation for refining diversity and separability.
\newblock \emph{arXiv preprint arXiv:2501.02739}, 2025.

\bibitem[Cegin et~al.(2024{\natexlab{b}})Cegin, Pecher, Simko, Srba, Bielikova, and Brusilovsky]{cegin2024effects}
Jan Cegin, Branislav Pecher, Jakub Simko, Ivan Srba, Maria Bielikova, and Peter Brusilovsky.
\newblock Effects of diversity incentives on sample diversity and downstream model performance in llm-based text augmentation.
\newblock \emph{arXiv preprint arXiv:2401.06643}, 2024{\natexlab{b}}.

\bibitem[Cai et~al.(2024)Cai, Ghosh, Adatia, Hayat, Dhall, Gedeon, and Stefanov]{cai2024av}
Zhixi Cai, Shreya Ghosh, Aman~Pankaj Adatia, Munawar Hayat, Abhinav Dhall, Tom Gedeon, and Kalin Stefanov.
\newblock Av-deepfake1m: A large-scale llm-driven audio-visual deepfake dataset.
\newblock In \emph{Proceedings of the 32nd ACM International Conference on Multimedia}, pages 7414--7423, 2024.

\bibitem[Dhingra et~al.(2024{\natexlab{a}})Dhingra, Agrawal, Veerappan, Ho, Chng, and Tong]{dhingra2024speech}
Priyanshu Dhingra, Satyam Agrawal, Chandra~Sekar Veerappan, Thi~Nga Ho, Eng~Siong Chng, and Rong Tong.
\newblock Speech de-identification data augmentation leveraging large language model.
\newblock In \emph{2024 International Conference on Asian Language Processing (IALP)}, pages 97--102. IEEE, 2024{\natexlab{a}}.

\bibitem[Dhingra et~al.(2024{\natexlab{b}})Dhingra, Agrawal, Veerappan, Chng, and Tong]{dhingra2024enhancing}
Priyanshu Dhingra, Satyam Agrawal, Chandra~Sekar Veerappan, Eng~Siong Chng, and Rong Tong.
\newblock Enhancing speech de-identification with llm-based data augmentation.
\newblock In \emph{2024 11th International Conference on Advanced Informatics: Concept, Theory and Application (ICAICTA)}, pages 1--5. IEEE, 2024{\natexlab{b}}.

\bibitem[Ma et~al.(2024)Ma, Wu, Zheng, Guo, Chen, Zhang, and Chen]{ma2024leveraging}
Ziyang Ma, Wen Wu, Zhisheng Zheng, Yiwei Guo, Qian Chen, Shiliang Zhang, and Xie Chen.
\newblock Leveraging speech ptm, text llm, and emotional tts for speech emotion recognition.
\newblock In \emph{ICASSP 2024-2024 IEEE International Conference on Acoustics, Speech and Signal Processing (ICASSP)}, pages 11146--11150. IEEE, 2024.

\bibitem[Xu(2024)]{xu2024audiosetmix}
David Xu.
\newblock Audiosetmix: Enhancing audio-language datasets with llm-assisted augmentations.
\newblock \emph{arXiv preprint arXiv:2405.11093}, 2024.

\bibitem[Ghosh et~al.(2024{\natexlab{b}})Ghosh, Kumar, Kong, Valle, Catanzaro, and Manocha]{ghosh2024synthio}
Sreyan Ghosh, Sonal Kumar, Zhifeng Kong, Rafael Valle, Bryan Catanzaro, and Dinesh Manocha.
\newblock Synthio: Augmenting small-scale audio classification datasets with synthetic data.
\newblock \emph{arXiv preprint arXiv:2410.02056}, 2024{\natexlab{b}}.

\bibitem[Heakl et~al.(2024)Heakl, Zaghloul, Ali, Hossam, and Gomaa]{heakl2024arzen}
Ahmed Heakl, Youssef Zaghloul, Mennatullah Ali, Rania Hossam, and Walid Gomaa.
\newblock Arzen-llm: Code-switched egyptian arabic-english translation and speech recognition using llms.
\newblock \emph{Procedia Computer Science}, 244:\penalty0 113--120, 2024.

\bibitem[Hashmi et~al.(2024)Hashmi, Yayilgan, Yamin, Abomhara, and Ullah]{hashmi2024self}
Ehtesham Hashmi, Sule~Yildirim Yayilgan, Muhammad~Mudassar Yamin, Mohamed Abomhara, and Mohib Ullah.
\newblock Self-supervised hate speech detection in norwegian texts with lexical and semantic augmentations.
\newblock \emph{Expert Systems with Applications}, page 125843, 2024.

\bibitem[Xu et~al.(2024)Xu, Zhou, Nguyen, Bao, Lin, and Du]{xu2024integrating}
Fang Xu, Tianyu Zhou, Tri Nguyen, Haohui Bao, Christine Lin, and Jing Du.
\newblock Integrating augmented reality and llm for enhanced cognitive support in critical audio communications.
\newblock \emph{International Journal of Human-Computer Studies}, page 103402, 2024.

\bibitem[Cook and Karaku{\c{s}}(2024)]{cook2024llm}
Alec Cook and Oktay Karaku{\c{s}}.
\newblock Llm-commentator: Novel fine-tuning strategies of large language models for automatic commentary generation using football event data.
\newblock \emph{Knowledge-Based Systems}, 300:\penalty0 112219, 2024.

\bibitem[Gkournelos et~al.(2024)Gkournelos, Konstantinou, and Makris]{gkournelos2024llm}
Christos Gkournelos, Christos Konstantinou, and Sotiris Makris.
\newblock An llm-based approach for enabling seamless human-robot collaboration in assembly.
\newblock \emph{CIRP Annals}, 2024.

\bibitem[Alier et~al.(2025)Alier, Pereira, Garc{\'\i}a-Pe{\~n}alvo, Casa{\~n}, and Cabr{\'e}]{alier2025lamb}
Marc Alier, Juanan Pereira, Francisco~Jos{\'e} Garc{\'\i}a-Pe{\~n}alvo, Maria~Jose Casa{\~n}, and Jose Cabr{\'e}.
\newblock Lamb: An open-source software framework to create artificial intelligence assistants deployed and integrated into learning management systems.
\newblock \emph{Computer Standards \& Interfaces}, 92:\penalty0 103940, 2025.

\bibitem[Senthilselvi et~al.(2024)Senthilselvi, Prawin, Harshit, et~al.]{senthilselvi2024abstractive}
A~Senthilselvi, RP~Prawin, V~Harshit, et~al.
\newblock Abstractive summarization of youtube videos using lamini-flan-t5 llm.
\newblock In \emph{2024 Second International Conference on Advances in Information Technology (ICAIT)}, volume~1, pages 1--5. IEEE, 2024.

\bibitem[Wang et~al.(2024{\natexlab{b}})Wang, Shafran, Soltau, Han, Cao, Yu, and El~Shafey]{wang2024retrieval}
Mingqiu Wang, Izhak Shafran, Hagen Soltau, Wei Han, Yuan Cao, Dian Yu, and Laurent El~Shafey.
\newblock Retrieval augmented end-to-end spoken dialog models.
\newblock In \emph{ICASSP 2024-2024 IEEE International Conference on Acoustics, Speech and Signal Processing (ICASSP)}, pages 12056--12060. IEEE, 2024{\natexlab{b}}.

\bibitem[Qiu et~al.(2024)Qiu, Wu, Zhang, Lin, Wang, Zhang, Wang, and Xie]{qiu2024towards}
Pengcheng Qiu, Chaoyi Wu, Xiaoman Zhang, Weixiong Lin, Haicheng Wang, Ya~Zhang, Yanfeng Wang, and Weidi Xie.
\newblock Towards building multilingual language model for medicine.
\newblock \emph{Nature Communications}, 15\penalty0 (1):\penalty0 8384, 2024.

\bibitem[Hasebe et~al.(2024)Hasebe, Fujimura, Kojima, Tamura, Kawai, Kishimoto, and Omori]{hasebe2024effect}
Koki Hasebe, Shintaro Fujimura, Tsuyoshi Kojima, Keiichi Tamura, Yoshitaka Kawai, Yo~Kishimoto, and Koichi Omori.
\newblock The effect of noise on deep learning for classification of pathological voice.
\newblock \emph{The Laryngoscope}, 134\penalty0 (8):\penalty0 3537--3541, 2024.

\bibitem[Sridhar et~al.(2024)Sridhar, Guo, and Visser]{sridhar2024enhancing}
Arvind~Krishna Sridhar, Yinyi Guo, and Erik Visser.
\newblock Enhancing temporal understanding in audio question answering for large audio language models.
\newblock \emph{arXiv preprint arXiv:2409.06223}, 2024.

\bibitem[Lei et~al.(2024)Lei, Na, Xu, Pusateri, Van~Gysel, Zhang, Han, and Huang]{lei2024contextualization}
Zhihong Lei, Xingyu Na, Mingbin Xu, Ernest Pusateri, Christophe Van~Gysel, Yuanyuan Zhang, Shiyi Han, and Zhen Huang.
\newblock Contextualization of asr with llm using phonetic retrieval-based augmentation.
\newblock \emph{arXiv preprint arXiv:2409.15353}, 2024.

\bibitem[Goel et~al.(2024)Goel, Kong, Valle, and Catanzaro]{goel2024audio}
Arushi Goel, Zhifeng Kong, Rafael Valle, and Bryan Catanzaro.
\newblock Audio dialogues: Dialogues dataset for audio and music understanding.
\newblock \emph{arXiv preprint arXiv:2404.07616}, 2024.

\bibitem[Yang et~al.(2023)Yang, Tian, Tan, Huang, Liu, Chang, Shi, Zhao, Bian, Wu, et~al.]{yang2023uniaudio}
Dongchao Yang, Jinchuan Tian, Xu~Tan, Rongjie Huang, Songxiang Liu, Xuankai Chang, Jiatong Shi, Sheng Zhao, Jiang Bian, Xixin Wu, et~al.
\newblock Uniaudio: An audio foundation model toward universal audio generation.
\newblock \emph{arXiv preprint arXiv:2310.00704}, 2023.

\bibitem[Wang et~al.(2024{\natexlab{c}})Wang, Tai, and Tang]{wang2024audio}
Zixuan Wang, Yu-Wing Tai, and Chi-Keung Tang.
\newblock Audio-agent: Leveraging llms for audio generation, editing and composition.
\newblock \emph{arXiv preprint arXiv:2410.03335}, 2024{\natexlab{c}}.

\bibitem[Cuskley et~al.(2024)Cuskley, Woods, and Flaherty]{cuskley2024limitations}
Christine Cuskley, Rebecca Woods, and Molly Flaherty.
\newblock The limitations of large language models for understanding human language and cognition.
\newblock \emph{Open Mind}, 8:\penalty0 1058--1083, 2024.

\bibitem[Lee et~al.(2024{\natexlab{c}})Lee, Song, and Kim]{lee2024performance}
Do~Hyun Lee, Yoonah Song, and Hong~Kook Kim.
\newblock Performance improvement of language-queried audio source separation based on caption augmentation from large language models for dcase challenge 2024 task 9.
\newblock \emph{arXiv preprint arXiv:2406.11248}, 2024{\natexlab{c}}.

\bibitem[Srivastava et~al.(2024)Srivastava, Khanna, Pan, Nguyen, and Jain]{srivastava2024unvoiced}
Tanmay Srivastava, Prerna Khanna, Shijia Pan, Phuc Nguyen, and Shubham Jain.
\newblock Unvoiced: Designing an llm-assisted unvoiced user interface using earables.
\newblock In \emph{Proceedings of the 22nd ACM Conference on Embedded Networked Sensor Systems}, pages 784--798, 2024.

\bibitem[Kheddar et~al.(2024)Kheddar, Hemis, and Himeur]{kheddar2024automatic}
Hamza Kheddar, Mustapha Hemis, and Yassine Himeur.
\newblock Automatic speech recognition using advanced deep learning approaches: A survey.
\newblock \emph{Information Fusion}, page 102422, 2024.

\bibitem[Li et~al.(2024{\natexlab{e}})Li, Xu, Zhang, Song, Cai, Qi, Zhou, Pan, Li, Tu, et~al.]{li2024groundinggpt}
Zhaowei Li, Qi~Xu, Dong Zhang, Hang Song, Yiqing Cai, Qi~Qi, Ran Zhou, Junting Pan, Zefeng Li, Vu~Tu, et~al.
\newblock Groundinggpt: Language enhanced multi-modal grounding model.
\newblock In \emph{Proceedings of the 62nd Annual Meeting of the Association for Computational Linguistics (Volume 1: Long Papers)}, pages 6657--6678, 2024{\natexlab{e}}.

\bibitem[Zhu et~al.(2024)Zhu, He, Jing, Song, Lian, Zhang, and Li]{zhu2024llm}
Yuanyuan Zhu, Jiaxu He, Ruihao Jing, Yaodong Song, Jie Lian, Xiao-lei Zhang, and Jie Li.
\newblock Llm-based expressive text-to-speech synthesizer with style and timbre disentanglement.
\newblock In \emph{2024 IEEE 14th International Symposium on Chinese Spoken Language Processing (ISCSLP)}, pages 596--600. IEEE, 2024.

\bibitem[Heidemann(2016)]{heidemann2016system}
Kate Heidemann.
\newblock A system for describing vocal timbre in popular song.
\newblock \emph{Music Theory Online}, 22\penalty0 (1), 2016.

\bibitem[Schneider(2018)]{schneider2018perception}
Albrecht Schneider.
\newblock Perception of timbre and sound color.
\newblock \emph{Springer Handbook of Systematic Musicology}, pages 687--725, 2018.

\bibitem[Chen et~al.(2024{\natexlab{a}})Chen, Feng, He, He, He, Hu, Lin, Lin, Pan, Tan, et~al.]{chen2024takin}
Sijing Chen, Yuan Feng, Laipeng He, Tianwei He, Wendi He, Yanni Hu, Bin Lin, Yiting Lin, Yu~Pan, Pengfei Tan, et~al.
\newblock Takin: A cohort of superior quality zero-shot speech generation models.
\newblock \emph{arXiv preprint arXiv:2409.12139}, 2024{\natexlab{a}}.

\bibitem[Ye et~al.(2024{\natexlab{b}})Ye, Sun, Lei, Lin, Tan, Dai, Kong, Chen, Pan, Liu, et~al.]{ye2024codec}
Zhen Ye, Peiwen Sun, Jiahe Lei, Hongzhan Lin, Xu~Tan, Zheqi Dai, Qiuqiang Kong, Jianyi Chen, Jiahao Pan, Qifeng Liu, et~al.
\newblock Codec does matter: Exploring the semantic shortcoming of codec for audio language model.
\newblock \emph{arXiv preprint arXiv:2408.17175}, 2024{\natexlab{b}}.

\bibitem[Barreto et~al.(2023)Barreto, Moharkar, Shirodkar, Sarode, Gonsalves, and Johns]{barreto2023generative}
Fabian Barreto, Lalita Moharkar, Madhura Shirodkar, Vidya Sarode, Saniya Gonsalves, and Aaron Johns.
\newblock Generative artificial intelligence: Opportunities and challenges of large language models.
\newblock In \emph{International Conference on Intelligent Computing and Networking}, pages 545--553. Springer, 2023.

\bibitem[Zhang et~al.(2024{\natexlab{c}})Zhang, Zhang, Ni, Wei, Yang, Jin, Huang, Liang, Zhang, Li, et~al.]{zhang2024multimodal}
Zhenwei Zhang, Shengming Zhang, Dong Ni, Zhaoguo Wei, Kongjun Yang, Shan Jin, Gan Huang, Zhen Liang, Li~Zhang, Linling Li, et~al.
\newblock Multimodal sensing for depression risk detection: integrating audio, video, and text data.
\newblock \emph{Sensors}, 24\penalty0 (12):\penalty0 3714, 2024{\natexlab{c}}.

\bibitem[Mariya~Celin et~al.(2023)Mariya~Celin, Vijayalakshmi, and Nagarajan]{mariya2023data}
TA~Mariya~Celin, P~Vijayalakshmi, and T~Nagarajan.
\newblock Data augmentation techniques for transfer learning-based continuous dysarthric speech recognition.
\newblock \emph{Circuits, Systems, and Signal Processing}, 42\penalty0 (1):\penalty0 601--622, 2023.

\bibitem[Peng et~al.(2024{\natexlab{a}})Peng, Puvvada, Chen, Zelasko, Huang, Dhawan, Hu, Watanabe, Balam, and Ginsburg]{peng2024voicetextblender}
Yifan Peng, Krishna~C Puvvada, Zhehuai Chen, Piotr Zelasko, He~Huang, Kunal Dhawan, Ke~Hu, Shinji Watanabe, Jagadeesh Balam, and Boris Ginsburg.
\newblock Voicetextblender: Augmenting large language models with speech capabilities via single-stage joint speech-text supervised fine-tuning.
\newblock \emph{arXiv preprint arXiv:2410.17485}, 2024{\natexlab{a}}.

\bibitem[Nautsch et~al.(2019)Nautsch, Jim{\'e}nez, Treiber, Kolberg, Jasserand, Kindt, Delgado, Todisco, Hmani, Mtibaa, et~al.]{nautsch2019preserving}
Andreas Nautsch, Abelino Jim{\'e}nez, Amos Treiber, Jascha Kolberg, Catherine Jasserand, Els Kindt, H{\'e}ctor Delgado, Massimiliano Todisco, Mohamed~Amine Hmani, Aymen Mtibaa, et~al.
\newblock Preserving privacy in speaker and speech characterisation.
\newblock \emph{Computer Speech \& Language}, 58:\penalty0 441--480, 2019.

\bibitem[Hu et~al.(2019)Hu, Shang, Qin, Li, Wang, and Wang]{hu2019adversarial}
Shengshan Hu, Xingcan Shang, Zhan Qin, Minghui Li, Qian Wang, and Cong Wang.
\newblock Adversarial examples for automatic speech recognition: Attacks and countermeasures.
\newblock \emph{IEEE Communications Magazine}, 57\penalty0 (10):\penalty0 120--126, 2019.

\bibitem[Chen et~al.(2022)Chen, Zhao, Song, Chen, Fan, Wang, and Wang]{chen2022towards}
Guangke Chen, Zhe Zhao, Fu~Song, Sen Chen, Lingling Fan, Feng Wang, and Jiashui Wang.
\newblock Towards understanding and mitigating audio adversarial examples for speaker recognition.
\newblock \emph{IEEE Transactions on Dependable and Secure Computing}, 20\penalty0 (5):\penalty0 3970--3987, 2022.

\bibitem[Han et~al.(2024)Han, Wang, Zhang, Li, and Duan]{han2024review}
Songyue Han, Mingyu Wang, Jialong Zhang, Dongdong Li, and Junhong Duan.
\newblock A review of large language models: Fundamental architectures, key technological evolutions, interdisciplinary technologies integration, optimization and compression techniques, applications, and challenges.
\newblock \emph{Electronics}, 13\penalty0 (24):\penalty0 5040, 2024.

\bibitem[Ogof et~al.(2024)Ogof, Romanov, and Polanski]{ogof2024enhancing}
Daniel Ogof, Anastasia Romanov, and Viktor Polanski.
\newblock Enhancing audio comprehension in large language models: Integrating audio knowledge.
\newblock \emph{Authorea Preprints}, 2024.

\bibitem[Dekel et~al.(2024)Dekel, Shechtman, Fernandez, Haws, Kons, and Hoory]{dekel2024speak}
Avihu Dekel, Slava Shechtman, Raul Fernandez, David Haws, Zvi Kons, and Ron Hoory.
\newblock Speak while you think: Streaming speech synthesis during text generation.
\newblock In \emph{ICASSP 2024-2024 IEEE International Conference on Acoustics, Speech and Signal Processing (ICASSP)}, pages 11931--11935. IEEE, 2024.

\bibitem[Zheng et~al.(2025)Zheng, Jiang, Gu, Li, Wang, and Zhang]{zheng2025teaching}
Longwei Zheng, Fei Jiang, Xiaoqing Gu, Yuanyuan Li, Gong Wang, and Haomin Zhang.
\newblock Teaching via llm-enhanced simulations: Authenticity and barriers to suspension of disbelief.
\newblock \emph{The Internet and Higher Education}, 65:\penalty0 100990, 2025.

\bibitem[Ghiur{\u{a}}u and Popescu(2024)]{ghiuruau2024distinguishing}
David Ghiur{\u{a}}u and Daniela~Elena Popescu.
\newblock Distinguishing reality from ai: Approaches for detecting synthetic content.
\newblock \emph{Computers}, 14\penalty0 (1):\penalty0 1, 2024.

\bibitem[Bak{\i}r et~al.(2024)Bak{\i}r, {\c{C}}ay{\i}r, and Navruz]{bakir2024comprehensive}
Halit Bak{\i}r, Ay{\c{s}}e~Nur {\c{C}}ay{\i}r, and Tu{\u{g}}ba~Selcen Navruz.
\newblock A comprehensive experimental study for analyzing the effects of data augmentation techniques on voice classification.
\newblock \emph{Multimedia Tools and Applications}, 83\penalty0 (6):\penalty0 17601--17628, 2024.

\bibitem[Mushtaq and Su(2020)]{mushtaq2020environmental}
Zohaib Mushtaq and Shun-Feng Su.
\newblock Environmental sound classification using a regularized deep convolutional neural network with data augmentation.
\newblock \emph{Applied Acoustics}, 167:\penalty0 107389, 2020.

\bibitem[Shen et~al.(2023)Shen, Liu, and Zhou]{shen2023mingling}
Siyuan Shen, Feng Liu, and Aimin Zhou.
\newblock Mingling or misalignment? temporal shift for speech emotion recognition with pre-trained representations.
\newblock In \emph{ICASSP 2023-2023 IEEE International Conference on Acoustics, Speech and Signal Processing (ICASSP)}, pages 1--5. IEEE, 2023.

\bibitem[Fathullah et~al.(2024)Fathullah, Wu, Lakomkin, Li, Jia, Shangguan, Mahadeokar, Kalinli, Fuegen, and Seltzer]{fathullah2024audiochatllama}
Yassir Fathullah, Chunyang Wu, Egor Lakomkin, Ke~Li, Junteng Jia, Yuan Shangguan, Jay Mahadeokar, Ozlem Kalinli, Christian Fuegen, and Mike Seltzer.
\newblock Audiochatllama: Towards general-purpose speech abilities for llms.
\newblock In \emph{Proceedings of the 2024 Conference of the North American Chapter of the Association for Computational Linguistics: Human Language Technologies (Volume 1: Long Papers)}, pages 5522--5532, 2024.

\bibitem[Vu et~al.(2024)Vu, Wang, Chen, Li, Zhao, Xing, and Chen]{vu2024gptvoicetasker}
Minh~Duc Vu, Han Wang, Jieshan Chen, Zhuang Li, Shengdong Zhao, Zhenchang Xing, and Chunyang Chen.
\newblock Gptvoicetasker: Advancing multi-step mobile task efficiency through dynamic interface exploration and learning.
\newblock In \emph{Proceedings of the 37th Annual ACM Symposium on User Interface Software and Technology}, pages 1--17, 2024.

\bibitem[Panagopoulou et~al.(2025)Panagopoulou, Xue, Yu, Li, Li, Joty, Xu, Savarese, Xiong, and Niebles]{panagopoulou2025x}
Artemis Panagopoulou, Le~Xue, Ning Yu, Junnan Li, Dongxu Li, Shafiq Joty, Ran Xu, Silvio Savarese, Caiming Xiong, and Juan~Carlos Niebles.
\newblock X-instructblip: A framework for aligning image, 3d, audio, video to llms and its emergent cross-modal reasoning.
\newblock In \emph{European Conference on Computer Vision}, pages 177--197. Springer, 2025.

\bibitem[Majhi and Saha(2024)]{majhi2024automatic}
Malay~Kumar Majhi and Sujan~Kumar Saha.
\newblock An automatic speech recognition system in odia language using attention mechanism and data augmentation.
\newblock \emph{International Journal of Speech Technology}, 27\penalty0 (3):\penalty0 717--728, 2024.

\bibitem[Rossenbach et~al.(2020)Rossenbach, Zeyer, Schl{\"u}ter, and Ney]{rossenbach2020generating}
Nick Rossenbach, Albert Zeyer, Ralf Schl{\"u}ter, and Hermann Ney.
\newblock Generating synthetic audio data for attention-based speech recognition systems.
\newblock In \emph{ICASSP 2020-2020 IEEE International Conference on Acoustics, Speech and Signal Processing (ICASSP)}, pages 7069--7073. IEEE, 2020.

\bibitem[Fan et~al.(2024)Fan, Zhang, Xu, Fang, Zhang, Zhao, and Yu]{fan2024transformer}
Huiting Fan, Xingnan Zhang, Yingying Xu, Jiangxiong Fang, Shiqing Zhang, Xiaoming Zhao, and Jun Yu.
\newblock Transformer-based multimodal feature enhancement networks for multimodal depression detection integrating video, audio and remote photoplethysmograph signals.
\newblock \emph{Information Fusion}, 104:\penalty0 102161, 2024.

\bibitem[Kachris(2025)]{kachris2025survey}
Christoforos Kachris.
\newblock A survey on hardware accelerators for large language models.
\newblock \emph{Applied Sciences}, 15\penalty0 (2):\penalty0 586, 2025.

\bibitem[Wen et~al.(2024{\natexlab{b}})Wen, Zhang, Niyato, Kang, Du, Zhang, and Han]{wen2024generative}
Jinbo Wen, Ruichen Zhang, Dusit Niyato, Jiawen Kang, Hongyang Du, Yang Zhang, and Zhu Han.
\newblock Generative ai for low-carbon artificial intelligence of things with large language models.
\newblock \emph{IEEE Internet of Things Magazine}, 8\penalty0 (1):\penalty0 82--91, 2024{\natexlab{b}}.

\bibitem[Luccioni et~al.(2024)Luccioni, Jernite, and Strubell]{luccioni2024power}
Sasha Luccioni, Yacine Jernite, and Emma Strubell.
\newblock Power hungry processing: Watts driving the cost of ai deployment?
\newblock In \emph{The 2024 ACM Conference on Fairness, Accountability, and Transparency}, pages 85--99, 2024.

\bibitem[Tan and Wang(2021)]{tan2021towards}
Ke~Tan and DeLiang Wang.
\newblock Towards model compression for deep learning based speech enhancement.
\newblock \emph{IEEE/ACM transactions on audio, speech, and language processing}, 29:\penalty0 1785--1794, 2021.

\bibitem[Cai and Li(2024)]{cai2024leveraging}
Danwei Cai and Ming Li.
\newblock Leveraging asr pretrained conformers for speaker verification through transfer learning and knowledge distillation.
\newblock \emph{IEEE/ACM Transactions on Audio, Speech, and Language Processing}, 2024.

\bibitem[Kim(2021)]{kim2021fpga}
Joo-Young Kim.
\newblock Fpga based neural network accelerators.
\newblock In \emph{Advances in Computers}, volume 122, pages 135--165. Elsevier, 2021.

\bibitem[Bell et~al.(2020)Bell, Fainberg, Klejch, Li, Renals, and Swietojanski]{bell2020adaptation}
Peter Bell, Joachim Fainberg, Ondrej Klejch, Jinyu Li, Steve Renals, and Pawel Swietojanski.
\newblock Adaptation algorithms for neural network-based speech recognition: An overview.
\newblock \emph{IEEE Open Journal of Signal Processing}, 2:\penalty0 33--66, 2020.

\bibitem[Bhattarai and Lee(2023)]{bhattarai2023comprehensive}
Bhuwan Bhattarai and Joonwhoan Lee.
\newblock A comprehensive review on music transcription.
\newblock \emph{Applied Sciences}, 13\penalty0 (21):\penalty0 11882, 2023.

\bibitem[Meerza et~al.(2024)Meerza, Liu, and Sun]{meerza2024harmonycloak}
Syed Irfan~Ali Meerza, Jian Liu, and Lichao Sun.
\newblock Harmonycloak: Making music unlearnable for generative ai.
\newblock In \emph{2025 IEEE Symposium on Security and Privacy (SP)}, pages 85--85. IEEE Computer Society, 2024.

\bibitem[Peng et~al.(2024{\natexlab{b}})Peng, Wang, Xi, Li, and Yu]{peng2024survey}
Jing Peng, Yucheng Wang, Yu~Xi, Xv~Li, and Kai Yu.
\newblock A survey on speech large language models.
\newblock \emph{arXiv preprint arXiv:2410.18908}, 2024{\natexlab{b}}.

\bibitem[Satish et~al.(2024)Satish, Naidu, Mogera, Karthik, et~al.]{satish2024voice}
EG~Satish, P~Ramesh Naidu, Girish~Madhava Mogera, HV~Karthik, et~al.
\newblock Voice over vision: A sequence-to-sequence model by text to speech technology.
\newblock In \emph{2024 First International Conference on Innovations in Communications, Electrical and Computer Engineering (ICICEC)}, pages 1--7. IEEE, 2024.

\bibitem[Chen et~al.(2024{\natexlab{b}})Chen, Xu, Zheng, Chen, Tolba, Zhao, Yu, and Feng]{chen2024evolution}
Zheyi Chen, Liuchang Xu, Hongting Zheng, Luyao Chen, Amr Tolba, Liang Zhao, Keping Yu, and Hailin Feng.
\newblock Evolution and prospects of foundation models: From large language models to large multimodal models.
\newblock \emph{Computers, Materials \& Continua}, 80\penalty0 (2), 2024{\natexlab{b}}.

\bibitem[Guo et~al.(2025)Guo, Yang, Zhang, Song, Zhang, Xu, Zhu, Ma, Wang, Bi, et~al.]{guo2025deepseek}
Daya Guo, Dejian Yang, Haowei Zhang, Junxiao Song, Ruoyu Zhang, Runxin Xu, Qihao Zhu, Shirong Ma, Peiyi Wang, Xiao Bi, et~al.
\newblock Deepseek-r1: Incentivizing reasoning capability in llms via reinforcement learning.
\newblock \emph{arXiv preprint arXiv:2501.12948}, 2025.

\bibitem[Qian et~al.(2024)Qian, Yin, and Dou]{qian2024reasoning}
Rui Qian, Xin Yin, and Dejing Dou.
\newblock Reasoning to attend: Try to understand how< seg> token works.
\newblock \emph{arXiv preprint arXiv:2412.17741}, 2024.

\bibitem[Jin et~al.(2024)Jin, Wang, Zhu, Wang, and Li]{jin2024pedestrian}
Jiandong Jin, Xiao Wang, Qian Zhu, Haiyang Wang, and Chenglong Li.
\newblock Pedestrian attribute recognition: A new benchmark dataset and a large language model augmented framework.
\newblock \emph{arXiv preprint arXiv:2408.09720}, 2024.

\bibitem[Liu and Nguyen(2024)]{liu2024rephrasing}
Jinghui Liu and Anthony Nguyen.
\newblock Rephrasing electronic health records for pretraining clinical language models.
\newblock \emph{arXiv preprint arXiv:2411.18940}, 2024.

\bibitem[Abane et~al.(2024)Abane, Bekri, and Battou]{abane2024fastrag}
Amar Abane, Anis Bekri, and Abdella Battou.
\newblock Fastrag: Retrieval augmented generation for semi-structured data.
\newblock \emph{arXiv preprint arXiv:2411.13773}, 2024.

\bibitem[Yang et~al.(2024{\natexlab{a}})Yang, Shi, Le, Hsu, and Tjandra]{yang2024audiobox}
Mu~Yang, Bowen Shi, Matthew Le, Wei-Ning Hsu, and Andros Tjandra.
\newblock Audiobox tta-rag: Improving zero-shot and few-shot text-to-audio with retrieval-augmented generation.
\newblock \emph{arXiv preprint arXiv:2411.05141}, 2024{\natexlab{a}}.

\bibitem[Fuad and Chen(2024)]{fuad2024llm}
Kazi Ahmed~Asif Fuad and Lizhong Chen.
\newblock Llm-ref: Enhancing reference handling in technical writing with large language models.
\newblock \emph{arXiv preprint arXiv:2411.00294}, 2024.

\bibitem[Wang et~al.(2024{\natexlab{d}})Wang, Xu, and Ren]{wang2024llm}
Zhenhua Wang, Guang Xu, and Ming Ren.
\newblock Llm-generated natural language meets scaling laws: New explorations and data augmentation methods.
\newblock \emph{arXiv preprint arXiv:2407.00322}, 2024{\natexlab{d}}.

\bibitem[Song et~al.(2024{\natexlab{b}})Song, Zhang, Tian, Yang, Huang, and Li]{song2024llm}
Yiping Song, Juhua Zhang, Zhiliang Tian, Yuxin Yang, Minlie Huang, and Dongsheng Li.
\newblock Llm-based privacy data augmentation guided by knowledge distillation with a distribution tutor for medical text classification.
\newblock \emph{arXiv preprint arXiv:2402.16515}, 2024{\natexlab{b}}.

\bibitem[Yang et~al.(2024{\natexlab{b}})Yang, Zhao, Huang, Li, and Xu]{yang2024latex}
Haoran Yang, Xiangyu Zhao, Sirui Huang, Qing Li, and Guandong Xu.
\newblock Latex-gcl: Large language models (llms)-based data augmentation for text-attributed graph contrastive learning.
\newblock \emph{arXiv preprint arXiv:2409.01145}, 2024{\natexlab{b}}.

\bibitem[Liu et~al.(2024{\natexlab{d}})Liu, Tao, Guo, and Yang]{liu2024improving}
Yizhu Liu, Ran Tao, Shengyu Guo, and Yifan Yang.
\newblock Improving topic relevance model by mix-structured summarization and llm-based data augmentation.
\newblock \emph{arXiv preprint arXiv:2404.02616}, 2024{\natexlab{d}}.

\bibitem[Cegin et~al.(2024{\natexlab{c}})Cegin, Pecher, Simko, Srba, Bielikova, and Brusilovsky]{cegin2024use}
Jan Cegin, Branislav Pecher, Jakub Simko, Ivan Srba, Maria Bielikova, and Peter Brusilovsky.
\newblock Use random selection for now: Investigation of few-shot selection strategies in llm-based text augmentation for classification.
\newblock \emph{arXiv preprint arXiv:2410.10756}, 2024{\natexlab{c}}.

\bibitem[Jia et~al.(2024)Jia, Wu, and Li]{jia2024curriculum}
Kaidi Jia, Yanxia Wu, and Rongsheng Li.
\newblock Curriculum-style data augmentation for llm-based metaphor detection.
\newblock \emph{arXiv preprint arXiv:2412.02956}, 2024.

\bibitem[Zeng(2024)]{zeng2024leveraging}
Linda Zeng.
\newblock Leveraging large language models for code-mixed data augmentation in sentiment analysis.
\newblock \emph{arXiv preprint arXiv:2411.00691}, 2024.

\bibitem[Litake et~al.(2024)Litake, Yagnik, and Labhsetwar]{litake2024inditext}
Onkar Litake, Niraj Yagnik, and Shreyas Labhsetwar.
\newblock Inditext boost: Text augmentation for low resource india languages.
\newblock \emph{arXiv preprint arXiv:2401.13085}, 2024.

\bibitem[Sahu et~al.(2023)Sahu, Vechtomova, Bahdanau, and Laradji]{sahu2023promptmix}
Gaurav Sahu, Olga Vechtomova, Dzmitry Bahdanau, and Issam~H Laradji.
\newblock Promptmix: A class boundary augmentation method for large language model distillation.
\newblock \emph{arXiv preprint arXiv:2310.14192}, 2023.

\bibitem[Chowdhury and Chadha(2023)]{chowdhury2023generative}
Arijit~Ghosh Chowdhury and Aman Chadha.
\newblock Generative data augmentation using llms improves distributional robustness in question answering.
\newblock \emph{arXiv preprint arXiv:2309.06358}, 2023.

\bibitem[Wang et~al.(2024{\natexlab{e}})Wang, Wang, Ni, Zhao, and Derr]{wang2024large}
Leyao Wang, Yu~Wang, Bo~Ni, Yuying Zhao, and Tyler Derr.
\newblock Large language model-based augmentation for imbalanced node classification on text-attributed graphs.
\newblock \emph{arXiv preprint arXiv:2410.16882}, 2024{\natexlab{e}}.

\bibitem[Ghosal et~al.(2023)Ghosal, Majumder, Mehrish, and Poria]{ghosal2023text}
Deepanway Ghosal, Navonil Majumder, Ambuj Mehrish, and Soujanya Poria.
\newblock Text-to-audio generation using instruction-tuned llm and latent diffusion model.
\newblock \emph{arXiv preprint arXiv:2304.13731}, 2023.

\bibitem[Manco et~al.(2024)Manco, Salamon, and Nieto]{manco2024augment}
Ilaria Manco, Justin Salamon, and Oriol Nieto.
\newblock Augment, drop \& swap: Improving diversity in llm captions for efficient music-text representation learning.
\newblock \emph{arXiv preprint arXiv:2409.11498}, 2024.

\bibitem[Li et~al.(2024{\natexlab{f}})Li, Xie, Xu, Guo, Yan, Zhang, Yu, and Wu]{li2024divesound}
Baihan Li, Zeyu Xie, Xuenan Xu, Yiwei Guo, Ming Yan, Ji~Zhang, Kai Yu, and Mengyue Wu.
\newblock Divesound: Llm-assisted automatic taxonomy construction for diverse audio generation.
\newblock \emph{arXiv preprint arXiv:2407.13198}, 2024{\natexlab{f}}.

\bibitem[Shu et~al.(2023)Shu, Zhang, Jiang, and Xie]{shu2023audio}
Fangxun Shu, Lei Zhang, Hao Jiang, and Cihang Xie.
\newblock Audio-visual llm for video understanding.
\newblock \emph{arXiv preprint arXiv:2312.06720}, 2023.

\bibitem[Huang et~al.(2023{\natexlab{b}})Huang, Ren, Huang, Yang, Ye, Zhang, Liu, Yin, Ma, and Zhao]{huang2023make}
Jiawei Huang, Yi~Ren, Rongjie Huang, Dongchao Yang, Zhenhui Ye, Chen Zhang, Jinglin Liu, Xiang Yin, Zejun Ma, and Zhou Zhao.
\newblock Make-an-audio 2: Temporal-enhanced text-to-audio generation.
\newblock \emph{arXiv preprint arXiv:2305.18474}, 2023{\natexlab{b}}.

\bibitem[Ok et~al.(2024)Ok, Yoo, and Lee]{ok2024audiobert}
Hyunjong Ok, Suho Yoo, and Jaeho Lee.
\newblock Audiobert: Audio knowledge augmented language model.
\newblock \emph{arXiv preprint arXiv:2409.08199}, 2024.

\bibitem[Lu et~al.(2024)Lu, Xie, Fu, Wen, Tao, Wang, Qi, Liu, Li, Liu, et~al.]{lu2024codecfake}
Yi~Lu, Yuankun Xie, Ruibo Fu, Zhengqi Wen, Jianhua Tao, Zhiyong Wang, Xin Qi, Xuefei Liu, Yongwei Li, Yukun Liu, et~al.
\newblock Codecfake: An initial dataset for detecting llm-based deepfake audio.
\newblock \emph{arXiv preprint arXiv:2406.08112}, 2024.

\bibitem[Das et~al.(2024)Das, Dingliwal, Ronanki, Paturi, Huang, Mathur, Yuan, Bekal, Niu, Jayanthi, et~al.]{das2024speechverse}
Nilaksh Das, Saket Dingliwal, Srikanth Ronanki, Rohit Paturi, Zhaocheng Huang, Prashant Mathur, Jie Yuan, Dhanush Bekal, Xing Niu, Sai~Muralidhar Jayanthi, et~al.
\newblock Speechverse: A large-scale generalizable audio language model.
\newblock \emph{arXiv preprint arXiv:2405.08295}, 2024.

\bibitem[Vallaeys et~al.(2024)Vallaeys, Shukor, Cord, and Verbeek]{vallaeys2024improved}
Th{\'e}ophane Vallaeys, Mustafa Shukor, Matthieu Cord, and Jakob Verbeek.
\newblock Improved baselines for data-efficient perceptual augmentation of llms.
\newblock \emph{arXiv preprint arXiv:2403.13499}, 2024.

\end{thebibliography}

\appendix

\section{Results and Discussion}

\subsection{LLM-Based Image Data Augmentation}

\begin{table*}[ht!]
\centering
\caption{List of Multi-modal LLMs in Image Data Augmentation (Preprints)}
\label{tab:surveyTableImagePreprint}
\fontsize{4.5}{4.5}\selectfont
\begin{tabular}{|p{1cm}|p{4cm}|p{4cm}|p{4cm}|}
\hline
\textbf{LLM Name} & \textbf{Augmentation Method} & \textbf{Application and Outcomes} & \textbf{Limitations}  \\ \hline
Image Augmentation Agent (IAA) \cite{wu2024image} & Utilizes LLMs and diffusion models to generate high-quality, diverse training images for weakly supervised semantic segmentation, incorporating a self-refinement mechanism for prompt and image quality control. & Demonstrates significant improvements in semantic segmentation on PASCAL VOC 2012 and MS COCO 2014 datasets by enriching training data diversity and quality. & Dependent on the sophisticated integration of LLMs and diffusion models, which requires careful calibration and high computational resources. \\ \hline
READ \cite{qian2024reasoning} & Empowers LLMs to discern 'where to attend' using a novel architecture, enhancing task-specific model integration through the innovative use of <SEG> tokens in multimodal contexts. & Introduces READ to improve reasoning by refining point-based activations from similarity maps, significantly boosting reasoning accuracy in visual tasks. & Relies on detailed similarity maps and may need precise calibration to ensure effectiveness across varied tasks. \\ \hline
T2Vid \cite{yin2024t2vid} & Explores enhancing video understanding in MLLMs using synthesized video-like samples for training, reducing reliance on extensive real video datasets. & Demonstrates efficient fine-tuning techniques with synthetic data to achieve superior performance, even with reduced sample sizes. & Depend heavily on the quality of synthetic data and its alignment with real video characteristics.\\ \hline 
LaB-RAG \cite{song2024lab} & Introduces Label Boosted Retrieval Augmented Generation (LaB-RAG), a method combining image-derived labels with retrieval-augmented generation to improve radiology report generation (RRG) without training deep learning models. & Demonstrates superior performance in both natural language and radiology-specific metrics, surpassing other retrieval-based and some fine-tuned methods. & Critiques commonly used RRG metrics and discusses the potential overvaluation of results, advocating for more accurate evaluation methods. \\ \hline 
DIAGen \cite{lingenberg2024diagen} & Introduces DIAGen, an approach to semantically diverse image augmentation using Gaussian noise and class-specific text prompts from LLMs, built on DA-Fusion. & Demonstrates that DIAGen enhances semantic diversity and improves classification performance, especially for out-of-distribution samples. & Discusses the fidelity-diversity tradeoff, using a weighting mechanism to mitigate the impact of poorly generated samples, thus enhancing the quality of the augmented dataset. \\ \hline 
T2Vid \cite{yin2024t2vid} & Investigates the use of pre-trained image LLMs for video understanding, exploring zero-shot inference and fine-tuning methods. & Develops the T2Vid method to synthesize video-like samples to improve instruction diversity in training, achieving comparable results with just 15\% of the sample size. & Highlights the capability of T2Vid to enhance long video understanding without actual long video data, aiming to refine video LLM training and data curation. \\ \hline 
ForgeryGPT \cite{li2024forgerygpt} & Introduces a novel multimodal LLM, ForgeryGPT, for image forgery detection and localization, integrating high-order forensics knowledge and explainable AI capabilities. & Develops innovative training strategies and architecture enhancements, such as the Mask-Aware Forgery Extractor for precise tampering detection and localization. & Demonstrates superior performance across multiple benchmarks, providing detailed, convincing explanations and supporting multi-turn dialogue, advancing the field towards robust, explainable image forgery analysis. \\ \hline 
Visual Editing GPT 3.5 Turb \cite{sultan2024visual} & Utilizes a distillation approach with data augmentation to improve fine-tuning in low-data regimes by 25\%. & Applied to real-time visual editing tasks, enabling effective color grading based on user input. Demonstrates cost and latency reduction while maintaining high performance. & Fixed sequential tool use, overly strict offline metrics, and limitation to one-hop responses may not fully capture detailed user preferences. \\ \hline
Pedestrian Attribute Recognition \cite{jin2024pedestrian} & Utilizes a Large Language Model augmented framework with a Vision Transformer backbone to enhance feature extraction and recognition in pedestrian attribute recognition. & Applied to the MSP60K dataset, achieving new state-of-the-art performances. The framework is validated across multiple PAR benchmark datasets. & Performance still highly dependent on computational resources. Limited by the need for extensive training data and the computational intensity of the models. \\ \hline
DALL-M: Context-Aware Clinical Data Augmentation with LLMs \cite{hsieh2024dall} & Employs LLMs to generate synthetic clinical data, enhancing context awareness in medical diagnostics. Introduces a novel three-phase feature generation process. & Demonstrates significant improvements in machine learning model performance in medical diagnostics using augmented features. & Relies on extensive clinical data for training, with potential computational and ethical challenges due to data sensitivity. \\ \hline
MM-Instruct: Generated Visual Instructions for Large Multimodal Model Alignment \cite{liu2024mm} & Leveraging existing LLMs to generate diverse visual instruction data from image captioning datasets. & Enhances the instruction-following capabilities of LMMs; significant performance improvements demonstrated with LLaVA-Instruct model. & Primarily focuses on generating instruction data; may not directly address other multimodal interactions or complex scenarios. \\ \hline

\end{tabular}
\end{table*}

\subsection{LLM-Based Text Data Augmentation}

\begin{table*}[ht!]
\centering
\caption{List of Multi-modal LLMs in Text Data Augmentation (Preprints)}
\label{tab:surveyTablePreprintsTextAudio}
\fontsize{4.5}{4.5}\selectfont
\begin{tabular}{|p{1cm}|p{4cm}|p{4cm}|p{4cm}|}
\hline
\textbf{LLM Name} & \textbf{Augmentation Method} & \textbf{Application and Outcomes} & \textbf{Limitations}  \\ \hline
\cite{kang2024synthetic} & Utilizes a chain-of-thought prompting approach with an LLM to generate synthetic summaries and sentiment analyses for improving depression prediction & Significantly enhances the prediction of depression severity while balancing dataset distributions & Dependent on the quality and relevance of initial transcripts for effective synthetic data generation \\ \hline
\cite{song2024lab} & Uses image descriptors as labels to enhance retrieval augmented generation for radiology report creation & Applied to radiology report generation, LaBRAG achieves superior results without fine-tuning DL models & Performance heavily reliant on the quality of image-derived labels and categorical accuracy \\ \hline
\cite{fischer2024swissadt} & Leverages video and textual data to improve AD translation for Swiss languages & Applied to ADT for German, French, Italian, and English, showing promising results in multilingual accessibility & Depends heavily on the quality and synchronization of video data for accurate AD translation \\  \hline 
\cite{glazkova2024evaluating} & Prompt-based data augmentation to detect green practices in Russian social media & Demonstrated effective use of LLM prompts for generating realistic text samples, improving multi-label classification of ecological texts & Limited to Russian language texts; further testing in other languages needed \\ \hline
\cite{wen2024aidbench} & AIDBench addresses the authorship identification capability of LLMs, introducing a benchmark with various datasets. It includes two methods, one-to-one and one-to-many identification, with a focus on privacy risks related to anonymous texts in systems like peer reviews. & AIDBench shows that LLMs can significantly exceed random chance in identifying authorship, revealing new privacy risks. & The study is limited to the datasets and methods used, which may not cover all aspects or types of texts, possibly affecting generalizability. \\ \hline 
\cite{alyafeai2024arabic} & Fine-tuning with synthetic dialogue data & Improves Arabic NLP performance on benchmarks with fewer parameters & Limited benchmarks for Arabic, overt tokenization issues \\ \hline
\cite{liu2024rephrasing} & Rephrasing EHRs using LLMs & Synthetic pretraining corpora improve language model performance & Limited real clinical text, potential LLM hallucinations \\ \hline
\cite{abane2024fastrag} & Schema and script learning in FastRAG & Improves accuracy and efficiency in network data processing & Challenges with complex queries and explicit entity retrieval \\ \hline
\cite{yang2024audiobox} & Retrieval-augmented TTA with Audiobox & Enhances zero-shot and few-shot TTA performance & Requires diverse audio retrieval, complex query handling \\ \hline
\cite{fuad2024llm} & Enhanced reference handling with LLM-Ref & Improves reference synthesis and handling in writing tools & Implements direct paragraph retrieval, avoiding chunking and indexing issues \\ \hline
LLM-Generated NL Meets Scaling Laws \cite{wang2024llm} & Introduces scaling laws for evaluating LLM-generated text & Proposes data augmentation methods for enhancing text classification & Uses fuzzy computing to assess data value, aligning it with human language standards \\ \hline
AugGPT: Leveraging ChatGPT for Text Data Augmentation \cite{dai2023auggpt} & Utilizes ChatGPT for generating auxiliary samples for few-shot text classification & Demonstrates double-digit improvements in sentence classification accuracy & Generates diversified and accurate augmented samples, improving model performance \\ \hline
LLMs vs Established Methods \cite{cegin2024llms} & Compares LLM-based and traditional text augmentation methods across multiple datasets and classifiers. & Finds LLMs advantageous mainly when using few seeds, with diminishing returns as seed count increases. & Suggests traditional methods often perform comparably or better in terms of cost-effectiveness and accuracy. \\ \hline
LLM2LLM \cite{lee2024llm2llm} & Introduces a novel iterative data enhancement technique using a teacher LLM to improve student LLM training. & Focuses on low-data regimes, significantly enhancing LLM performance by targeting incorrect predictions. & Demonstrates substantial performance improvements across several datasets, reducing the need for extensive data collection. \\ \hline
LLM-based Privacy Data Augmentation \cite{song2024llm} & Introduces a DP-based data augmentation method using LLMs and a DP-based discriminator for private domain text classification. & Employs a knowledge distillation approach for DP-based discrimination and introduces a DP-based tutor for distribution control. & Demonstrates significant performance improvements in text classification within private domains while ensuring privacy protection. \\ \hline
LATEX-GCL \cite{yang2024latex} & Introduces a novel framework utilizing LLMs for data augmentation in Text-Attributed Graphs (TAGs) for Graph Contrastive Learning (GCL), addressing limitations of conventional feature augmentation and improving semantic richness. & Proposes three types of textual augmentations via LLMs: shorten, rewriting, expansion. Uses carefully crafted prompts to guide LLMs, enhancing transparency and control over augmentation processes. & Shows superior performance on TAG datasets, effectively overcoming information loss and semantic deficits, thus setting a new standard for GCL applications in TAG settings. \\ \hline
GPT-4 \cite{liu2024improving} & Improving Real-Time Response in IoT Devices Using Edge Computing & Investigates the impact of integrating edge computing with IoT devices to enhance real-time data processing capabilities. & Proposes a new architecture that reduces latency and increases efficiency, validated through multiple real-world tests. \\ \hline
LLM-based Augmentation \cite{cegin2024use} & Random selection of samples for few-shot text augmentation & Explores few-shot sample selection strategies for enhancing classifier performance in LLM-based text augmentation, assessing effectiveness on both in-distribution and out-of-distribution data & While some informed strategies occasionally improve performance, random selection remains comparably effective, suggesting limited benefits from more complex selection methods. \\ \hline
Curriculum-style Augmentation \cite{jia2024curriculum} & Utilizes Curriculum-style Data Augmentation (CDA) to tackle data scarcity in metaphor detection by incrementally fine-tuning with progressively challenging data & Applies to metaphor detection to enhance the performance of open-source LLMs, reducing inference costs and improving efficiency & Limited by data imbalance and diminishing returns in later iterations, leading to potential declines in model performance over time. \\ \hline
DALDA \cite{jung2024dalda} & Leverages the LLM and Diffusion Model (DM) with Adaptive Guidance Scaling (AGS) for generating semantically rich synthetic images & Applied to enhance data augmentation in few-shot settings by dynamically adjusting the guidance weight based on CLIPScore, ensuring target distribution adherence & Faces challenges with data diversity management, potentially generating less diverse data under constrained guidance settings. \\ \hline
\cite{cegin2024effects} & Implements diversity incentives including taboo words, chaining, and hints in LLM paraphrasing & Applied in paraphrase generation to enhance the lexical diversity and performance of downstream classification models & Inconsistent improvements in lexical diversity and mixed results on downstream model performance, with only hints method showing significant positive outcomes. \\ \hline
\cite{zeng2024leveraging} & Employs LLMs to generate synthetic code-mixed data for sentiment analysis, enhancing model performance & Demonstrates the effectiveness of LLMs in improving sentiment analysis in multilingual contexts, particularly in Spanish-English setups & Significant F1 score improvement in Spanish-English; mixed results in Malayalam-English depending on baseline quality. \\  \hline
\cite{litake2024inditext} & Implements data augmentation techniques like EDA, Back Translation, and LLM-based methods for text classification in six Indian languages & Shows that basic data augmentation methods can outperform complex LLM techniques in scenarios of data scarcity & Finds that EDA and simple augmentation strategies often yield better results than LLM-based augmentation in text classification tasks. \\ \hline
 GPT3.5-turbo \cite{sahu2023promptmix} & Introduces PromptMix, a method that enhances data augmentation by generating challenging text augmentations near class boundaries and relabeling them for accuracy & Demonstrates that PromptMix can effectively improve knowledge transfer from large LLMs to smaller models, enhancing text classification in data-scarce scenarios & Finds that 2-shot PromptMix outperforms traditional 5-shot data augmentation methods across multiple datasets, indicating its efficiency in leveraging few-shot setups. \\ \hline
 GPT-3.5 \cite{chowdhury2023generative} & Explores the impact of LLM-generated datasets on the distributional robustness of QA models under natural distribution shifts & Demonstrates how "in-the-wild" generative data augmentation can enhance domain generalization in QA systems & Shows that models trained on a mix of real and generated data perform better on naturally shifted distributions compared to those trained solely on real data. \\ \hline
LLM-based Augmentation on Text-Attributed Graphs (LA-TAG) \cite{wang2024large} & Introduces LA-TAG, a novel method that leverages LLMs for text-based data augmentation to address class imbalance in text-attributed graphs & Demonstrates superior performance in node classification on various datasets compared to traditional non-textual data augmentation strategies & Shows that integrating synthetic text-attributed nodes into graphs significantly reduces the performance gap between minority and majority nodes. \\ \hline
\end{tabular}
\end{table*}

\subsection{LLM-Based Image Data Augmentation}

\begin{table*}[ht!]
\centering
\caption{List of Multi-modal LLMs in Speech Data Augmentation (Preprints)}
\label{tab:surveyTablePreprintsTextAudio}
\fontsize{4.5}{4.5}\selectfont
\begin{tabular}{|p{1cm}|p{4cm}|p{4cm}|p{4cm}|}
\hline
\textbf{LLM Name} & \textbf{Augmentation Method} & \textbf{Application and Outcomes} & \textbf{Limitations}  \\ \hline
AudioSetMix \cite{xu2024audiosetmix} & LLM-assisted augmentation for audio-language datasets & Enhanced performance on audio-language benchmarks by diversifying training examples and addressing data set limitations. & Relies heavily on the alignment and quality of LLM-genera\\ \hline
Synthio \cite{ghosh2024synthio} & Enhances small-scale audio classification datasets with synthetic data via text-to-audio models & Improves classification accuracy by generating acoustically consistent and compositionally diverse synthetic audio data & Depends on the alignment of synthetic audio with actual dataset characteristics for effective augmentation \\ \hline
LLM-powered Data Augmentation \cite{whitehouse2023llm} & Leverages LLMs to enhance data for multilingual commonsense reasoning tasks & Demonstrates that LLM-generated data can effectively improve the performance of multilingual models on limited datasets & Highlights challenges in generating coherent data in some languages like Tamil; shows LLMs' uneven performance across different languages \\ \hline
Text-to-Audio Generation \cite{ghosal2023text} & Introduces the TANGO model, leveraging instruction-tuned LLMs for effective text-to-audio generation, significantly outperforming previous models on a smaller dataset & Highlights the integration of FLAN-T5 for enhanced text comprehension and audio generation, with specific focus on maintaining model diversity without fine-tuning during training & Demonstrates improved model performance through innovative audio mixing techniques based on audio pressure levels \\ \hline
Audio Dialogues \cite{goel2024audio} & Introduces a novel dataset for enhancing audio and music understanding through multi-turn dialogues, consisting of over 163,000 samples, aimed at training models to handle complex interactive tasks in audio analysis. & Utilizes a unique prompting-based approach with existing datasets to create dialogues that explore the subtleties of audio events, enabling deeper model training on enhanced audio features. & Provides an extensive evaluation of the dataset's utility, demonstrating its potential to significantly advance the field of audio interaction models. \\ \hline
UniAudio \cite{yang2023uniaudio} & Introduces a universal audio generation model capable of handling multiple audio tasks including speech, sound, music, and singing synthesis with a single model architecture. & Utilizes innovative LLM techniques for effective tokenization and sequence prediction, enhancing the model's ability to handle complex audio generation tasks efficiently. & Demonstrates superior performance across various audio tasks, supported by extensive training on diverse datasets, showcasing its adaptability and potential as a foundational model for future audio applications. \\ \hline
Augment, Drop and Swap \cite{manco2024augment} & Introduces Augmented View Dropout and TextSwap techniques for diversifying text inputs in music-text contrastive learning & Applied to music-text representation learning, demonstrating significant improvements in model robustness and retrieval accuracy without additional computational costs & Does not address potential biases or inconsistencies introduced by synthetic text augmentation and relies on the quality of initial tag annotations for effective training \\ \hline
DiveSound \cite{li2024divesound} & Employs LLMs to automatically construct a taxonomy for diverse audio generation, integrating multimodal data. & Enhances audio diversity in generation tasks by utilizing a new dataset with diverse subcategories informed by visual and textual data, improving sound quality and diversity. & The framework's effectiveness can be limited by the accuracy of the LLMs in generating coherent and relevant subcategories and the alignment of multimodal data. \\ \hline
Audio-Agent \cite{wang2024audio} & Utilizes a pre-trained TTA diffusion model and GPT-4 for decomposing text into specific instructions for audio generation; employs Gemma2-2B-it for bridging video and audio modalities with temporal alignment. & Provides high-quality audio generation aligned with complex textual or video inputs, supporting variable-length and multimodal audio generation tasks. & The method's dependency on the quality of LLM decompositions and the accuracy of semantic and temporal alignments might limit its application in more dynamic or unpredictable scenarios. \\ \hline
Audio-Visual LLM \cite{shu2023audio} & Integrates modality-specific tokens to activate visual or auditory encoders for holistic video understanding; includes a high-quality video instruction dataset from GPT-4. & Demonstrates strong zero-shot performance on diverse video understanding tasks like MSRVTT-QA, significantly surpassing non-LLM methods. & May be constrained by the reliance on accurate modality-specific token activation and the dataset's quality for training effectiveness. \\ \hline
Audio-Visual LLM \cite{shu2023audio} & Integrates modality-specific tokens to activate visual or auditory encoders for holistic video understanding; includes a high-quality video instruction dataset from GPT-4. & Demonstrates strong zero-shot performance on diverse video understanding tasks like MSRVTT-QA, significantly surpassing non-LLM methods. & May be constrained by the reliance on accurate modality-specific token activation and the dataset's quality for training effectiveness. \\ \hline    Contextual ASR \cite{lei2024contextualization} & Uses contextual cues to enhance ASR accuracy & Significant reduction in word and entity error rates in ASR systems & Relies heavily on precise initial entity recognition \\
\hline
Make-An-Audio 2 \cite{huang2023make} & Employs structured text parsing and temporal alignment for audio synthesis & Improved alignment and audio quality; surpasses existing models & Dependent on the precision of initial text parsing \\
\hline
AudioBERT \cite{ok2024audiobert} & Enhances BERT by injecting auditory knowledge & Integrates auditory commonsense into BERT, enhancing model performance & Requires accurate detection of auditory knowledge spans \\
\hline
Codecfake \cite{lu2024codecfake} & Utilizes neural codecs for waveform manipulation in deepfake detection & Efficiently detects deepfake audios by reducing equal error rate & Struggles to detect without traditional vocoder artifacts \\
\hline
SpeechVerse \cite{das2024speechverse} & Merges speech and text foundation models for robust multi-task training & Outperforms traditional models in diverse speech processing tasks & Needs further adjustments for consistent performance across tasks \\
\hline
Enhancing Temporal Understanding in AQA for LALMs \cite{sridhar2024enhancing} & Focuses on temporal reasoning in audio QA through curriculum learning & Introduces effective strategies and a novel metric, improving temporal reasoning & Performance gain without losing capabilities in other areas \\
\hline
Improved Baselines for Data-efficient Perceptual Augmentation of LLMs \cite{vallaeys2024improved} & Studies interfacing mechanisms between LLMs and perceptual backbones & Introduces DePALM, enhancing performance with reduced training time & Demonstrates advancements in LLM interface techniques in low-data settings \\
\hline

\end{tabular}
\end{table*}

\end{document}